\documentclass[conference]{IEEEtran}

\usepackage{graphicx}
\usepackage{booktabs}
\usepackage{multirow}
\usepackage{amsmath}
\usepackage{amssymb}
\usepackage{microtype}
\usepackage{xcolor}
\usepackage{hyperref}
\usepackage{cleveref}
\usepackage{array}
\usepackage{dblfloatfix}

\hypersetup{
    colorlinks=true,
    linkcolor=blue,
    citecolor=blue,
    urlcolor=blue
}

\newcommand{\pivot}{\textsc{PIVOT}}

\title{\textbf{PIVOT: A Multi-Trajectory Dataset and Testbed for Pose, Intrinsics, and Novel Viewpoint Evaluation in Real-World 3D Reconstruction}}
\author{
\IEEEauthorblockN{Mary Raymond}
\IEEEauthorblockA{
\textit{Independent Researcher}\\
mary.raymond.n@gmail.com
}
}
\date{August 2026}

\begin{document}
\maketitle

\begin{abstract}
Neural radiance fields (NeRFs), 3D Gaussian Splatting (3DGS), and related novel-view synthesis methods are commonly evaluated under capture and reconstruction conditions that can be substantially cleaner than those encountered by robots, drones, and autonomous systems. In particular, benchmark pipelines may rely on reconstruction-friendly camera trajectories, offline-optimized camera poses, scene-optimized intrinsics, and held-out views sampled from trajectories already represented during training. These choices make evaluation convenient, but couple several favorable assumptions and can obscure how reconstruction systems behave when deployed with measured poses, reusable camera calibration, or structurally different camera paths.

We introduce \pivot{} (Pose, Intrinsics and Viewpoint Oriented Testbed), a multi-trajectory dataset, processing pipeline, and evaluation framework for studying these factors independently. \pivot{} represents each scene using deliberately different camera trajectories and retains, where available, both sensor-derived measured poses and COLMAP-optimized poses for each frame, together with physical camera calibration and scene-optimized intrinsics. The testbed defines three benchmark families: (1) seen versus unseen trajectory novel-view generalization, (2) measured versus optimized pose sensitivity, and (3) calibrated versus optimized intrinsic sensitivity. We additionally introduce a directed pose-space Chamfer distance for describing how well a training pose distribution covers an evaluation trajectory. \pivot{} v1 contains five real-world scenes captured using a DJI Mini 4 Pro drone and provides an open processing and Nerfstudio-based evaluation toolchain intended to be reusable for additional scenes and capture devices.

This manuscript accompanies the initial \pivot{} release. Results from the benchmarks show a consistent quality gap between held-out views on represented trajectories and views from unseen trajectories. They also show substantial sensitivity to pose source and to the camera intrinsics used for reconstruction. 
\end{abstract}

\section{Introduction}
Modern 3D reconstruction and novel-view synthesis systems are increasingly relevant to robotics, autonomous platforms, inspection, mapping, and embodied perception. Yet common experimental pipelines can implicitly assume access to conditions that are difficult to reproduce online. A scene may be captured using a smooth inward-looking orbit with high image overlap; camera poses may be recovered and globally optimized offline by Structure-from-Motion (SfM); camera intrinsics may be optimized independently for every scene; and evaluation may use held-out images sampled from the same trajectory family as the training data.

These assumptions are individually reasonable for reconstruction research, but together they can make the experimental setting substantially cleaner than the conditions encountered by a deployed system. A robot or drone may instead receive poses from GPS/IMU, visual(-inertial) odometry, SLAM~\cite{campos2021orbslam3}, LiDAR, radar, or another online localization source. It may reuse a physical camera calibration across scenes. Its motion may follow traversals rather than reconstruction-friendly orbits. Most importantly, a requested novel view may lie on a camera path that is structurally different from the paths represented during training.

\pivot{} is designed to make these differences explicit, measurable, and reproducible. Rather than treating a scene as one undifferentiated image collection, \pivot{} makes the \emph{trajectory} a first-class unit of capture, processing, visualization, export, and evaluation. Each scene contains multiple named and typed trajectories covering reconstruction-friendly, robot-like, and extrapolation-oriented motion. Processed frames can retain both a sensor-derived measured pose and a COLMAP-optimized pose, while trajectories can retain both physical/offline camera calibration and COLMAP-optimized intrinsics.

\begin{figure*}[!t]
    \centering
    \includegraphics[width=\textwidth]{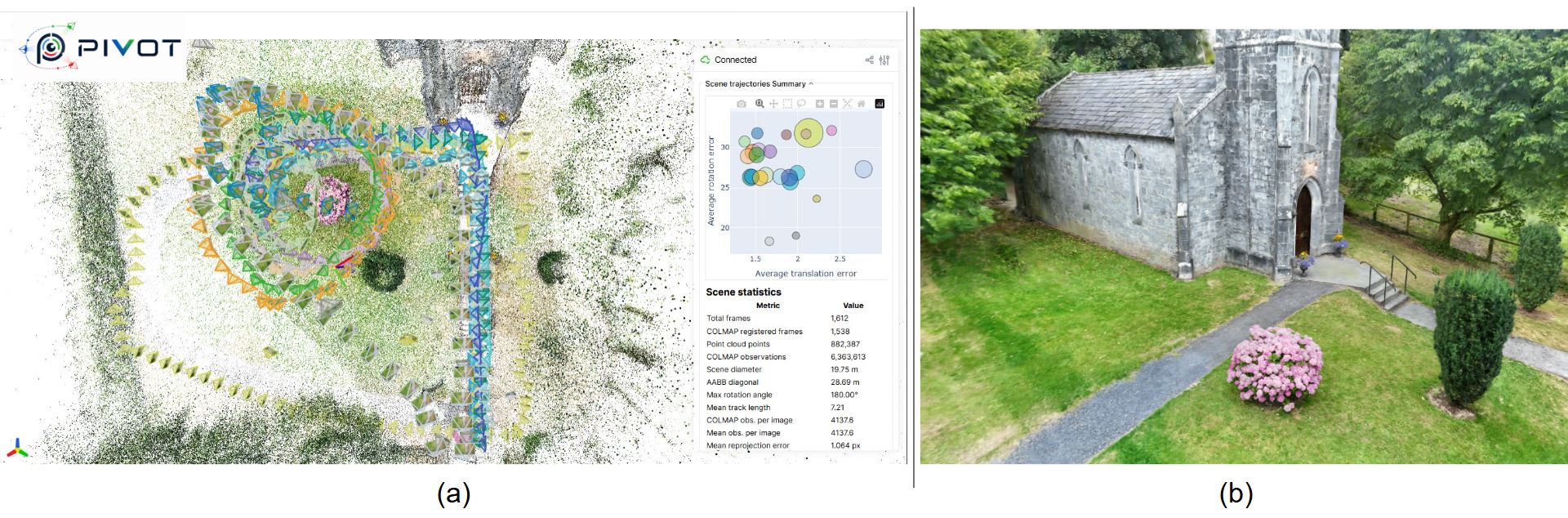}
    \caption{Overview of the \pivot{} testbed and reconstruction pipeline. (a) \pivot{}'s interactive trajectory-aware viewer jointly visualizes the COLMAP sparse reconstruction, camera trajectories, trajectory-level pose errors, and scene statistics. (b) Example Splatfacto reconstruction of the same scene, trained using 80\% of the frames from all trajectories with COLMAP-optimized poses and intrinsics. \pivot{} preserves trajectory identity together with measured and optimized camera parameters to support controlled reconstruction evaluation.}
    \label{fig:pivot_overview}
\end{figure*}

Figure~\ref{fig:pivot_overview} provides an overview of the trajectory-aware \pivot{} representation and an example reconstruction produced from the processed scene.

The resulting testbed supports four central questions:
\begin{enumerate}
    \item How well does a reconstruction model generalize to camera trajectories that are structurally different from its training trajectories?
    \item How does reconstruction quality change as an evaluation trajectory moves farther from the training pose distribution?
    \item How much reconstruction quality is gained by replacing measured poses with offline-optimized poses?
    \item How much does per-scene intrinsic optimization improve over a reusable physical camera calibration?
    
\end{enumerate}

The principal contributions of this work are:
\begin{itemize}
    \item a reusable multi-trajectory scene capture specification and processed dataset representation;
    \item a dual-pose representation that stores sensor-derived measured poses and COLMAP-optimized poses side by side;
    \item support for both calibrated and scene-optimized camera intrinsics;
    \item a directed pose-space Chamfer distance for quantifying evaluation-trajectory coverage relative to training poses;
    \item an end-to-end raw-data processing, visualization, export, and Nerfstudio integration toolchain; and
    \item three benchmark families for novel-view trajectory generalization, pose-source sensitivity, and intrinsic-source sensitivity.
\end{itemize}

\pivot{} does not propose a new NeRF, 3DGS, or SfM algorithm. Its purpose is to provide a controlled testbed for studying reconstruction under more realistic capture and evaluation conditions.

\section{Related Work}
\label{sec:related}

\paragraph{Novel-view synthesis and neural reconstruction.}
NeRF and subsequent neural rendering methods established high-quality novel-view synthesis from posed image collections \cite{mildenhall2020nerf}. More recent explicit scene representations, including 3D Gaussian Splatting, provide high-quality rendering with substantially different optimization and rendering characteristics \cite{kerbl20233dgs}. \pivot{} is model-agnostic at the dataset level; the initial benchmark integration targets Nerfacto and Splatfacto through Nerfstudio.

\paragraph{Camera pose estimation and Structure-from-Motion.}
COLMAP provides a widely used SfM and multi-view geometry pipeline for camera registration and sparse reconstruction \cite{schoenberger2016sfm,schoenberger2016mvs}. Many reconstruction datasets and pipelines use SfM-optimized camera parameters as model inputs. \pivot{} retains these optimized estimates while also preserving sensor-derived measured poses, enabling controlled experiments in which translation and rotation sources can be independently selected.

\paragraph{Reconstruction benchmarks and trajectory generalization.}
Real-world novel-view synthesis datasets span several capture regimes. LLFF introduced forward-facing real-world captures together with practical sampling guidance \cite{mildenhall2019llff}, while the NeRF synthetic and LLFF-style evaluations helped establish interpolation-oriented held-out-view protocols. Mip-NeRF 360 extended evaluation to challenging unbounded 360-degree scenes \cite{barron2022mipnerf360}. These datasets have been important for measuring rendering quality, but evaluation commonly samples test views from the same capture distribution used to construct the scene. \pivot{} instead makes the trajectory an explicit experimental unit and includes complete, independently captured trajectories whose motion structure differs from the training paths. The goal is not to replace existing NVS benchmarks, but to complement them with a controlled way to distinguish within-trajectory interpolation from trajectory-level generalization.

\paragraph{Calibration and pose robustness.}
Several neural reconstruction methods relax the assumption of perfectly known cameras by jointly optimizing scene representation and camera parameters. BARF jointly refines camera poses and a NeRF representation from imperfect initialization \cite{lin2021barf}, while NeRF-- jointly optimizes both camera intrinsics and poses \cite{wang2021nerfminusminus}. Such methods demonstrate that camera uncertainty can be absorbed or corrected during offline optimization. \pivot{} asks a complementary deployment-oriented question: what happens when reconstruction is intentionally evaluated using sensor-derived poses or a reusable physical calibration rather than allowing scene-specific camera optimization?

\paragraph{Reconstruction frameworks.}
Nerfstudio provides a modular framework for NeRF development and includes Nerfacto as a practical combination of established components \cite{tancik2023nerfstudio}. \pivot{} uses Nerfstudio as the initial benchmark backend and adds trajectory-aware export and evaluation so that the same processed scene can be tested under controlled pose, intrinsic, and viewpoint conditions.

\section{PIVOT Testbed Design}
\subsection{Scene as a Collection of Trajectories}

A \pivot{} scene is represented as a collection of deliberately different camera trajectories rather than a single reconstruction-friendly path. The trajectory protocol records properties including motion type, altitude band, whether the path is closed, camera direction, lens type, image resolution, capture mode, and capture device.

The core trajectory families include inward-looking orbits at multiple altitudes, outward-looking orbit at low altitude, directional traversals, a closed traversal loop, bird's-eye-view capture, vertical ascent, scattered still viewpoints, and 360-degree panorama stations. Optional trajectories extend the same design with outward-looking orbits, additional traversal altitudes, and additional scattered viewpoints.

This structure is intended to support both conventional reconstruction-friendly coverage and motion patterns that better resemble deployed robotic or aerial systems.

\subsection{Dual Pose Representation}

For each processed frame, \pivot{} can store:
\begin{align}
\mathbf{T}^{\mathrm{measured}}_{c2w} &: \text{sensor-derived measured camera-to-world pose},\\
\mathbf{T}^{\mathrm{COLMAP}}_{c2w} &: \text{COLMAP-optimized camera-to-world pose}.
\end{align}

Measured poses are derived from capture-device metadata without scene-level pose optimization. For the DJI Mini 4 Pro drone capture path used in \pivot{} v1, GPS position, flight attitude, and gimbal attitude are converted into a North-East-Down (NED) world frame and then into the OpenGL-style camera convention used by the processed dataset.

The COLMAP reconstruction uses measured positions as soft position priors. This allows the reconstructed model to benefit from SfM optimization while retaining a common spatial relationship with the measured trajectory.

The dual representation allows downstream experiments to independently select measured or optimized translation and rotation. We denote the four pose-source configurations as:
\begin{center}
\begin{tabular}{lll}
\toprule
Configuration & Translation & Rotation\\
\midrule
OO & optimized & optimized\\
OM & optimized & measured\\
MO & measured & optimized\\
MM & measured & measured\\
\bottomrule
\end{tabular}
\end{center}

\subsection{Dual Intrinsic Representation}

A physical robotic system generally carries a camera whose calibration is reused across scenes, whereas an offline reconstruction pipeline may optimize camera intrinsics for each scene. \pivot{} therefore retains both a physical/offline camera calibration and COLMAP-optimized per-scene intrinsics. The exporter can select the intrinsic source independently of the pose source.

\subsection{Processing Pipeline}

\begin{figure*}[!t]
    \centering
    \includegraphics[width=\textwidth]{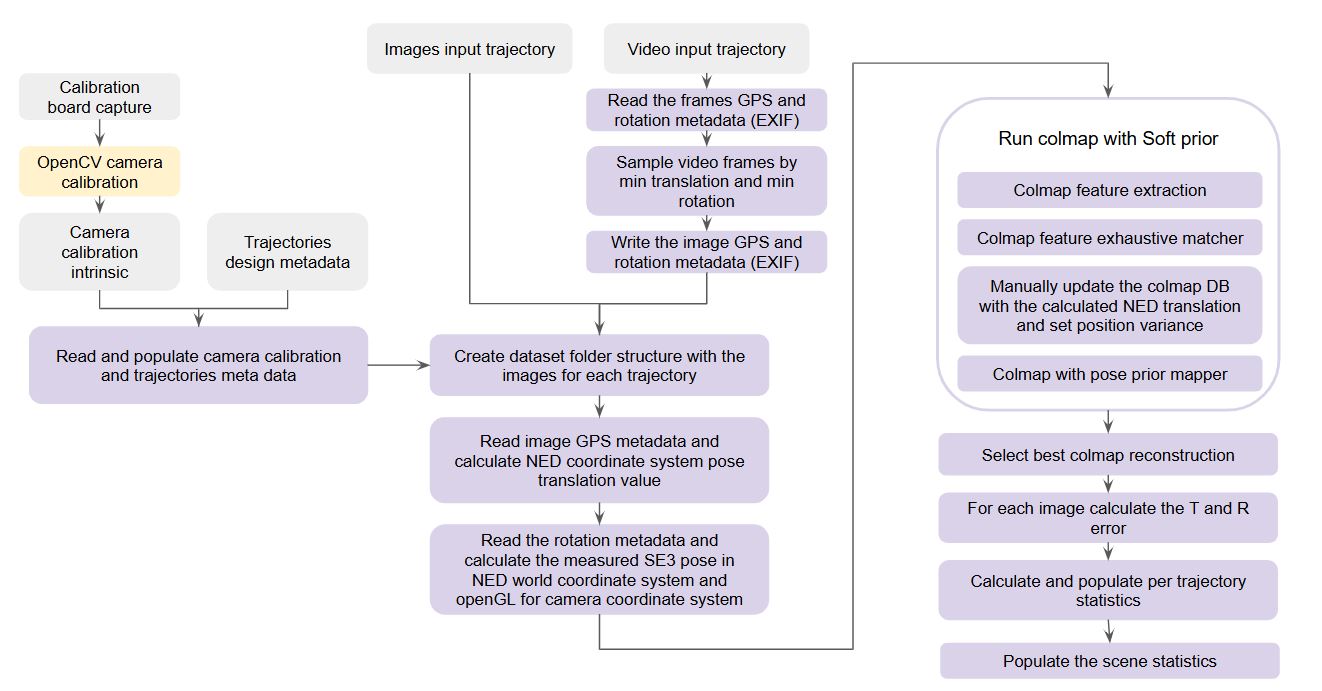}
    \caption{PIVOT raw-data processing pipeline.}
    \label{fig:raw_processing_pipeline}
\end{figure*}

The \pivot{} raw-data pipeline illustrated in
Figure~\ref{fig:raw_processing_pipeline}, transforms trajectory captures into a processed scene while preserving trajectory identity. The main stages are:
\begin{enumerate}
    \item read trajectory metadata and raw video/photo captures;
    \item sample video frames using translation and rotation thresholds, or use captured still images directly;
    \item extract and write EXIF/XMP metadata;
    \item compute measured camera poses from device position and orientation metadata;
    \item transform poses into the NED world frame and OpenGL camera convention;
    \item run COLMAP feature extraction and matching;
    \item inject measured camera positions and covariance as soft priors;
    \item run COLMAP pose-prior mapping and select the best reconstruction;
    \item retain optimized poses and intrinsics alongside measured poses and calibrated intrinsics;
    \item compute per-frame pose errors and trajectory/scene statistics; and
    \item compute the directed trajectory-distance matrix and export the processed scene.
\end{enumerate}

The processing core exposes capture-device metadata interfaces so that devices other than the DJI Mini 4 Pro drone can be integrated by implementing the required metadata mapping and image/video pose readers.

\section{Directed Pose Chamfer Distance}
\label{sec:chamfer}

To describe how well one set of camera poses covers another, \pivot{} extends the Chamfer distance commonly used for comparing point sets~\cite{fan2017point} to a directed pose-space measure. Let $A$ be an evaluation trajectory and $B$ a reference or training pose set. The directed distance is

\begin{equation}
D(A \rightarrow B)
=
\frac{1}{|A|}
\sum_{a \in A}
\operatorname{kNNDistance}(a,B).
\end{equation}

The pose distance combines normalized translation and rotation components. Translation-only and rotation-only variants are also retained. Because the measure is directed,

\begin{equation}
D(A \rightarrow B) \neq D(B \rightarrow A),
\end{equation}

which is intentional: the question ``how well does the training pose set cover the evaluation trajectory?'' is different from asking how well the evaluation trajectory covers the training set.

At scene-processing time, pairwise trajectory distances are stored as a matrix for visualization and experiment design. During reconstruction evaluation, the same formulation is used to measure each evaluation trajectory against the complete training pose set.

\paragraph{Interpretation.}
The metric should be interpreted as a \emph{pose-space coverage descriptor}, not as a direct measure of novel-view difficulty. The experiments in this work test whether increasing pose-space displacement is empirically associated with reconstruction-quality degradation.

\section{Dataset}
\subsection{PIVOT v1}

\pivot{} v1 is built around five real-world scenes captured using a DJI Mini 4 Pro drone. Released processed scenes contain trajectory images, per-frame measured and COLMAP-optimized poses, calibrated and optimized camera intrinsics, trajectory statistics, scene statistics, and sparse reconstruction assets.

\begin{table}[t]
\centering
\caption{PIVOT v1 scene summary. Frontyard and Backyard statistics are pending completion of the currently running processing/benchmark jobs and will be populated before release.}
\label{tab:dataset}
\small
\begin{tabular}{>{\raggedright\arraybackslash}p{0.14\linewidth}>{\centering\arraybackslash}p{0.08\linewidth}>{\centering\arraybackslash}p{0.1\linewidth}>{\centering\arraybackslash}p{0.08\linewidth}>{\centering\arraybackslash}p{0.12\linewidth}>{\centering\arraybackslash}p{0.09\linewidth}>{\centering\arraybackslash}p{0.08\linewidth}}
\toprule
Scene & Frames & Registered & Reg. (\%) & Sparse pts. & AABB (m) & Reproj. (px)\\
\midrule
Church            & 1,612& 1,538& 95.4& 882,387& 28.68&  1.06\\
Village Street    & 1,733& 1,726& 99.5& 1,430,217& 55.12& 0.89\\
Victorian Garden  & 1,547&  1,536& 99.2& 767,781& 44.62& 0.92\\
Frontyard         & 920& 913& 99.2& 719,123& 13.16& 1.09\\
Backyard          & 1,536& 1,527& 99.4& 110,1253& 25.3& 0.97\\
\bottomrule
\end{tabular}
\end{table}

Some \pivot{} trajectories are intentionally difficult for SfM. Consequently, the dataset records both total frames and COLMAP-registered frames. Registration rate is treated as useful information about the interaction between trajectory design and SfM rather than merely as a preprocessing detail.

\subsection{Trajectory Taxonomy}

\begin{table}[t]
\centering
\caption{Representative core trajectory types in PIVOT.}
\label{tab:trajectories}
\small
\begin{tabular}{llll}
\toprule
Trajectory & Motion & Altitude & Camera direction\\
\midrule
orbit\_inward\_low & orbit & low & scene inward\\
orbit\_inward\_mid & orbit & mid & scene inward\\
orbit\_inward\_high & orbit & high & scene inward\\
orbit\_outward\_low & orbit & low & scene outward\\
traversal\_forward\_low & traversal & low & along track\\
traversal\_backward\_low & traversal & low & along track\\
traversal\_left\_low & traversal & low & along track\\
traversal\_right\_low & traversal & low & along track\\
traverse\_loop\_low & closed traversal & low & along track\\
bev\_orbit\_area & BEV orbit & high & nadir\\
bev\_traverse\_area & BEV traverse & high & nadir\\
rocket\_upward & vertical ascent & low--high & scene inward\\
scattered\_low & scattered & low & multi-angle\\
panorama\_360\_station & panorama & low--mid & 360$^\circ$ sweep\\
\bottomrule
\end{tabular}
\end{table}

\section{Experimental Protocol}
\label{sec:experiments}

The \pivot{} evaluation uses the separate PIVOT Nerfstudio integration environment and targets Nerfacto and Splatfacto. Evaluation is performed per trajectory and reports SSIM~\cite{wang2004ssim}, PSNR, LPIPS~\cite{zhang2018lpips}, and directed pose Chamfer distance relative to the training pose set.

For reproducibility, the release will freeze the PIVOT and Nerfstudio-integration commits together with the model configurations, training iteration counts, random seeds, image resolution/scaling settings, and exact train/evaluation trajectory selections used for all reported runs. These values are recorded from the executed benchmark configuration rather than reconstructed after the fact.

\subsection{Benchmark 1: Seen vs. Unseen Trajectories}

This benchmark asks how reconstruction quality changes when evaluation moves from held-out frames on trajectories represented during training to complete camera trajectories not represented during training.

Training uses a mixture of inward-orbit and traversal trajectories. Evaluation is divided into:
\begin{itemize}
    \item \textbf{seen trajectories}: held-out frames from trajectories represented in training;
    \item \textbf{unseen trajectories}: complete trajectories absent from the training set.
\end{itemize}

Per-trajectory image-quality metrics are reported together with directed pose Chamfer distance to the training pose set.
The number of training iterations for benchmark 1 is 60k

\subsection{Benchmark 2: Measured vs. Optimized Poses}

This benchmark asks how strongly reconstruction quality depends on offline pose optimization. Translation and rotation are independently selected from measured or optimized estimates, producing OO, OM, MO, and MM conditions. This separation is intended to reveal whether translation error, rotation error, or their combination dominates reconstruction degradation.
The number of training iterations for benchmark 2 is 30k

\subsection{Benchmark 3: Calibrated vs. Optimized Intrinsics}

This benchmark asks how much benefit is obtained by allowing COLMAP to optimize camera intrinsics for a scene rather than using the camera's precomputed physical calibration. Pose source is held fixed while the intrinsic source changes.
The number of training iterations for benchmark 3 is 30k

\section{Quantitative Results}
\label{sec:quant}

\subsection{Novel-View Trajectory Generalization}

Table~\ref{tab:bm1_quantitative} summarizes BM1 reconstruction quality for seen and unseen evaluation trajectories. The table reports aggregate SSIM, LPIPS, and PSNR for Nerfacto and Splatfacto.

\begin{table*}[!tbp]
\centering
\caption{BM1 quantitative results comparing seen and unseen evaluation trajectories.
Higher SSIM and PSNR are better ($\uparrow$), while lower LPIPS is better ($\downarrow$).}
\label{tab:bm1_quantitative}
\small
\begin{tabular}{ll l c c c}
\hline
\textbf{Scene} &
\textbf{Model} &
\textbf{Eval. Type} &
\textbf{SSIM $\uparrow$} &
\textbf{LPIPS $\downarrow$} &
\textbf{PSNR $\uparrow$} \\
\hline

% ---------------- Church ----------------
\multirow{4}{*}{church}
& \multirow{2}{*}{Nerfacto}
    & Seen avg.   & 0.44 & 0.53 & 19.53 \\
&   & Unseen avg. & 0.31 & 0.70 & 16.47 \\
\cline{2-6}
& \multirow{2}{*}{Splatfacto}
    & Seen avg.   & 0.65 & 0.30 & 22.16 \\
&   & Unseen avg. & 0.40 & 0.57 & 16.86 \\
\hline

% ---------------- Victorian Garden ----------------
\multirow{4}{*}{victorian\_garden}
& \multirow{2}{*}{Nerfacto}
    & Seen avg.   & 0.33 & 0.62 & 17.00 \\
&   & Unseen avg. & 0.29 & 0.71 & 15.15 \\
\cline{2-6}
& \multirow{2}{*}{Splatfacto}
    & Seen avg.   & 0.51 & 0.38 & 19.22 \\
&   & Unseen avg. & 0.38 & 0.56 & 14.99 \\
\hline

% ---------------- Village Street ----------------
\multirow{4}{*}{village\_street}
& \multirow{2}{*}{Nerfacto}
    & Seen avg.   & 0.57 & 0.47 & 19.79 \\
&   & Unseen avg. & 0.45 & 0.61 & 16.81 \\
\cline{2-6}
& \multirow{2}{*}{Splatfacto}
    & Seen avg.   & 0.80 & 0.22 & 24.50 \\
&   & Unseen avg. & 0.56 & 0.49 & 16.07 \\
\hline

% ---------------- Frontyard ----------------
\multirow{4}{*}{frontyard}
& \multirow{2}{*}{Nerfacto}
    & Seen avg.   & 0.51 & 0.44 & 19.64 \\
&   & Unseen avg. & 0.42 & 0.59 & 17.02 \\
\cline{2-6}
& \multirow{2}{*}{Splatfacto}
    & Seen avg.   & 0.74 & 0.18 & 23.40 \\
&   & Unseen avg. & 0.53 & 0.43 & 16.96 \\
\hline

% ---------------- Backyard ----------------
\multirow{4}{*}{backyard}
& \multirow{2}{*}{Nerfacto}
    & Seen avg.   & 0.49 & 0.58 & 17.83 \\
&   & Unseen avg. & 0.38 & 0.70 & 16.17 \\
\cline{2-6}
& \multirow{2}{*}{Splatfacto}
    & Seen avg.   & 0.73 & 0.29 & 22.12 \\
&   & Unseen avg. & 0.46 & 0.56 & 15.63 \\
\hline

\end{tabular}
\end{table*}

\subsection{Trajectory Distance and Reconstruction Quality}

\begin{figure*}[!tbp]
    \centering
    \setlength{\tabcolsep}{3pt}
    \renewcommand{\arraystretch}{1.0}

    \begin{tabular}{c c}
        \textbf{Nerfacto} & \textbf{Splatfacto} \\

        \includegraphics[width=0.35\textwidth]{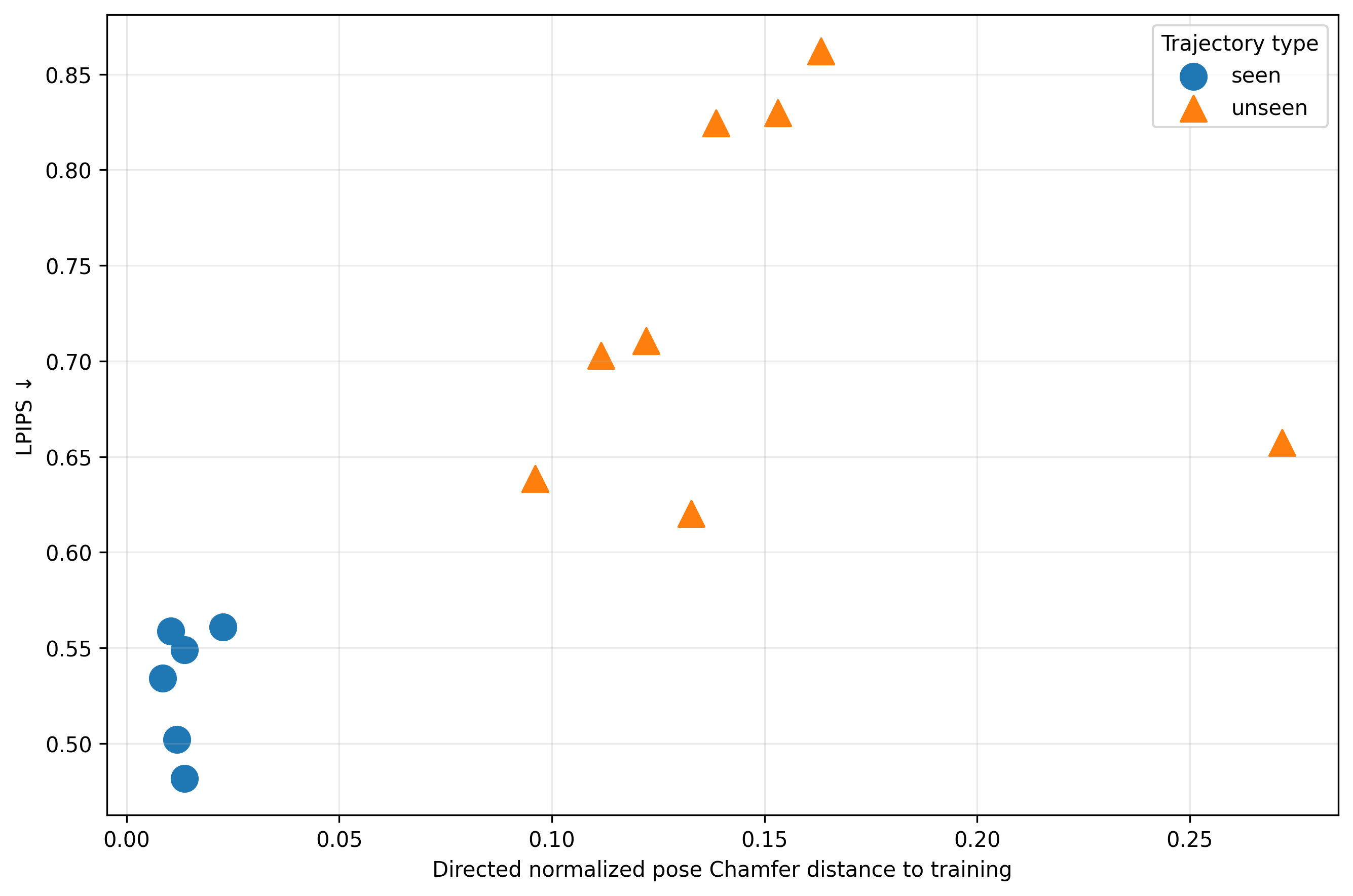} &
        \includegraphics[width=0.35\textwidth]{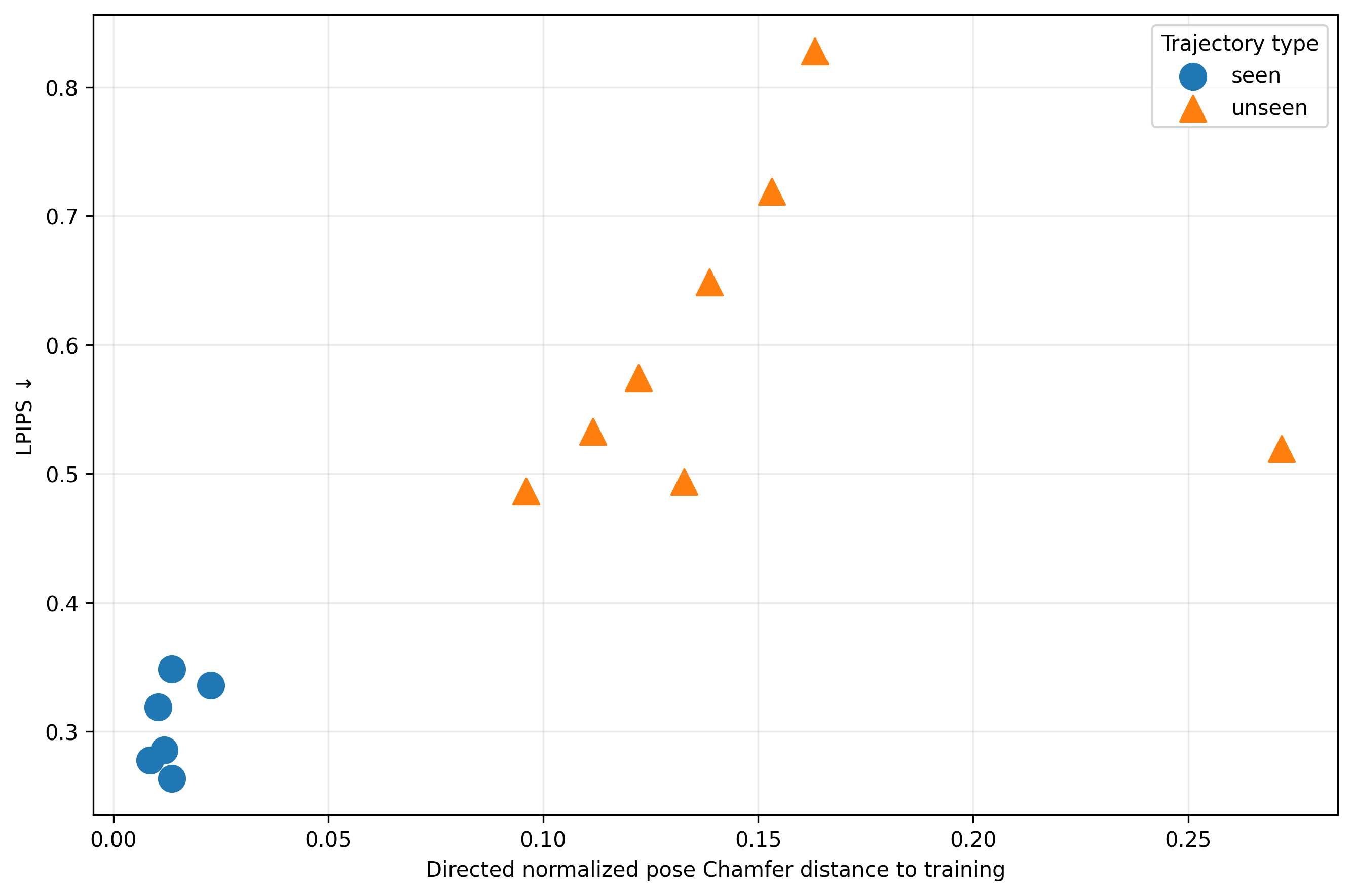} \\
        \multicolumn{2}{c}{\small (a) Church} \\[1.5mm]

        \includegraphics[width=0.35\textwidth]{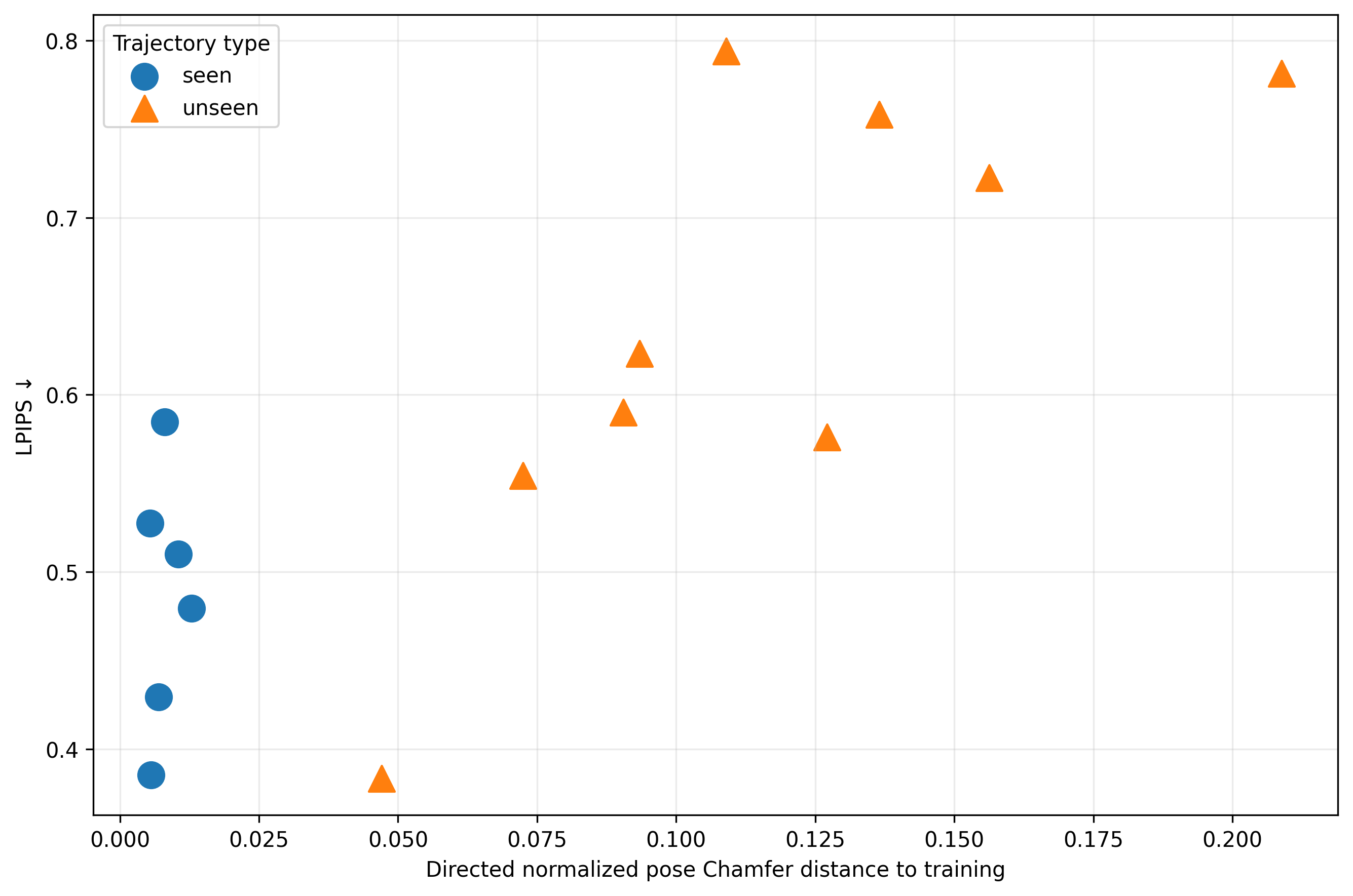} &
        \includegraphics[width=0.35\textwidth]{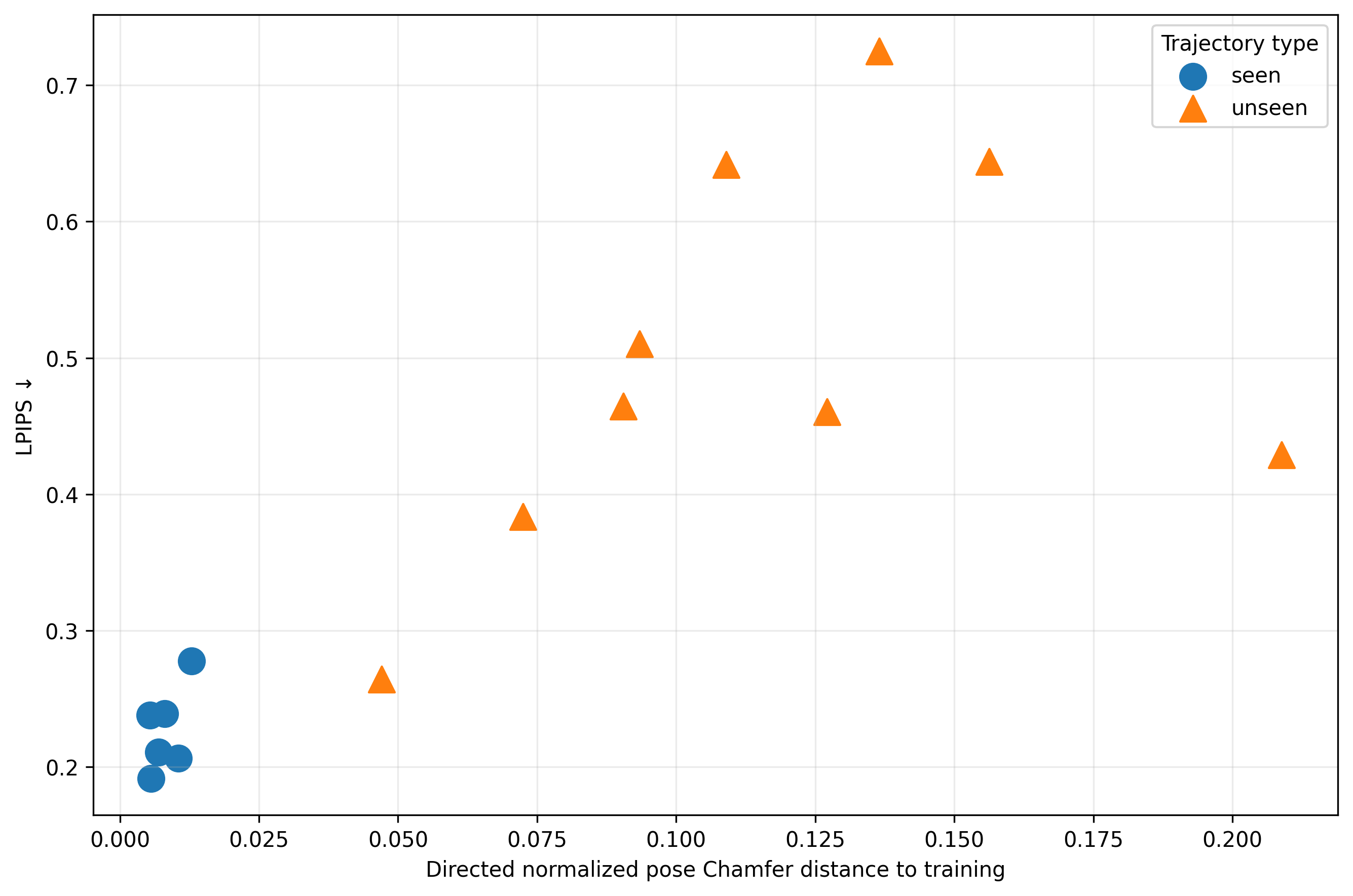} \\
        \multicolumn{2}{c}{\small (b) Village Street} \\[1.5mm]

        \includegraphics[width=0.35\textwidth]{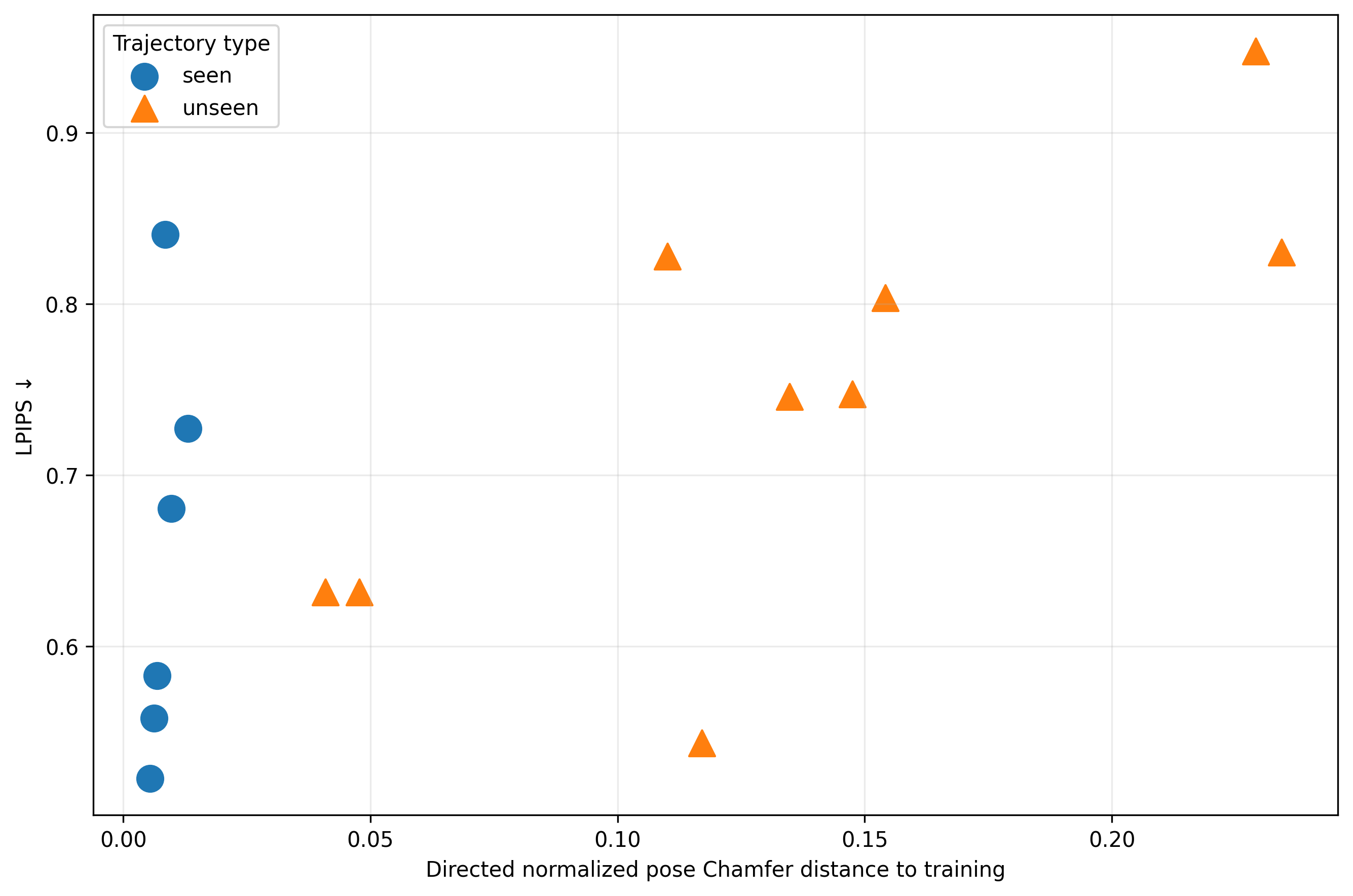} &
        \includegraphics[width=0.35\textwidth]{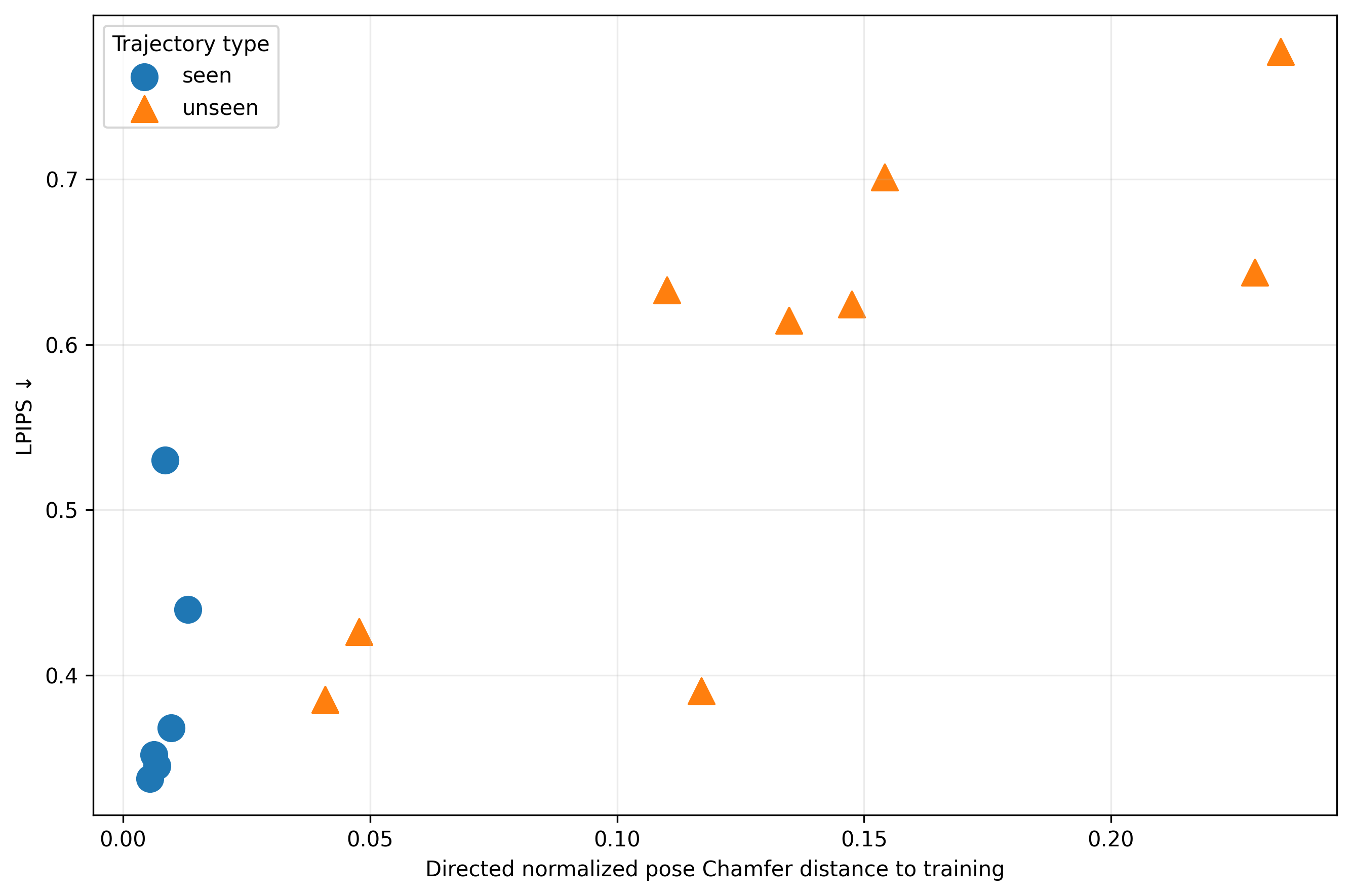} \\
        \multicolumn{2}{c}{\small (c) Victorian Garden} \\

        \includegraphics[width=0.35\textwidth]{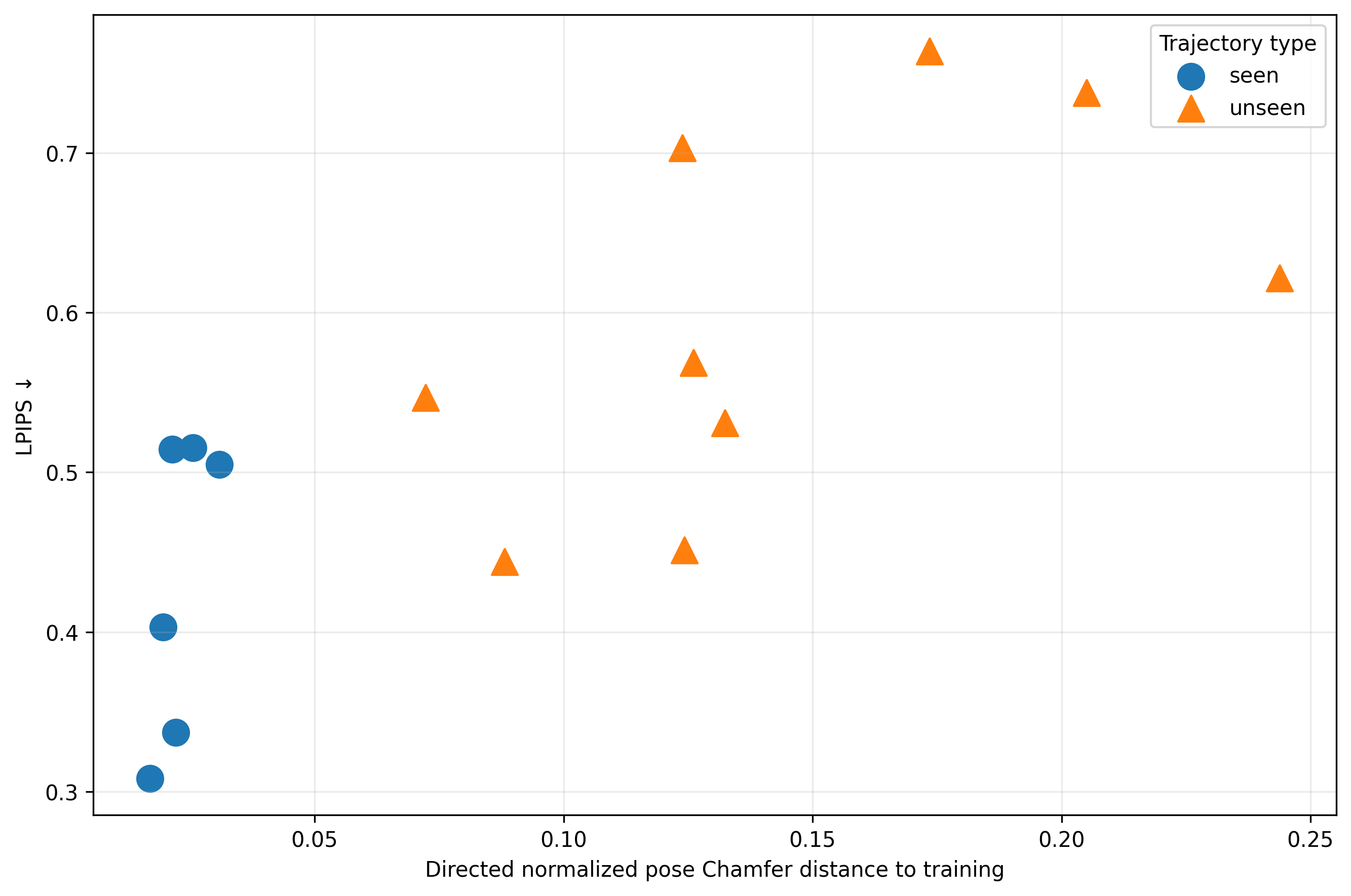} &
        \includegraphics[width=0.35\textwidth]{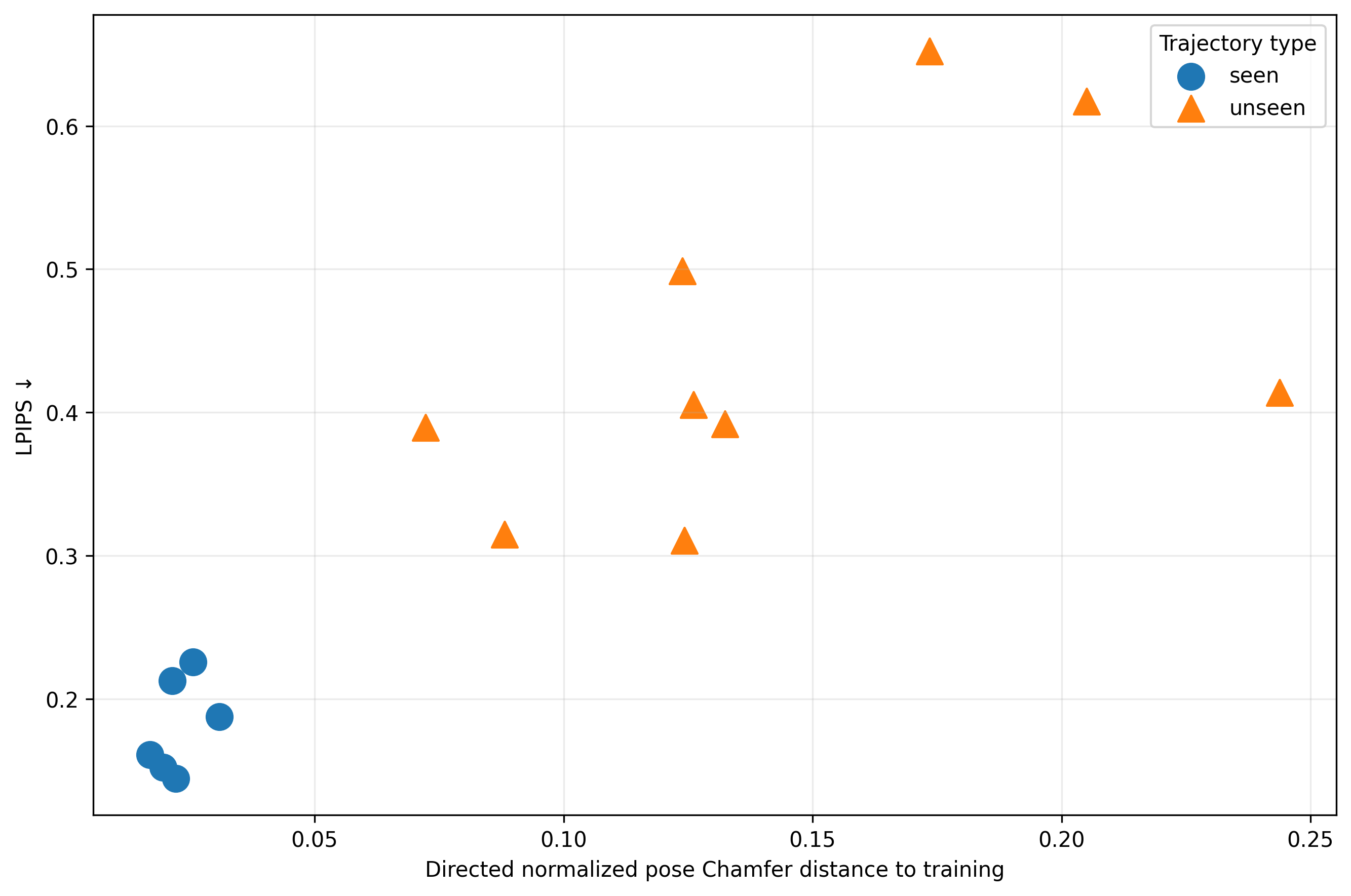} \\
        \multicolumn{2}{c}{\small (d) Frontyard} \\

        \includegraphics[width=0.35\textwidth]{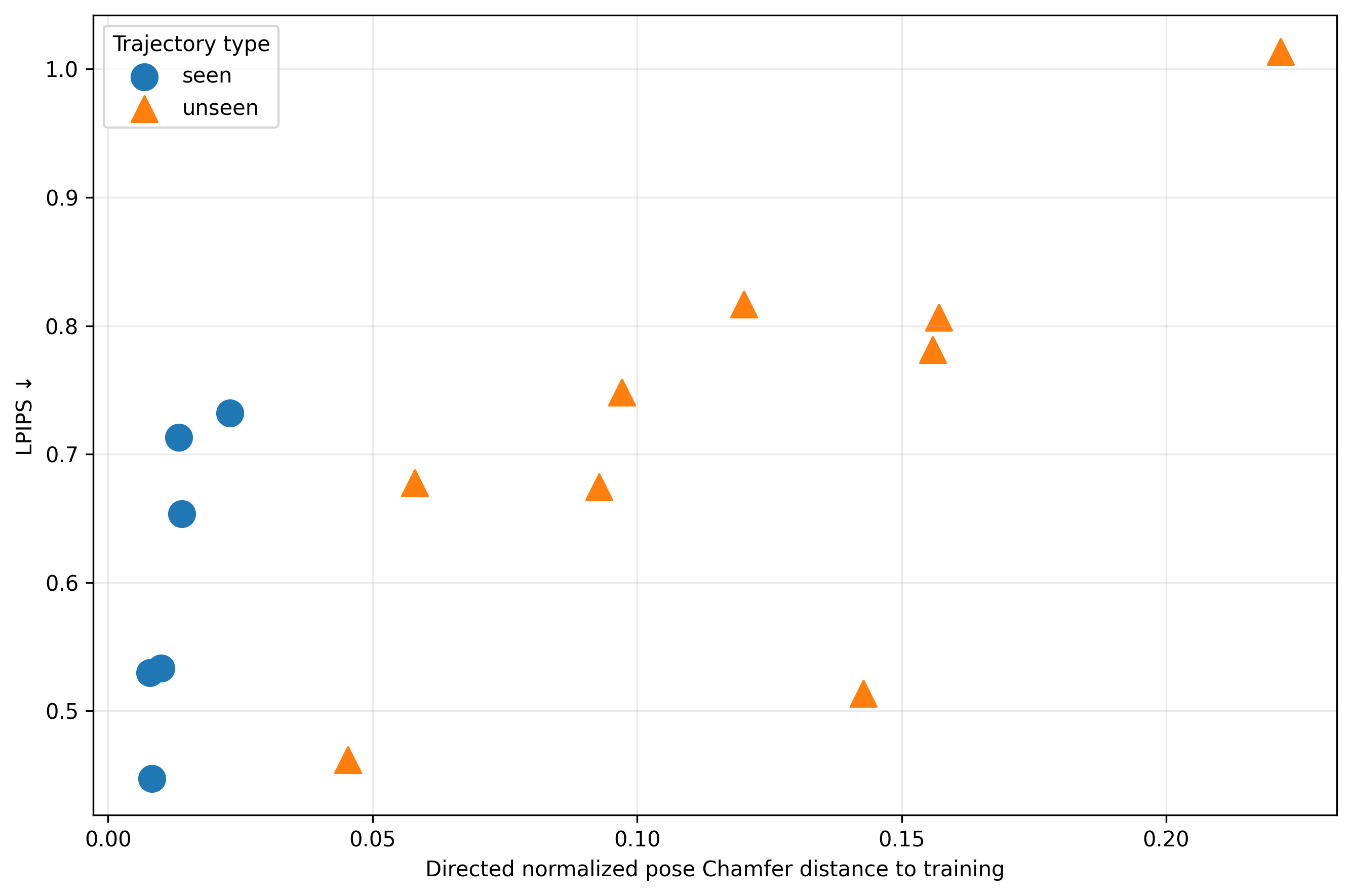} &
        \includegraphics[width=0.35\textwidth]{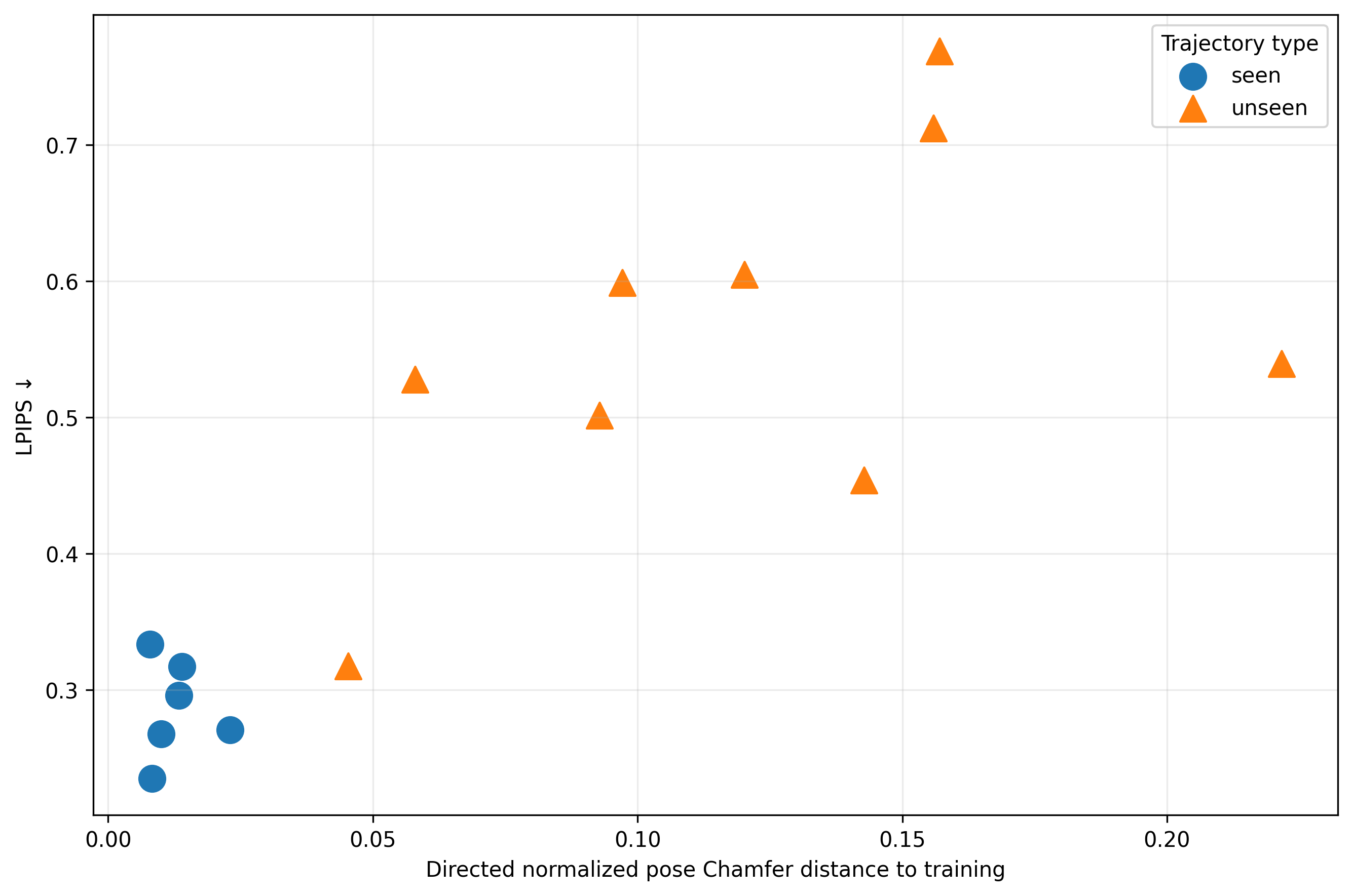} \\
        \multicolumn{2}{c}{\small (e) Backyard} \\
    \end{tabular}

    \caption{Relationship between directed normalized pose Chamfer distance to the training trajectories and reconstruction quality (LPIPS) across the PIVOT scenes. Left: Nerfacto. Right: Splatfacto. Seen trajectories (blue) correspond to held-out views from trajectories represented during training, while unseen (orange) trajectories correspond to independently captured evaluation trajectories. Lower LPIPS indicates better reconstruction quality.}
    \label{fig:chamfer_lpips_all}
\end{figure*}

Figure~\ref{fig:chamfer_lpips_all} shows the relationship between directed normalized pose Chamfer distance to the training pose set and LPIPS. Seen trajectories are held-out views from trajectories represented during training, while unseen trajectories are independently captured evaluation trajectories.

\subsection{Measured vs. Optimized Poses}

Table~\ref{tab:bm2_quantitative} reports BM2 results for the four translation/rotation source combinations. OO uses optimized translation and rotation, OM uses optimized translation and measured rotation, MO uses measured translation and optimized rotation, and MM uses measured translation and rotation.

\begin{table*}[!tbp]
\centering
\caption{BM2 quantitative results for optimized and measured camera poses.}
\label{tab:bm2_quantitative}
\small
\begin{tabular}{lllcccc}
\hline
\textbf{Scene} &
\textbf{Model} &
\textbf{Pose Type (TR)} &
\textbf{SSIM $\uparrow$} &
\textbf{LPIPS $\downarrow$} &
\textbf{PSNR $\uparrow$} &
\textbf{$\Delta$ PSNR vs. OO}\\
\hline

% ==================== Church ====================
\multirow{8}{*}{church}
& \multirow{4}{*}{Nerfacto}
& OO & 0.44 & 0.46 & 20.25 &  0.00 \\
& & OM & 0.20 & 0.79 & 13.42 & -6.82 \\
& & MO & 0.24 & 0.73 & 15.82 & -4.43 \\
& & MM & 0.20 & 0.80 & 13.55 & -6.69 \\
\cline{2-7}
& \multirow{4}{*}{Splatfacto}
& OO & 0.63 & 0.29 & 22.13 &  0.00 \\
& & OM & 0.25 & 0.60 & 14.64 & -7.48 \\
& & MO & 0.31 & 0.54 & 16.58 & -5.54 \\
& & MM & 0.25 & 0.61 & 14.56 & -7.56 \\
\hline

% ==================== Victorian Garden ====================
\multirow{8}{*}{victorian\_garden}
& \multirow{4}{*}{Nerfacto}
& OO & 0.22 & 0.60 & 16.44 &  0.00 \\
& & OM & 0.11   & 0.84   & 13.92   & -2.52   \\
& & MO & 0.12   & 0.82   & 14.82   & -1.62   \\
& & MM & 0.10 & 0.84 & 13.68 & -2.76 \\
\cline{2-7}
& \multirow{4}{*}{Splatfacto}
& OO & 0.39 & 0.36 & 17.28 &  0.00 \\
& & OM & 0.12   & 0.57   & 14.70   & -2.59   \\
& & MO & 0.15   & 0.54   & 14.92   & -2.37   \\
& & MM & 0.12 & 0.56 & 14.50 & -2.78 \\
\hline

% ==================== Village Street ====================
\multirow{8}{*}{village\_street}
& \multirow{4}{*}{Nerfacto}
& OO & 0.57 & 0.39 & 21.03 &   0.00 \\
& & OM & 0.26 & 0.75 & 14.64 &  -6.39 \\
& & MO & 0.25 & 0.76 & 14.10 &  -6.93 \\
& & MM & 0.25 & 0.77 & 14.10 &  -6.93 \\
\cline{2-7}
& \multirow{4}{*}{Splatfacto}
& OO & 0.80 & 0.17 & 24.61 &   0.00 \\
& & OM & 0.36 & 0.58 & 16.09 &  -8.52 \\
& & MO & 0.32 & 0.69 & 13.72 & -10.89 \\
& & MM & 0.30 & 0.72 & 13.05 & -11.56 \\
\hline

% ==================== Frontyard ====================
\multirow{8}{*}{frontyard}
& \multirow{4}{*}{Nerfacto}
& OO & 0.47 & 0.44 & 19.51 & 0.00 \\
& & OM & 0.15 & 0.86 & 12.98 & -6.54 \\
& & MO & 0.16 & 0.85 & 13.37 & -6.14 \\
& & MM & 0.16 & 0.87 & 13.08 & -6.43 \\
\cline{2-7}
& \multirow{4}{*}{Splatfacto}
& OO & 0.70 & 0.20 & 22.54 & 0.00 \\
& & OM & 0.17 & 0.63 & 13.22 & -9.32 \\
& & MO & 0.19 & 0.74 & 12.23 & -10.31 \\
& & MM & 0.17 & 0.74 & 11.96 & -10.58 \\
\hline

% ==================== Backyard ====================
\multirow{8}{*}{backyard}
& \multirow{4}{*}{Nerfacto}
& OO & & & & \\
& & OM & & & & \\
& & MO & & & & \\
& & MM & & & & \\
\cline{2-7}
& \multirow{4}{*}{Splatfacto}
& OO & & & & \\
& & OM & & & & \\
& & MO & & & & \\
& & MM & & & & \\
\hline

\end{tabular}

\vspace{1mm}
\footnotesize
O denotes COLMAP-optimized and M denotes measured pose components;
the first and second letters correspond to translation (T) and rotation (R),
respectively.
\end{table*}

\subsection{Calibrated vs. Optimized Intrinsics}

Table~\ref{tab:bm3_quantitative} compares scene-optimized COLMAP intrinsics with the fixed OpenCV calibration used by the capture device. Pose source is held fixed at the COLMAP optimized pose while the intrinsic source changes.

\begin{table*}[!tbp]
\centering
\caption{BM3 quantitative results comparing COLMAP-optimized and OpenCV-calibrated camera intrinsics.}
\label{tab:bm3_quantitative}
\small
\begin{tabular}{lllcccc}
\hline
\textbf{Scene} &
\textbf{Model} &
\textbf{Camera Calibration Type} &
\textbf{SSIM $\uparrow$} &
\textbf{LPIPS $\downarrow$} &
\textbf{PSNR $\uparrow$} &
\textbf{$\Delta$ PSNR vs. Optimized} \\
\hline

% ==================== Church ====================
\multirow{4}{*}{church}
& \multirow{2}{*}{Nerfacto}
& COLMAP optimized   & 0.44 & 0.45 & 20.26 & 0.00 \\
& & OpenCV calibrated & 0.24 & 0.74 & 15.78 & -4.47 \\
\cline{2-7}
& \multirow{2}{*}{Splatfacto}
& COLMAP optimized   & 0.63 & 0.29 & 22.13 & 0.00 \\
& & OpenCV calibrated & 0.29 & 0.54 & 16.42 & -5.71 \\
\hline

% ==================== Victorian Garden ====================
\multirow{4}{*}{victorian\_garden}
& \multirow{2}{*}{Nerfacto}
& COLMAP optimized   & 0.22 & 0.60 & 16.46 & 0.00 \\
& & OpenCV calibrated & 0.11 & 0.84 & 14.71 & -1.74 \\
\cline{2-7}
& \multirow{2}{*}{Splatfacto}
& COLMAP optimized   & 0.39 & 0.36 & 17.29 & 0.00 \\
& & OpenCV calibrated & 0.12 & 0.55 & 14.69 & -2.59 \\
\hline

% ==================== Village Street ====================
\multirow{4}{*}{village\_street}
& \multirow{2}{*}{Nerfacto}
& COLMAP optimized   & 0.57 & 0.39 & 21.00 & 0.00 \\
& & OpenCV calibrated & 0.25 & 0.75 & 14.35 & -6.65 \\
\cline{2-7}
& \multirow{2}{*}{Splatfacto}
& COLMAP optimized   & 0.80 & 0.17 & 24.63 & 0.00 \\
& & OpenCV calibrated & 0.35 & 0.61 & 15.30 & -9.33 \\
\hline

% ==================== Frontyard ====================
\multirow{4}{*}{frontyard}
& \multirow{2}{*}{Nerfacto}
& COLMAP optimized    & 0.46 & 0.45 & 19.50 & 0.00 \\
& & OpenCV calibrated & 0.17 & 0.84 & 14.78 & -4.72 \\
\cline{2-7}
& \multirow{2}{*}{Splatfacto}
& COLMAP optimized    & 0.70 & 0.20 & 22.56 & 0.00 \\
& & OpenCV calibrated & 0.21 & 0.54 & 14.95 & -7.61 \\
\hline

% ==================== Backyard ====================
\multirow{4}{*}{backyard}
& \multirow{2}{*}{Nerfacto}
& COLMAP optimized   & & & & \\
& & OpenCV calibrated & & & & \\
\cline{2-7}
& \multirow{2}{*}{Splatfacto}
& COLMAP optimized   & & & & \\
& & OpenCV calibrated & & & & \\
\hline

\end{tabular}
\end{table*}

\section{Qualitative Results}
\label{sec:qual}

The following qualitative comparisons use representative views from two scenes. Each panel uses the same ground-truth view across compared reconstruction conditions so that changes in rendering quality can be inspected directly.

\subsection{BM1: Seen vs. Unseen Trajectories}

\begin{figure*}[!tbp]
    \centering
    \setlength{\tabcolsep}{3pt}
    \renewcommand{\arraystretch}{1.05}

    \begin{tabular}{c c c c}
        \textbf{Scene} &
        \textbf{Model} &
        \textbf{Seen: orbit\_inward\_high} &
        \textbf{Unseen: scattered\_low} \\
        \hline

        % ================= Village Street =================
        \multirow{3}{*}{\shortstack{Village\\Street}}
        & GT
        & \includegraphics[width=0.28\textwidth]{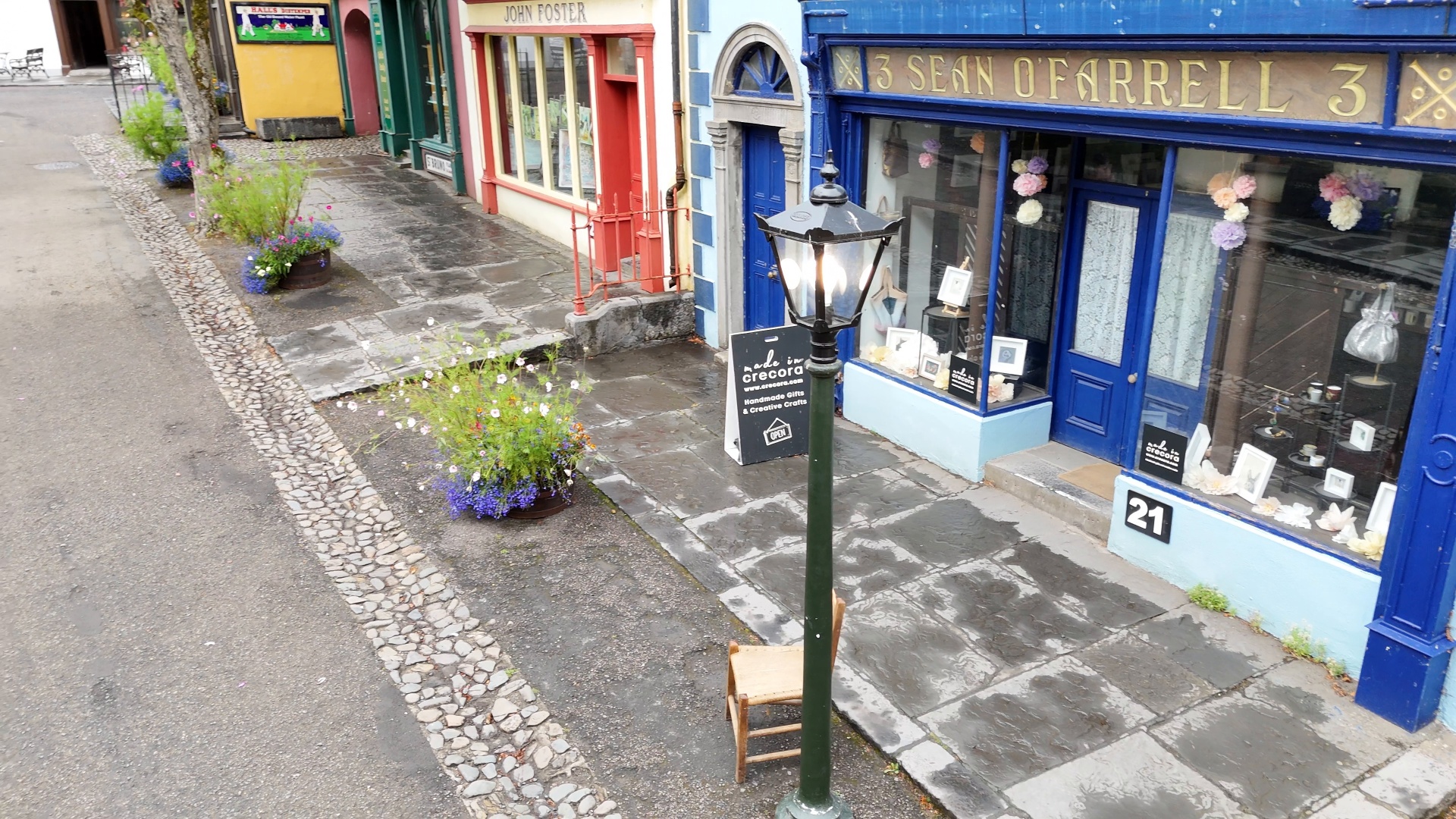}
        & \includegraphics[width=0.28\textwidth]{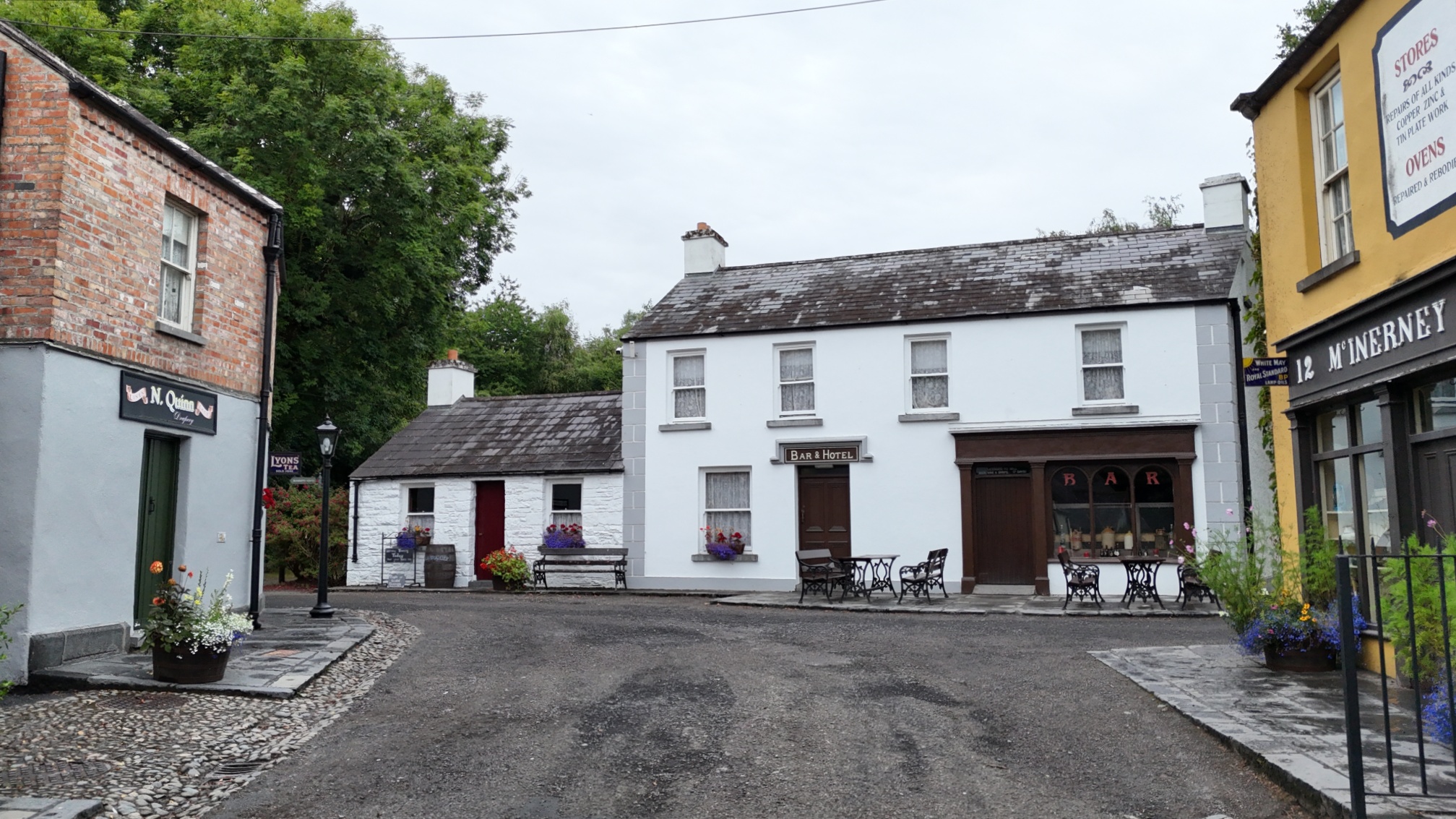} \\

        & Nerfacto
        & \includegraphics[width=0.28\textwidth]{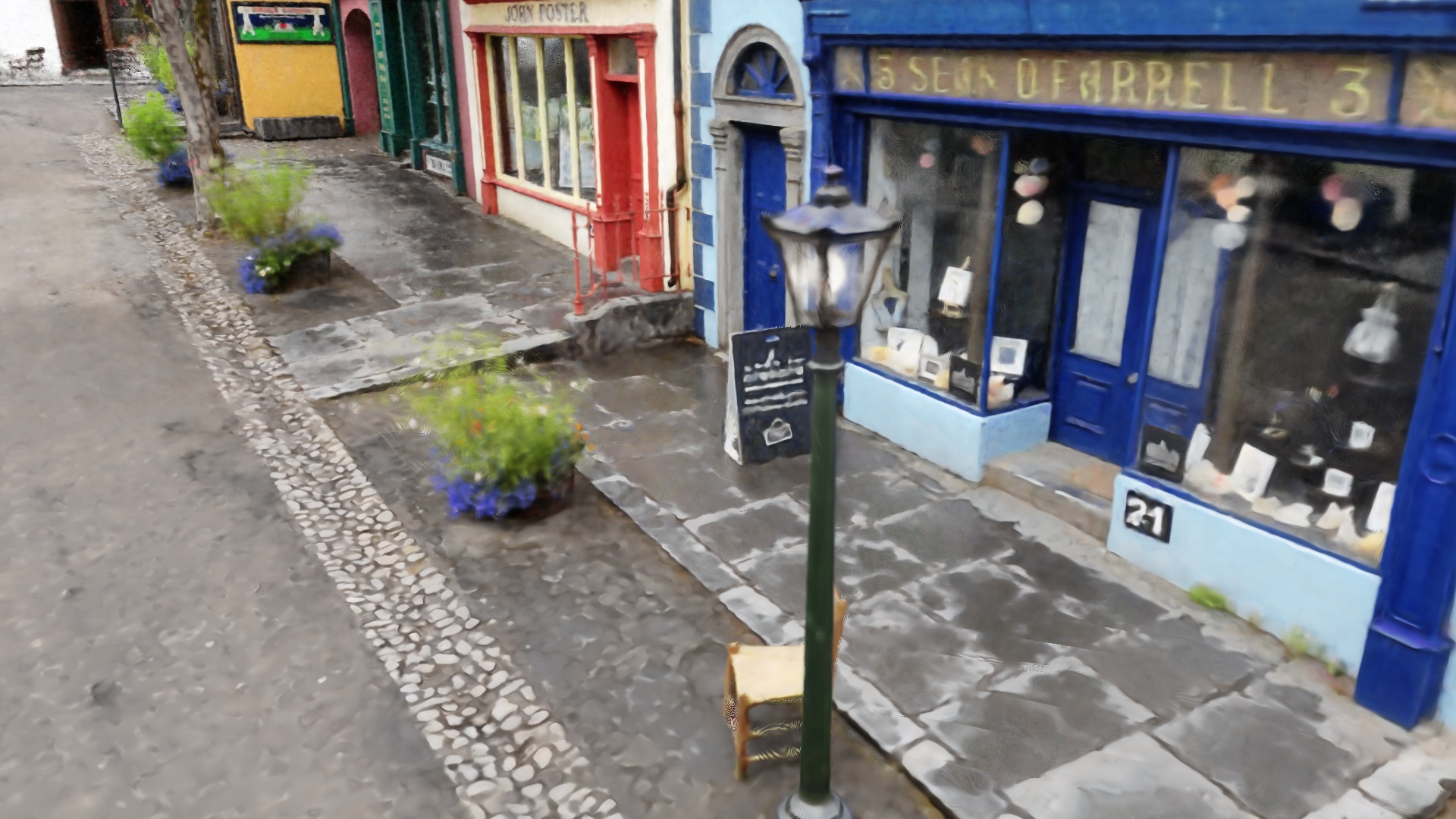}
        & \includegraphics[width=0.28\textwidth]{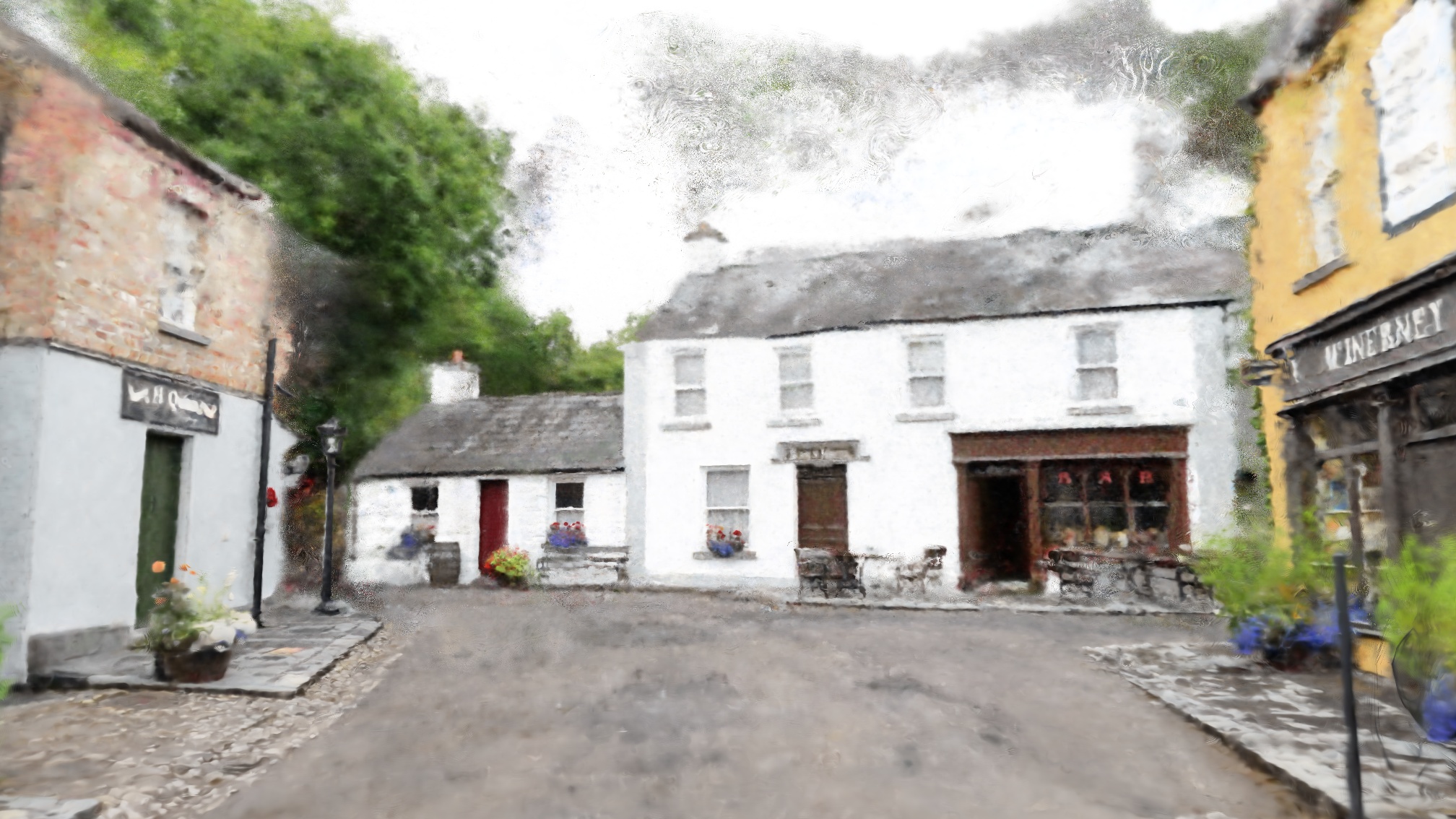} \\

        & Splatfacto
        & \includegraphics[width=0.28\textwidth]{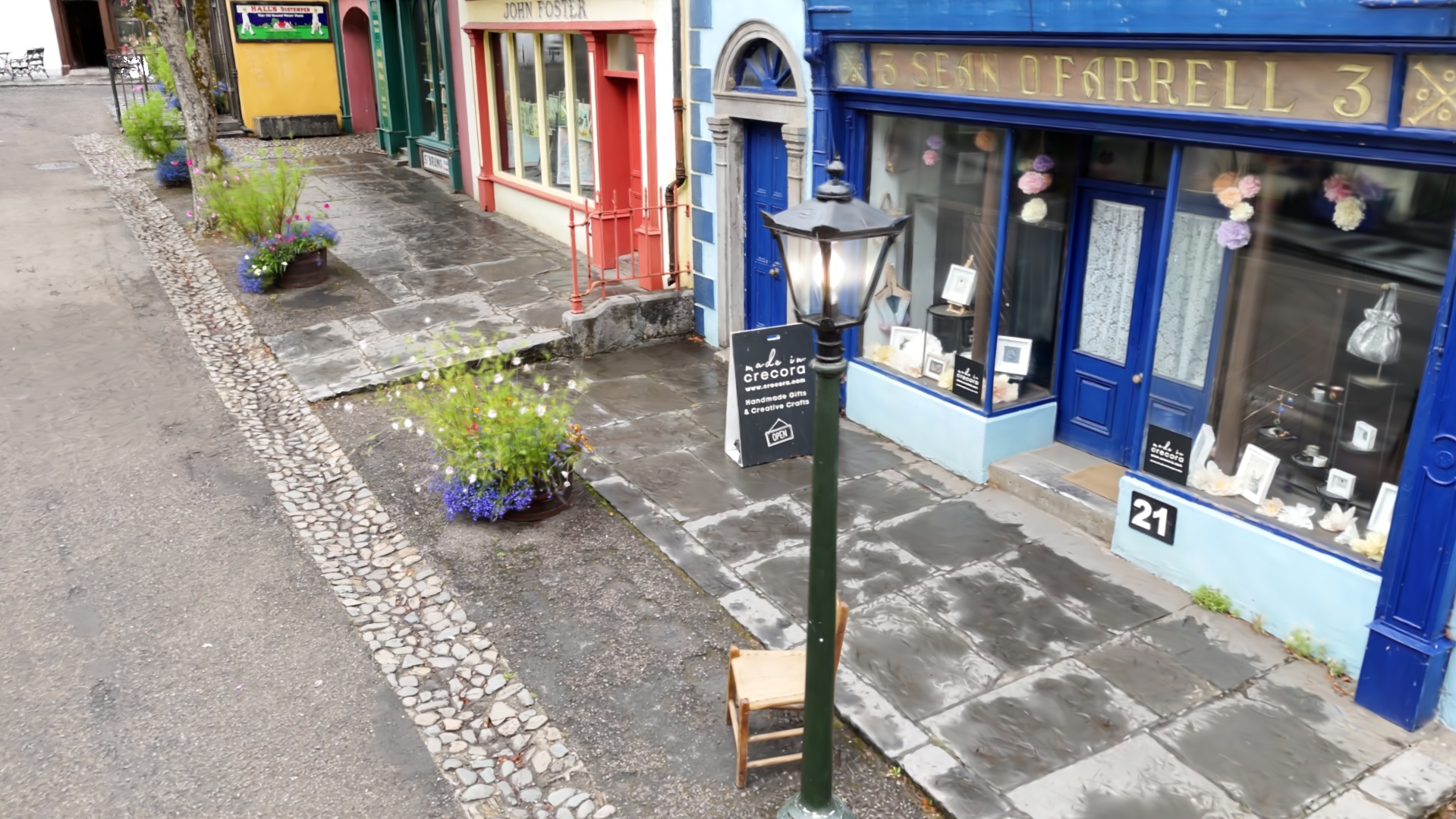}
        & \includegraphics[width=0.28\textwidth]{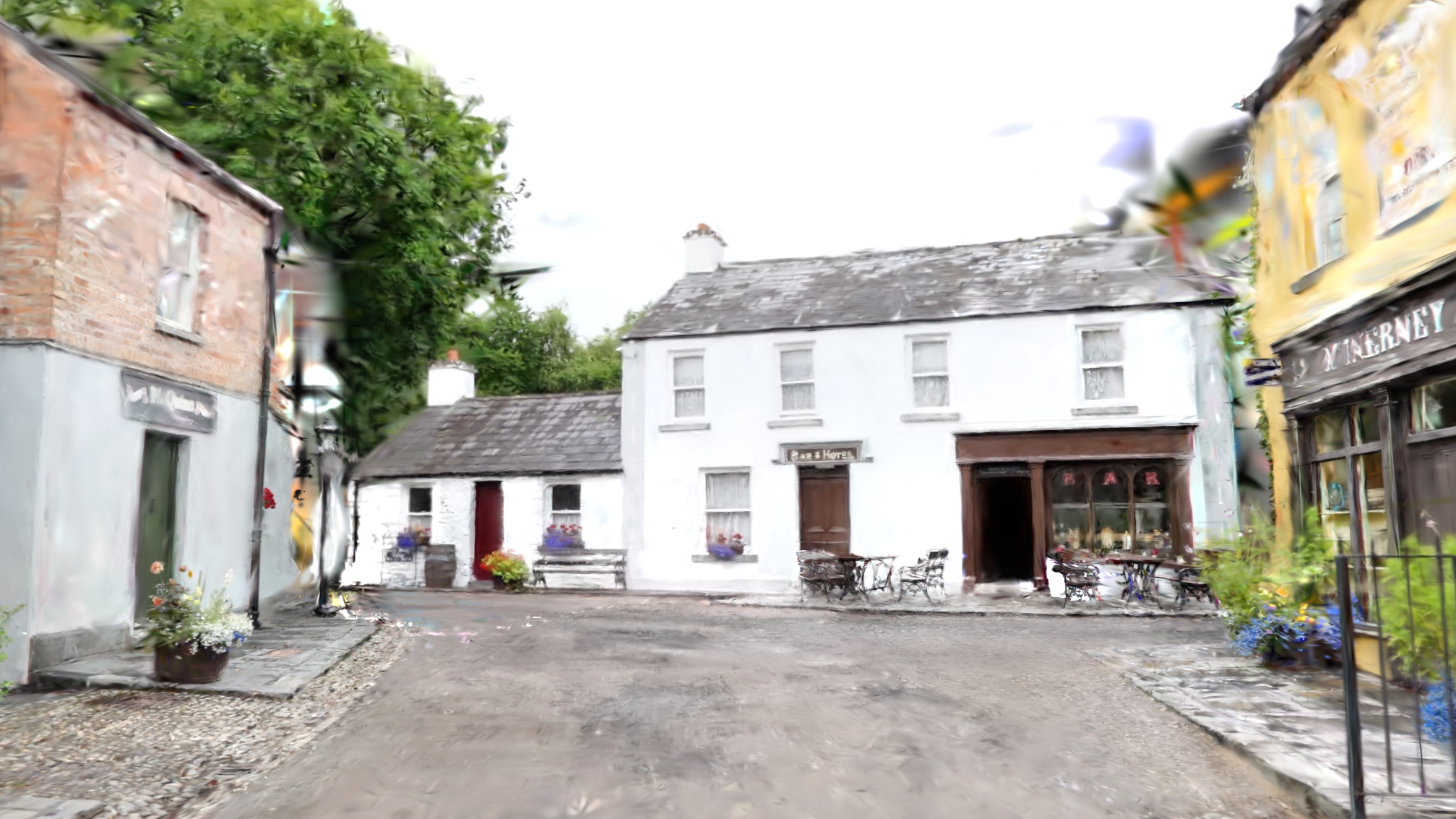} \\

        \hline

        % ================= Church =================
        \multirow{3}{*}{Church}
        & GT
        & \includegraphics[width=0.28\textwidth]{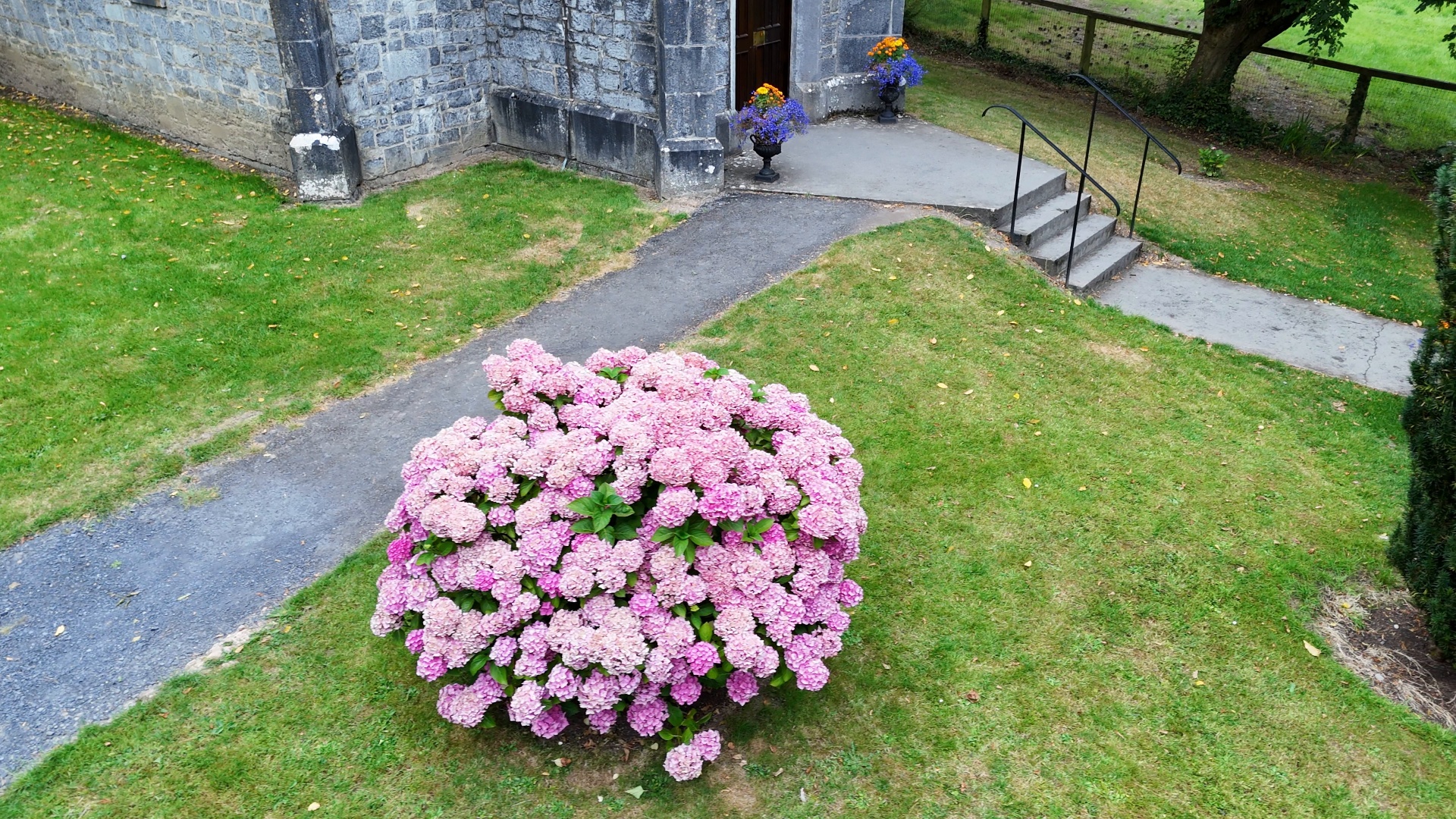}
        & \includegraphics[width=0.28\textwidth]{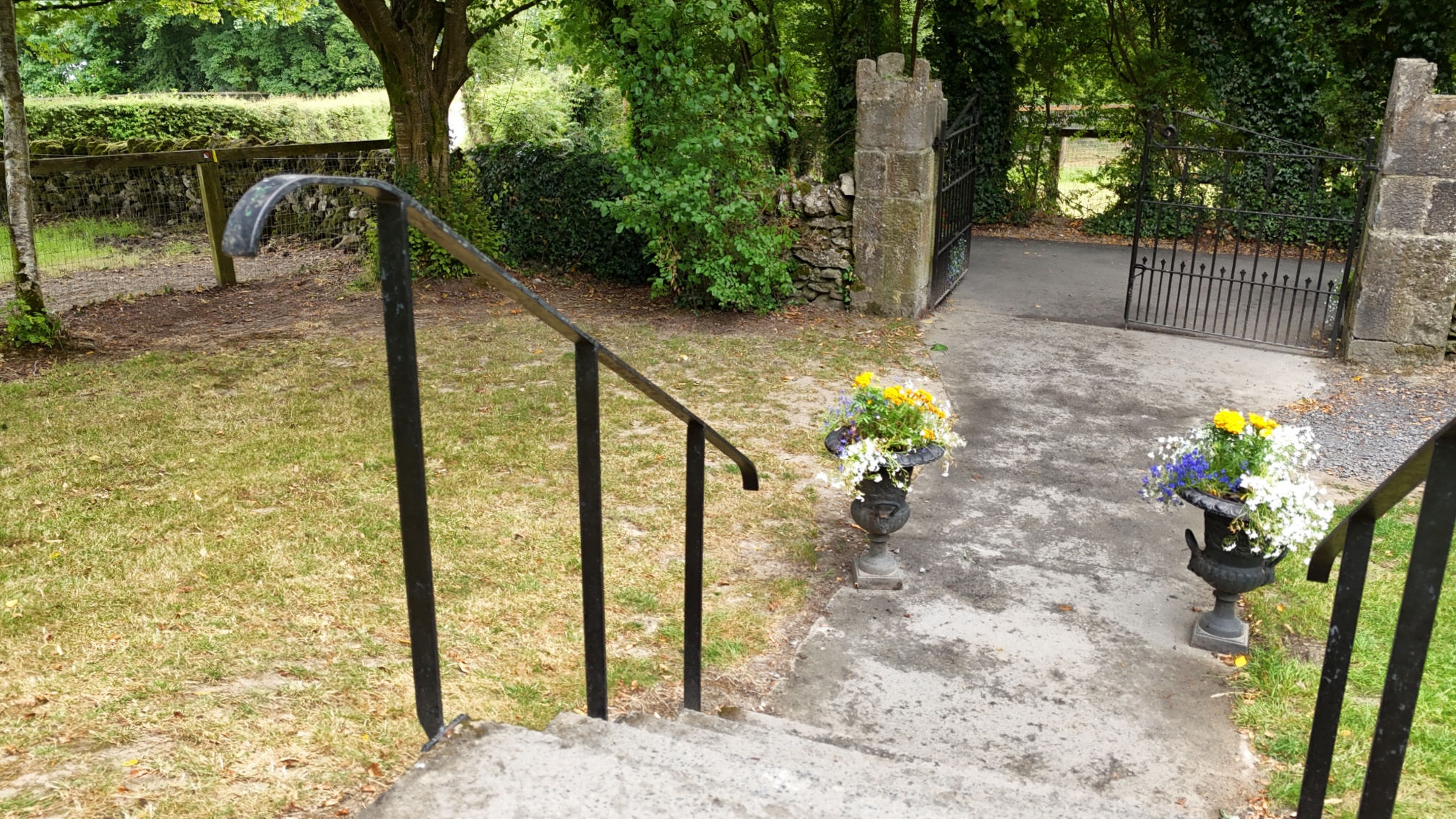} \\

        & Nerfacto
        & \includegraphics[width=0.28\textwidth]{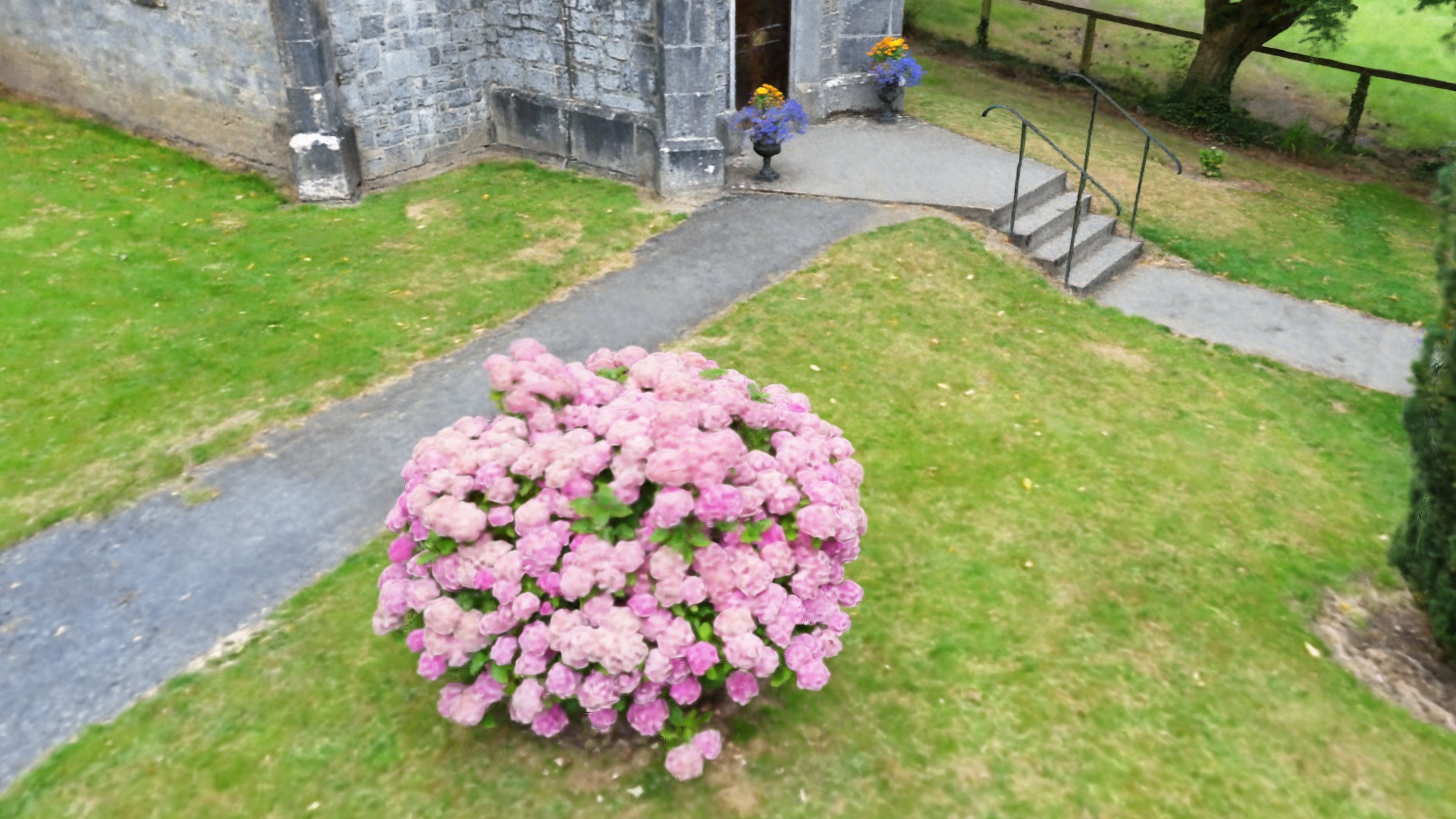}
        & \includegraphics[width=0.28\textwidth]{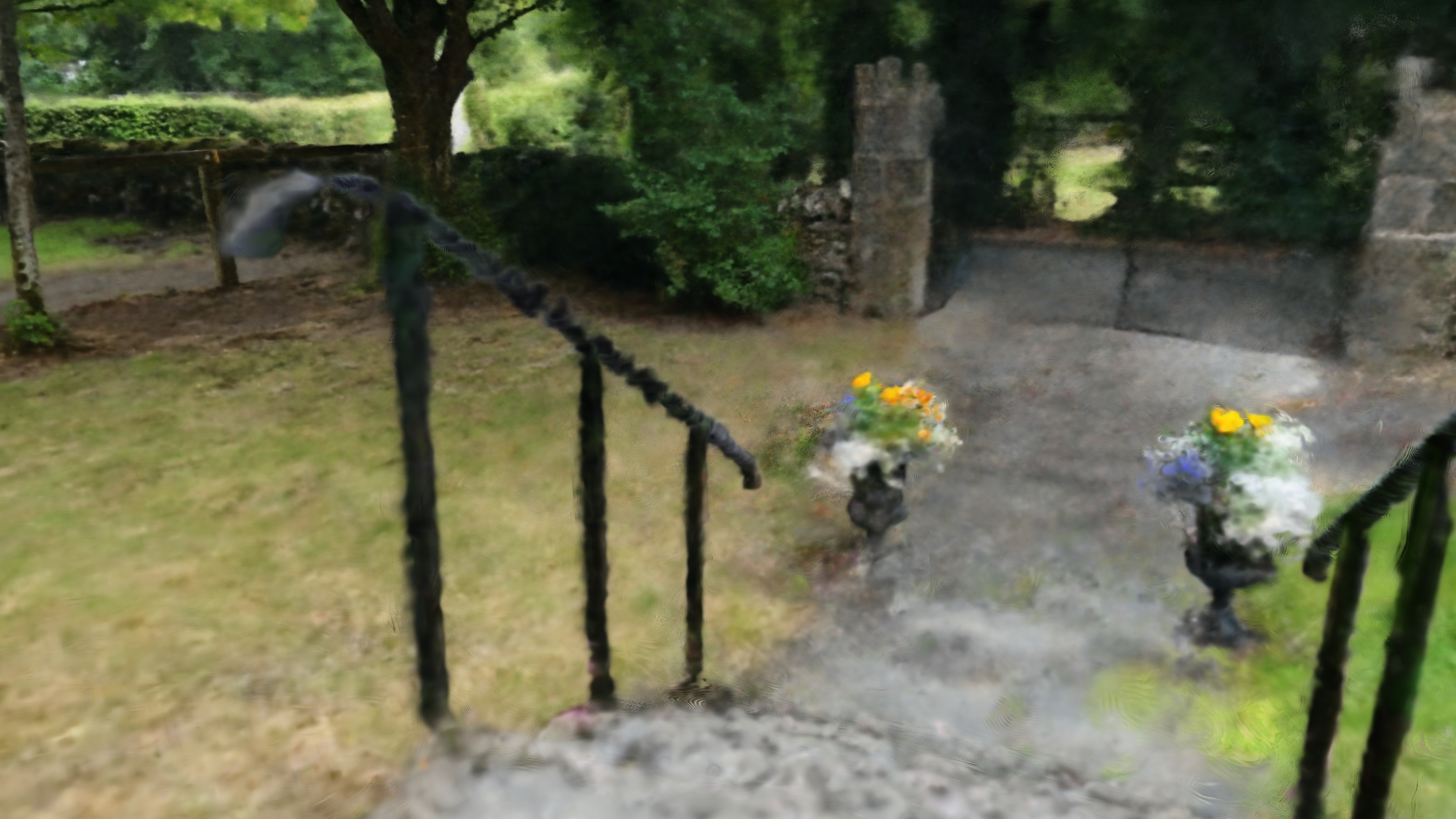} \\

        & Splatfacto
        & \includegraphics[width=0.28\textwidth]{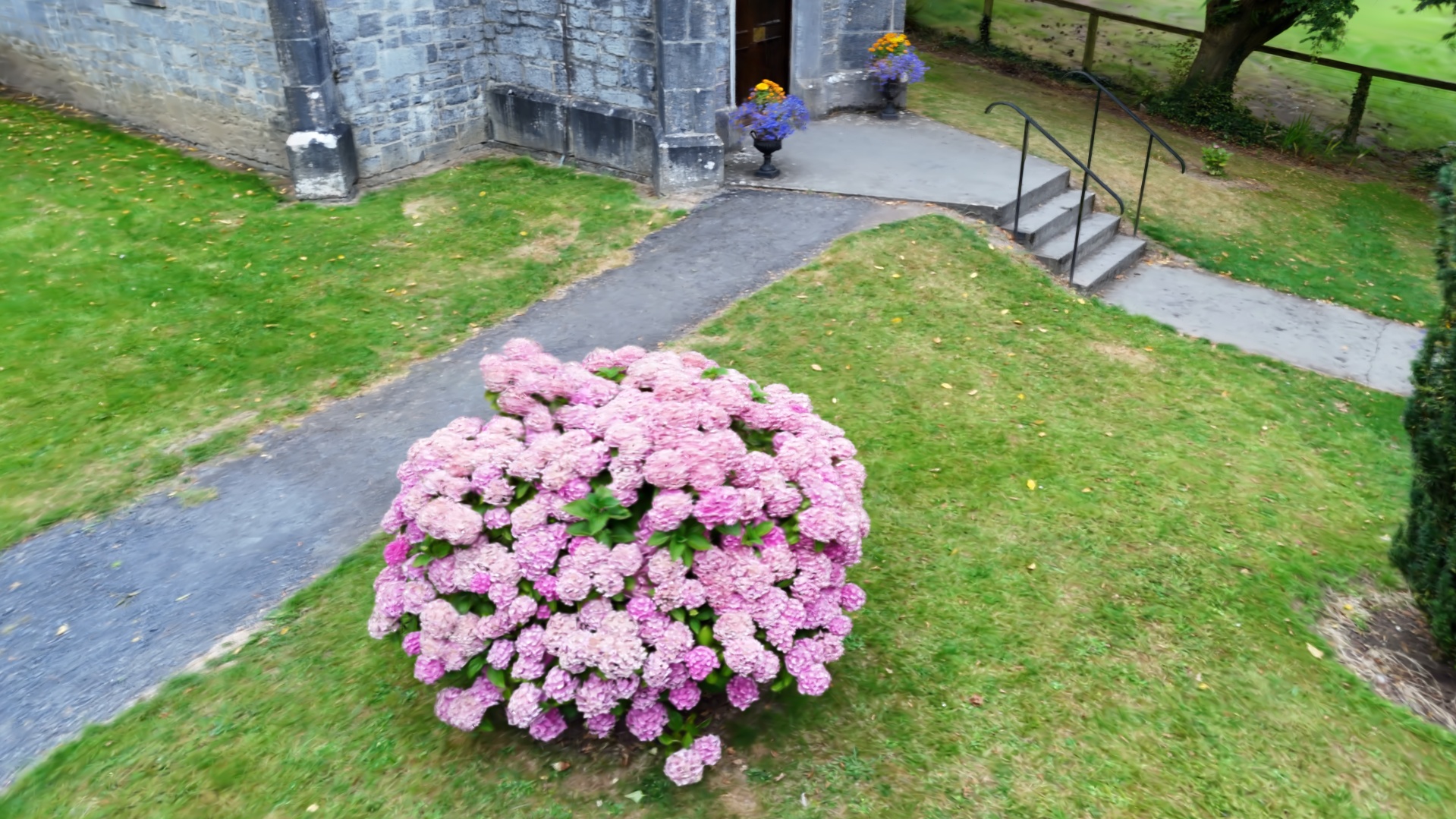}
        & \includegraphics[width=0.28\textwidth]{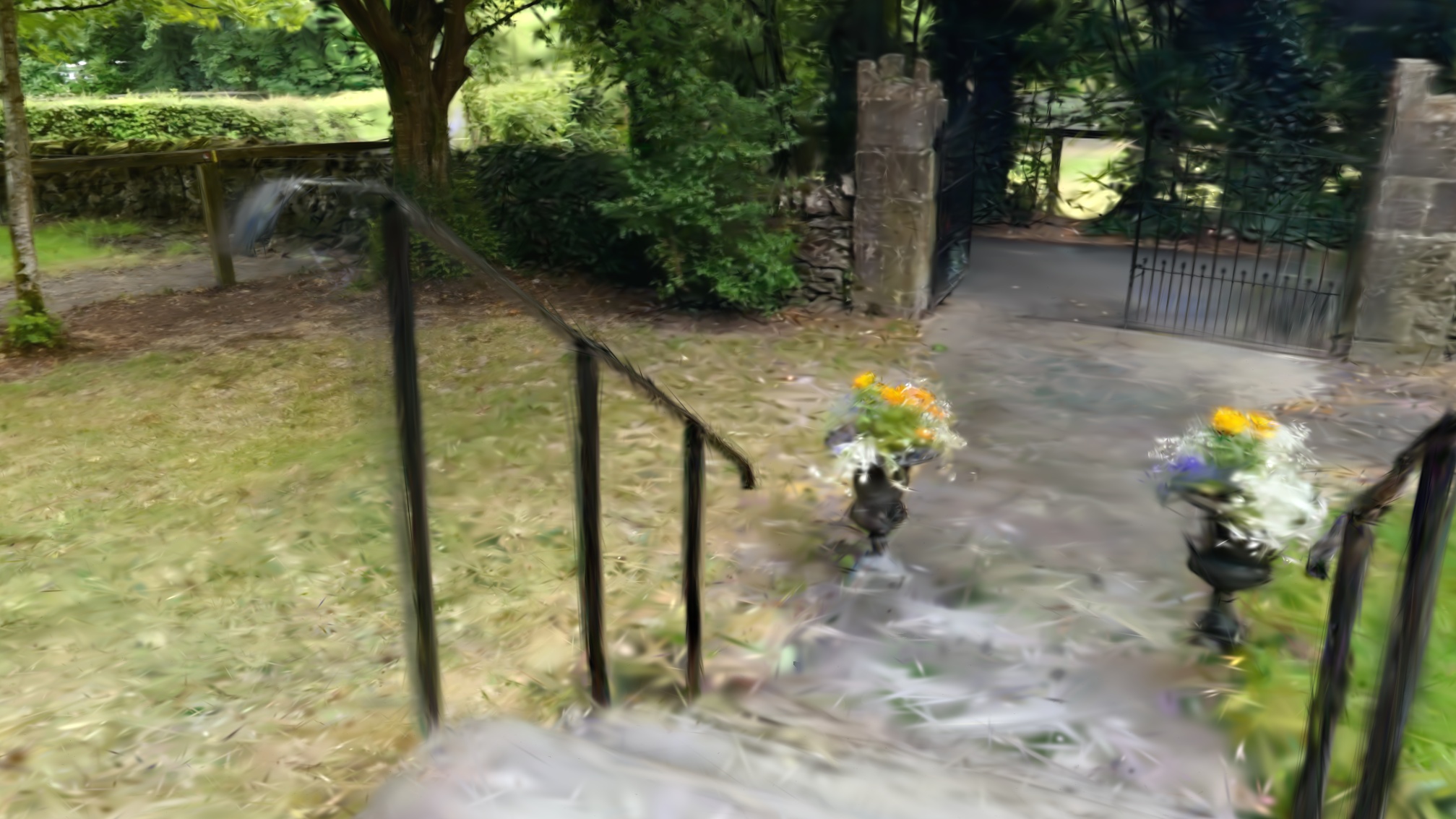} \\

    \end{tabular}

    \caption{
    Qualitative BM1: Seen vs unseen reconstruction results for representative seen and unseen
    evaluation trajectories. Seen examples are sampled from a training
    trajectory, while unseen examples are sampled from an independently
    captured trajectory. Ground-truth images are shown alongside Nerfacto and
    Splatfacto reconstructions.
    }

    \label{fig:bm1_qualitative}
\end{figure*}

Figure~\ref{fig:bm1_qualitative} compares representative seen and unseen views for Nerfacto and Splatfacto.

\subsection{BM2: Measured vs. Optimized Poses}

\begin{figure*}[!tbp]
    \centering
    \setlength{\tabcolsep}{4pt}
    \renewcommand{\arraystretch}{1.05}

    \begin{tabular}{c c c c}
        \textbf{Scene} &
        \textbf{Pose Type} &
        \textbf{Nerfacto} &
        \textbf{Splatfacto} \\
        \hline

        % ================= Village Street =================
        \multirow{5}{*}{\shortstack{Village\\Street}}

        & GT
        & \includegraphics[height=1.95cm]{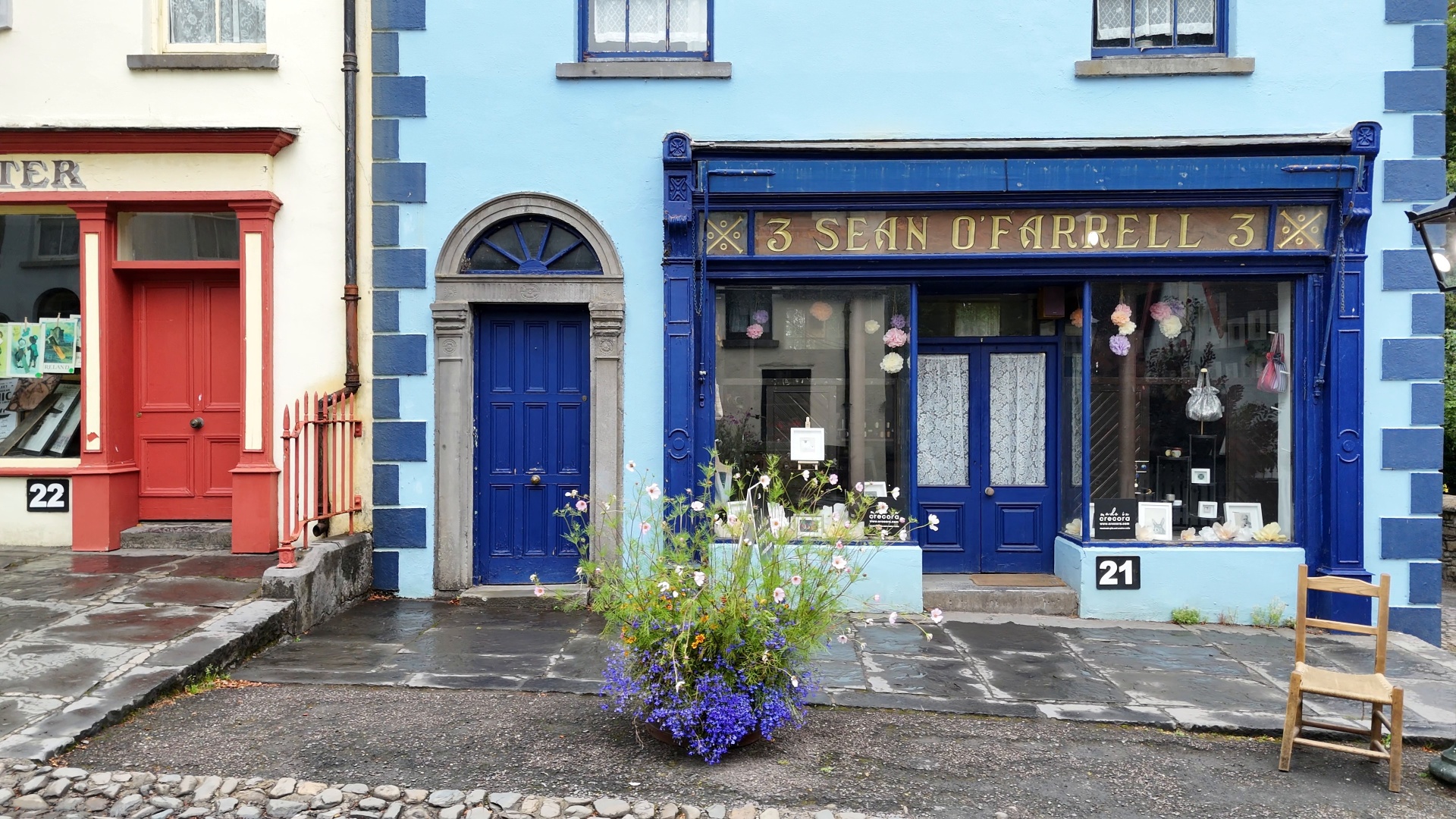}
        & \includegraphics[height=1.95cm]{fig/bm2/gt_eval_orbit_inward_low_frame_000562.JPG} \\

        & OO
        & \includegraphics[height=1.95cm]{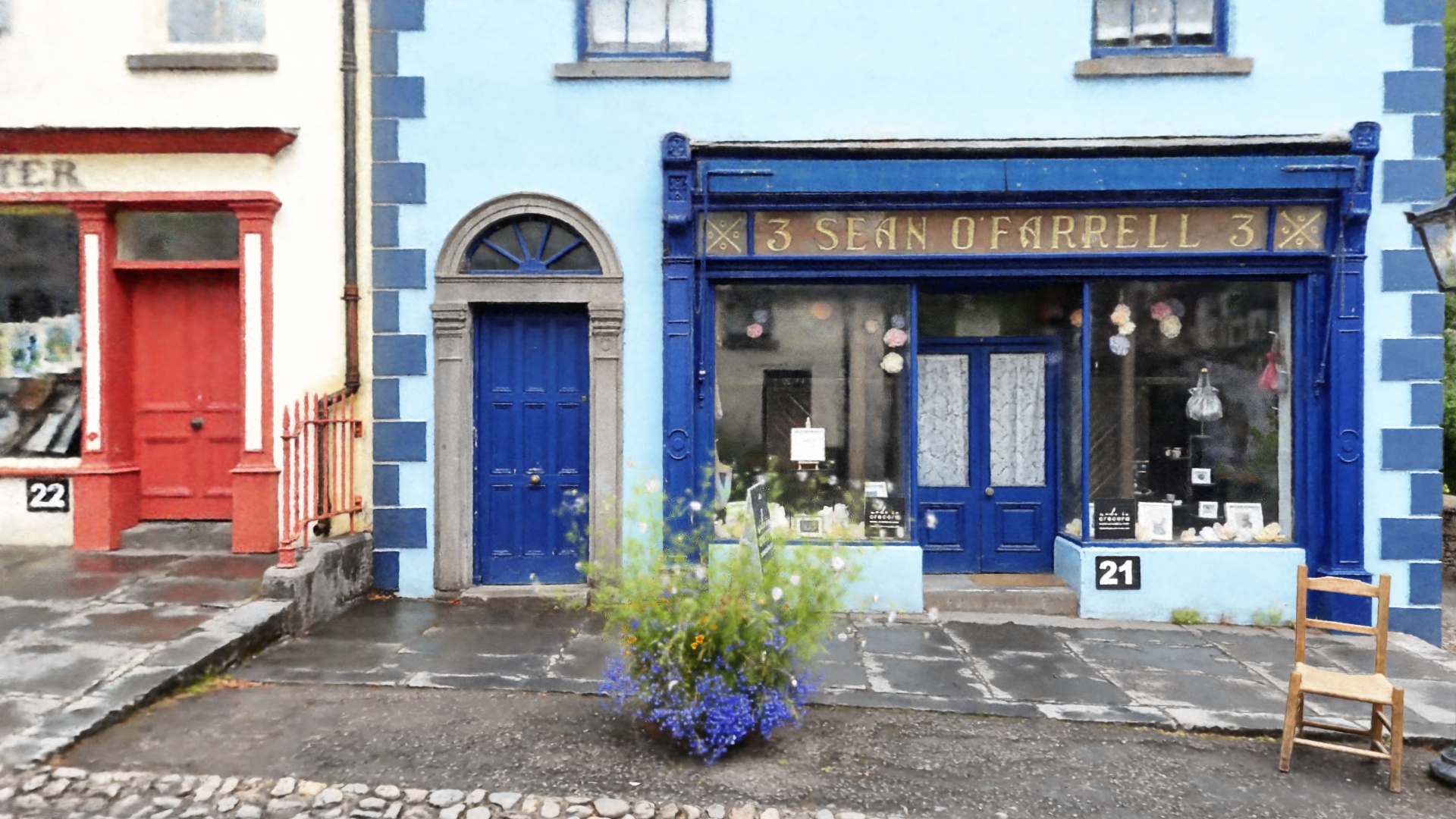}
        & \includegraphics[height=1.95cm]{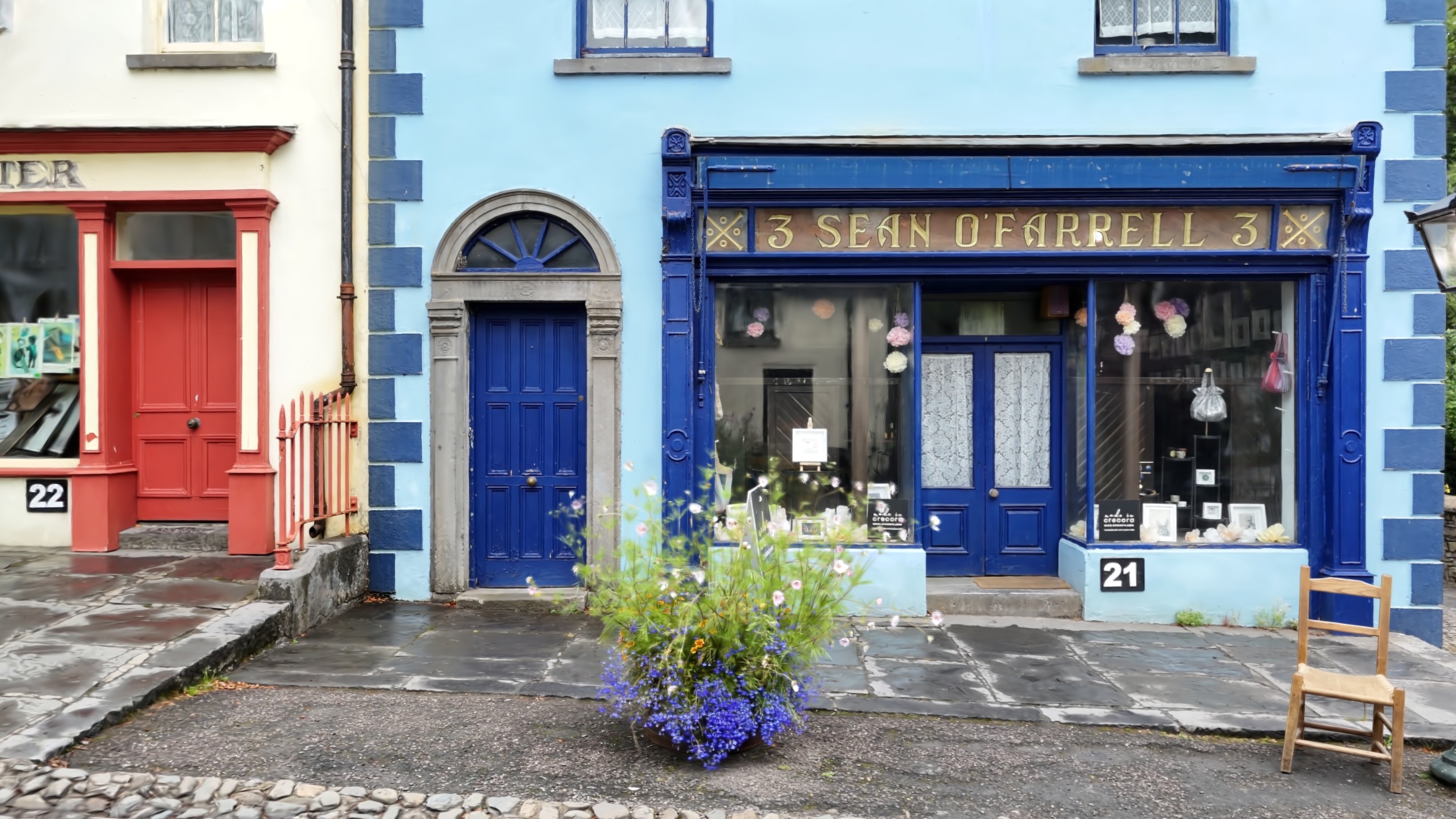} \\

        & MO
        & \includegraphics[height=1.95cm]{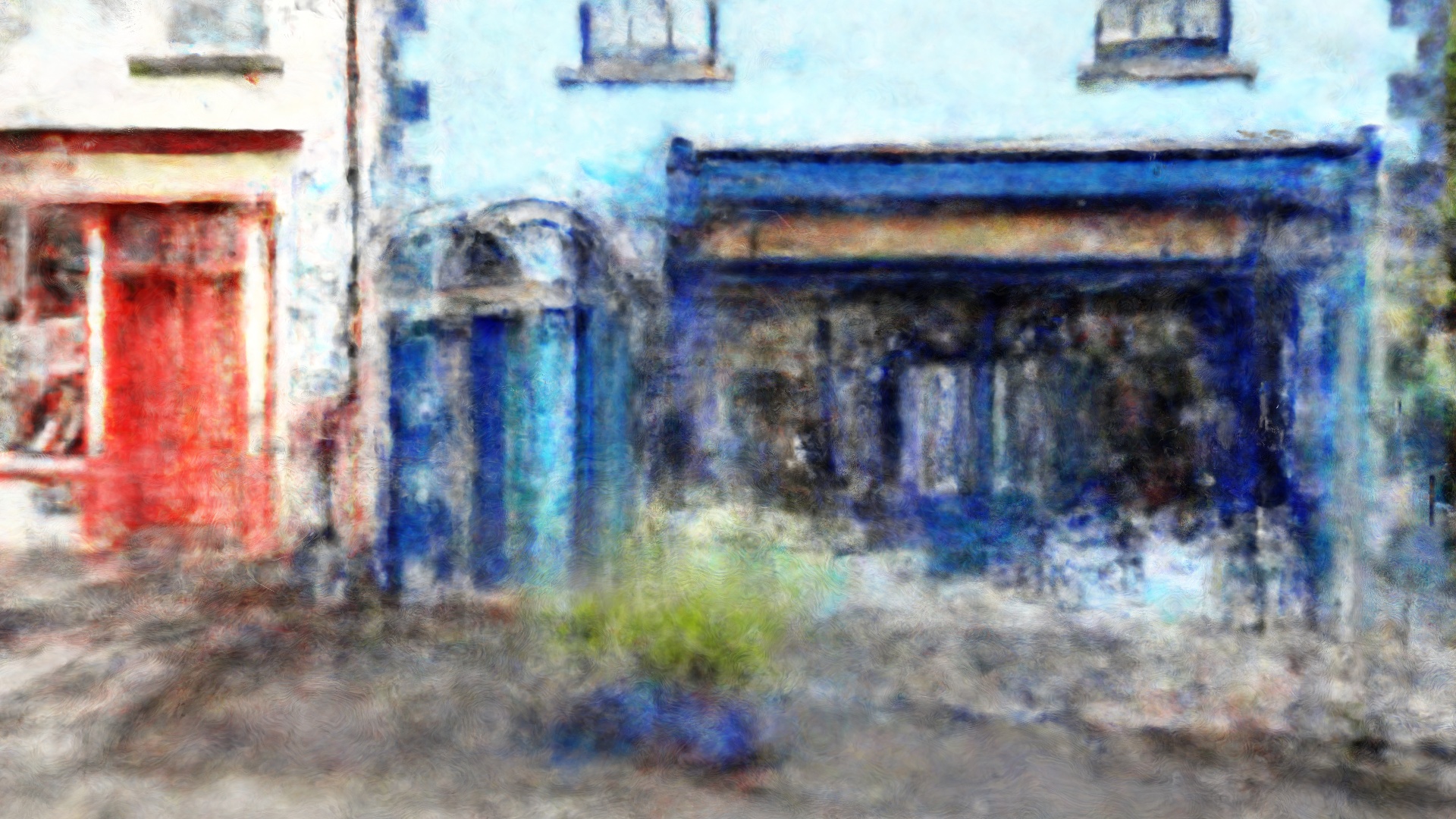}
        & \includegraphics[height=1.95cm]{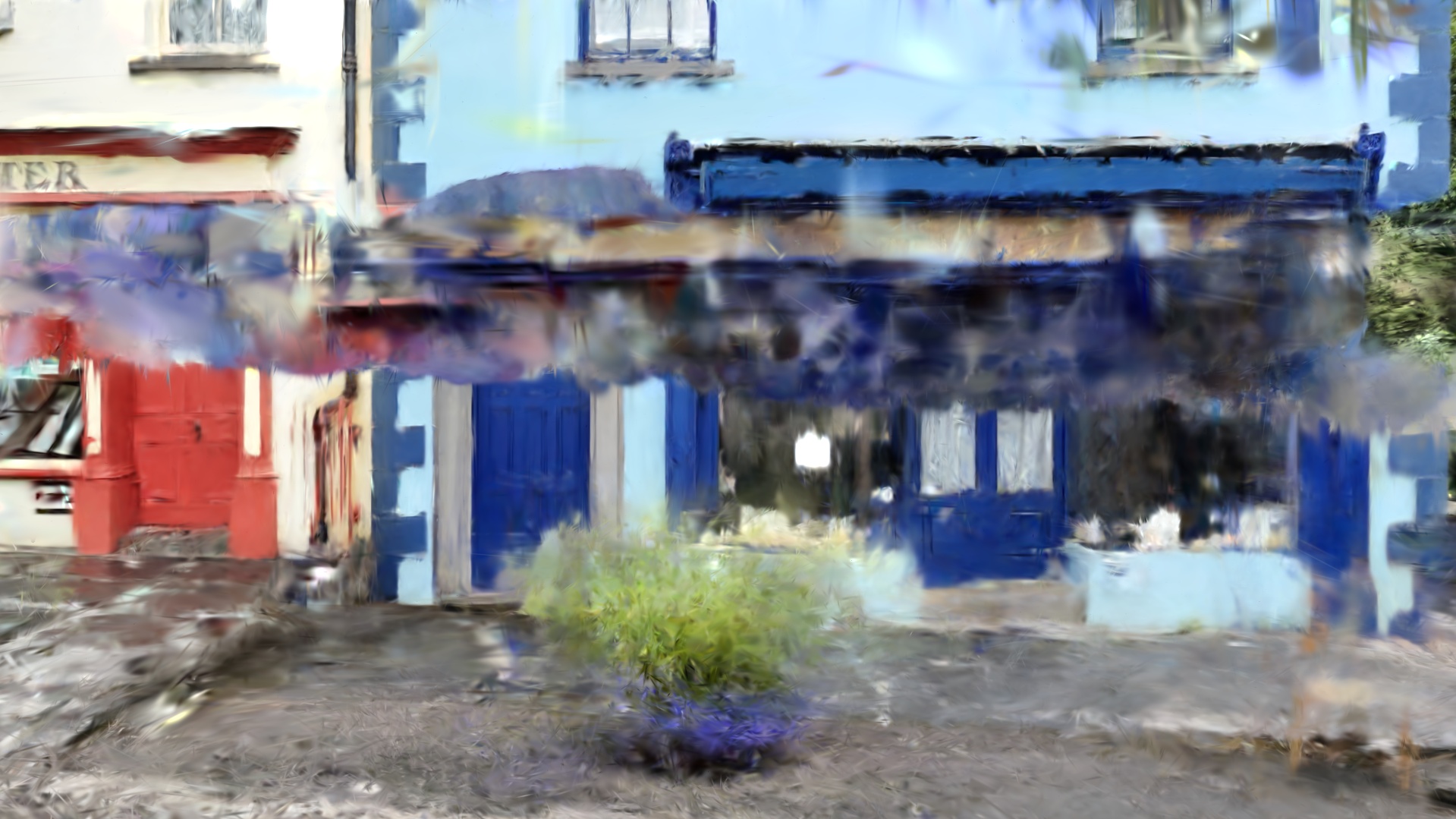} \\

        & OM
        & \includegraphics[height=1.95cm]{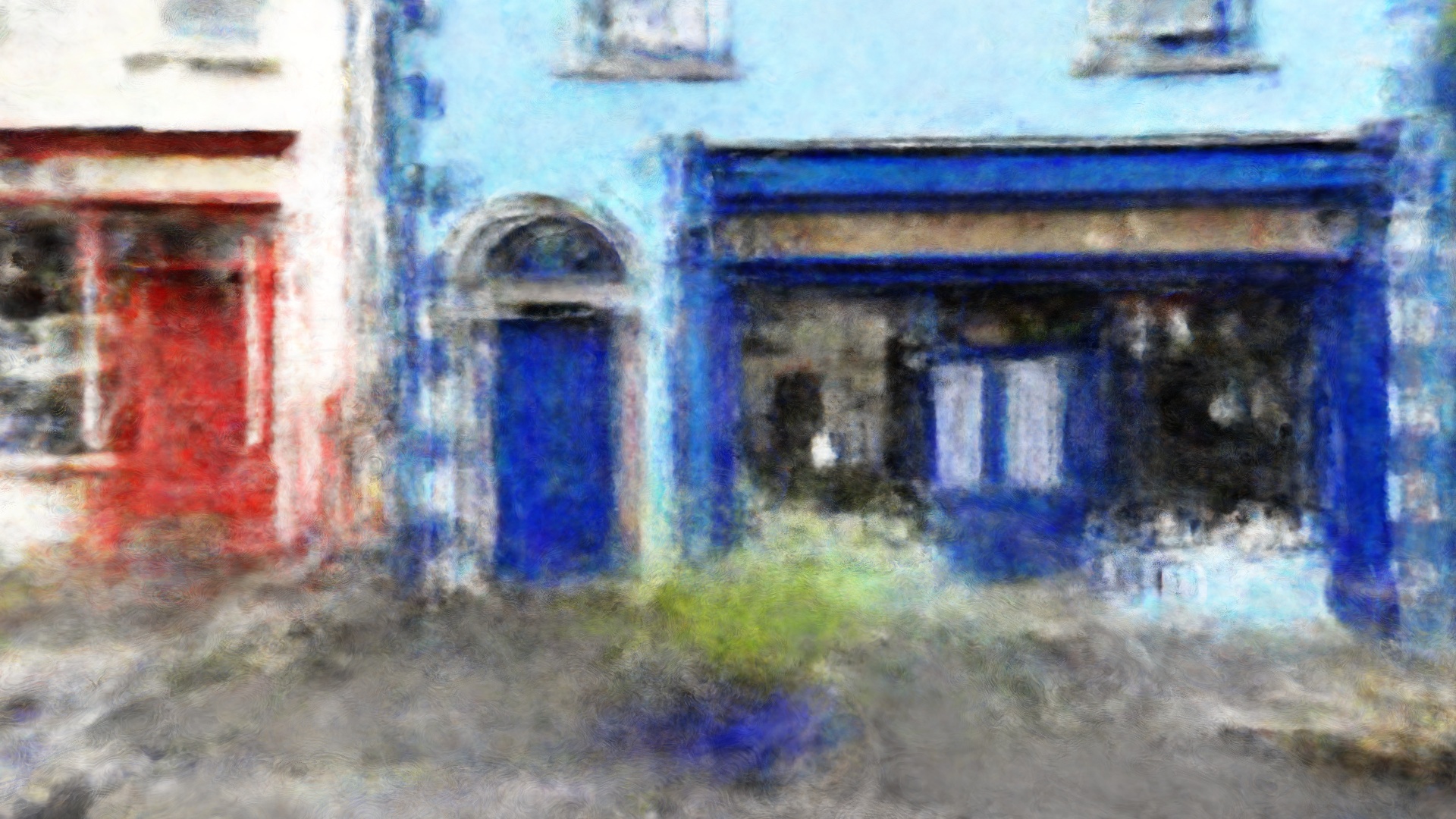}
        & \includegraphics[height=1.95cm]{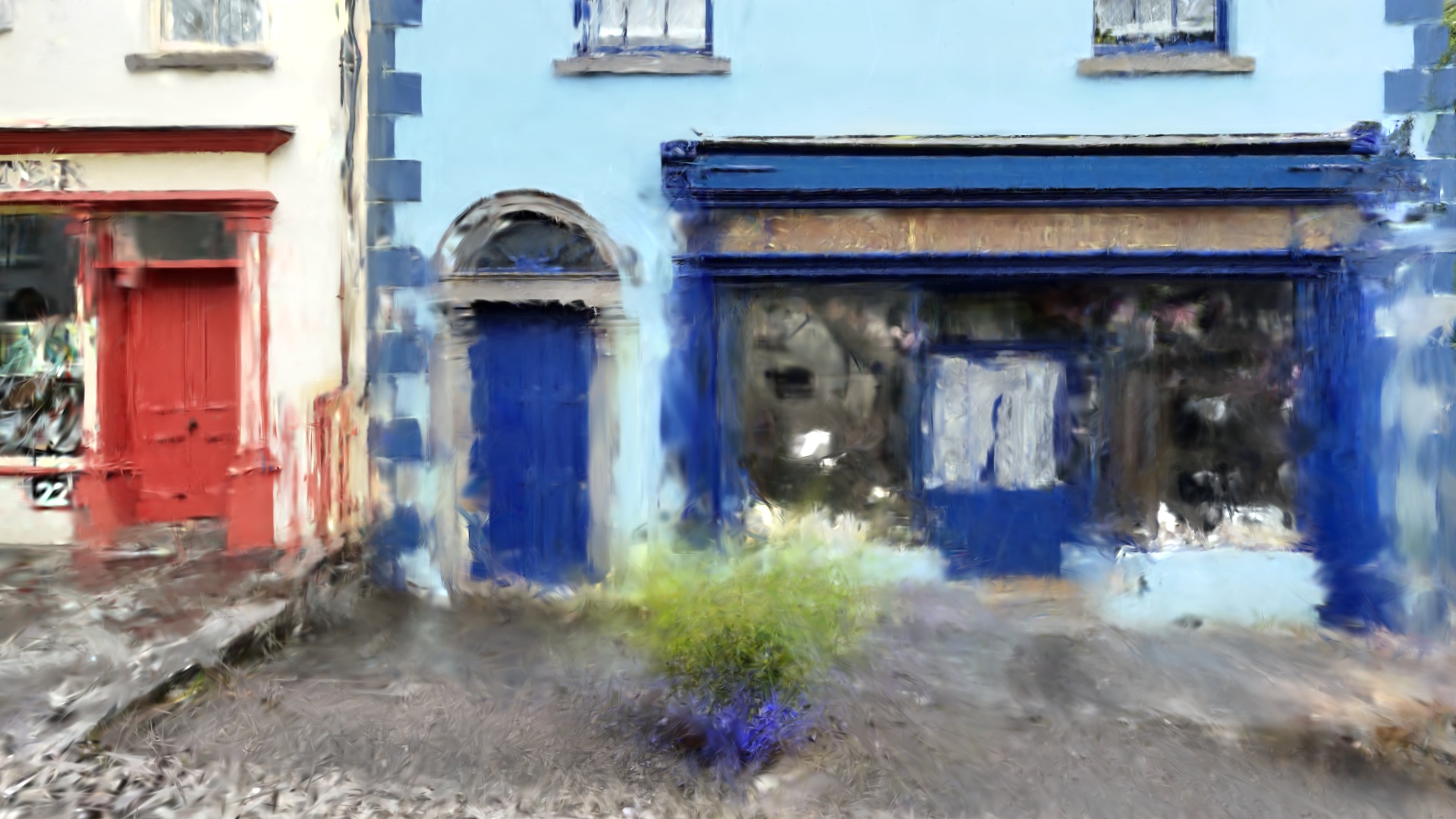} \\

        & MM
        & \includegraphics[height=1.95cm]{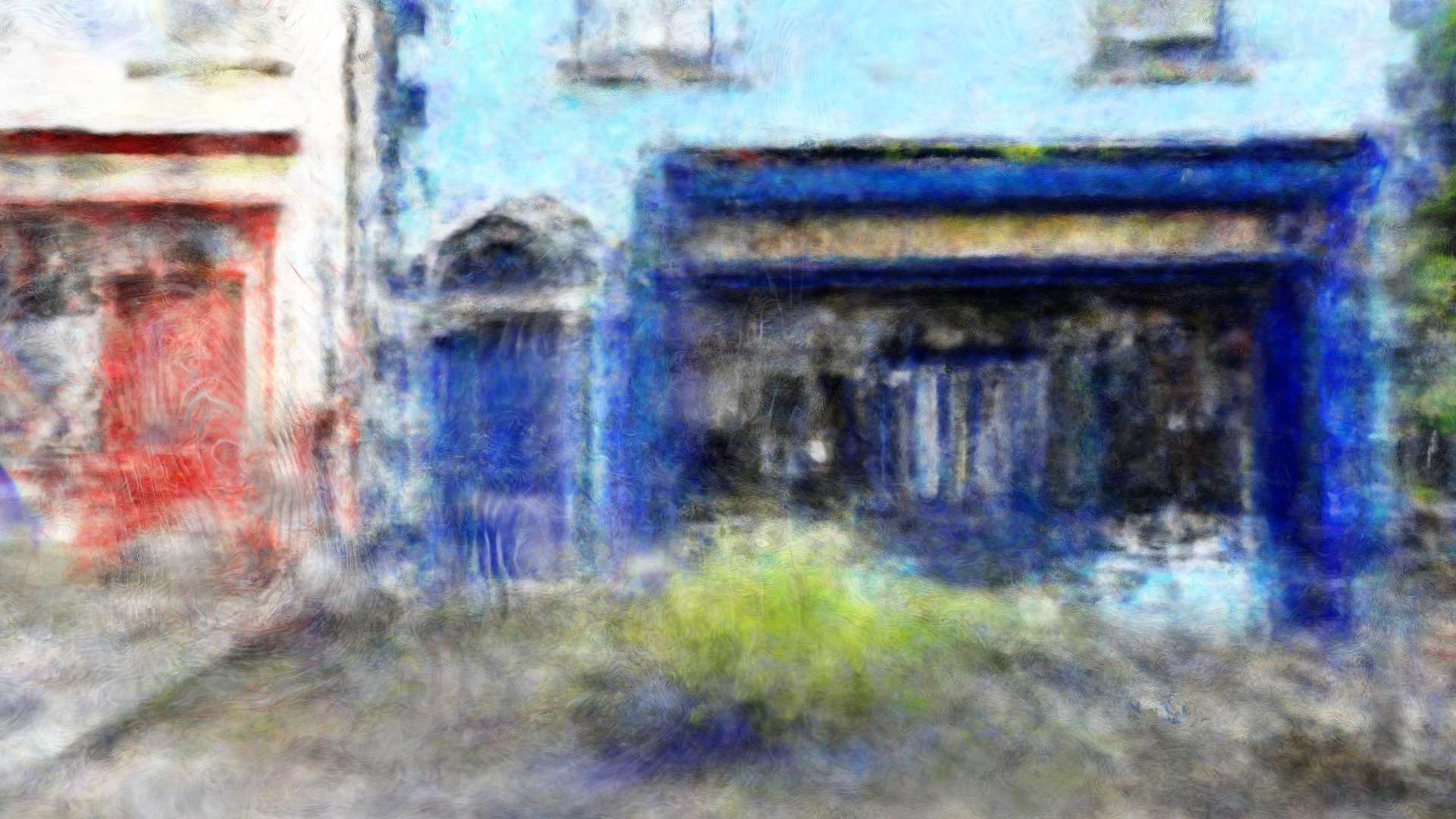}
        & \includegraphics[height=1.95cm]{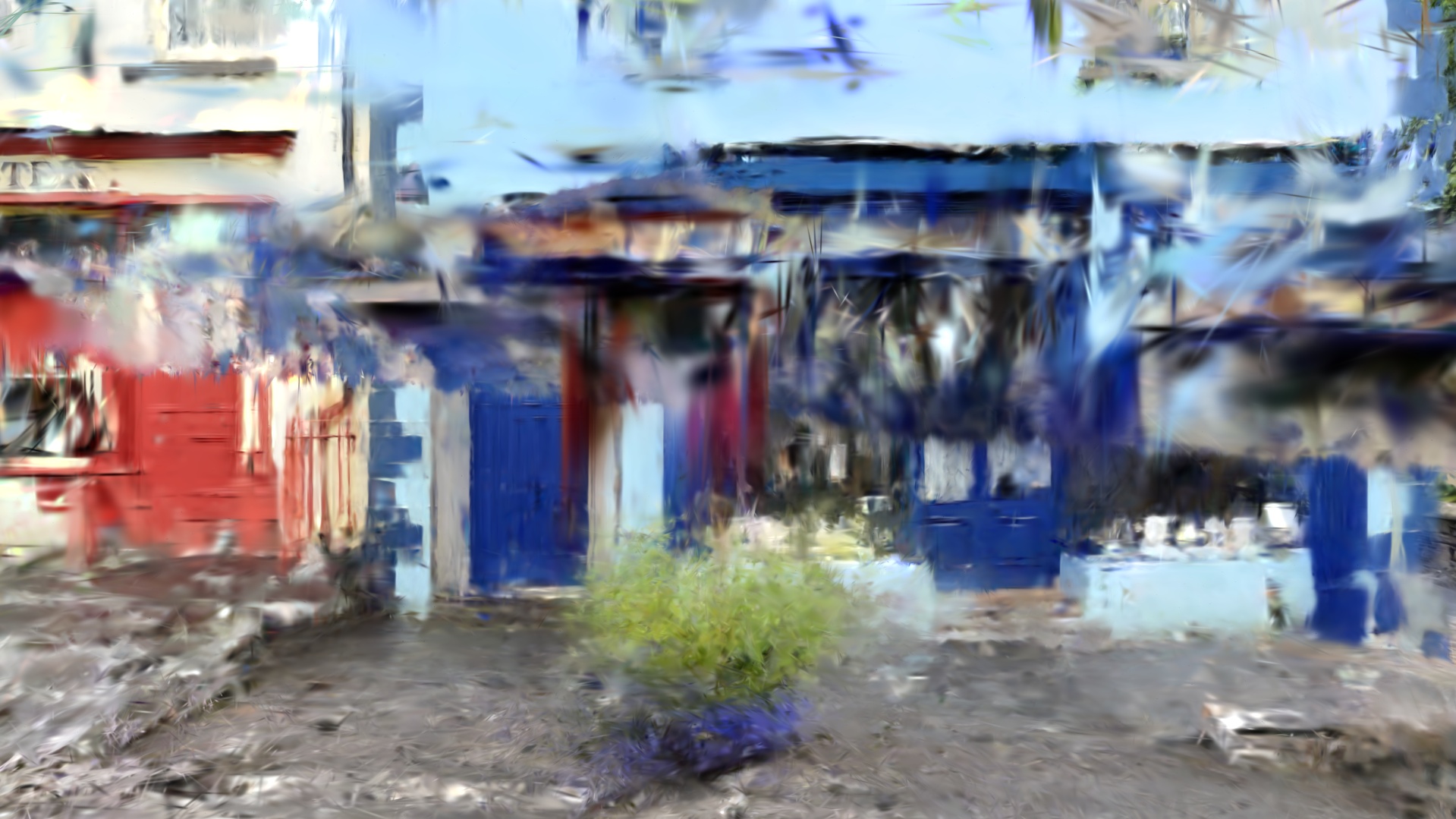} \\

        \hline

        % ================= Church =================
        \multirow{5}{*}{Church}

        & GT
        & \includegraphics[height=1.95cm]{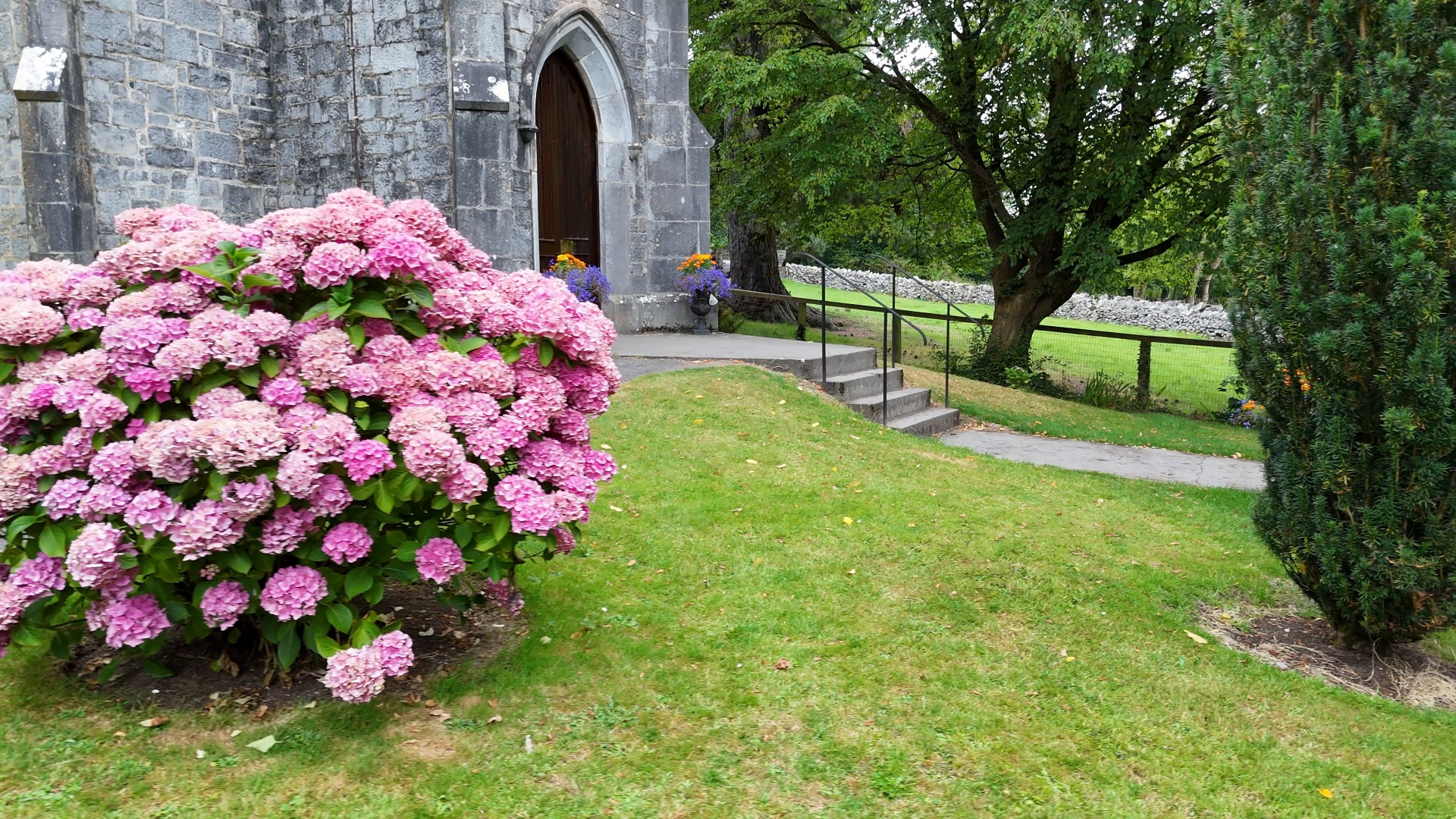}
        & \includegraphics[height=1.95cm]{fig/bm2/gt_eval_orbit_inward_low_frame_000597.JPG} \\

        & OO
        & \includegraphics[height=1.95cm]{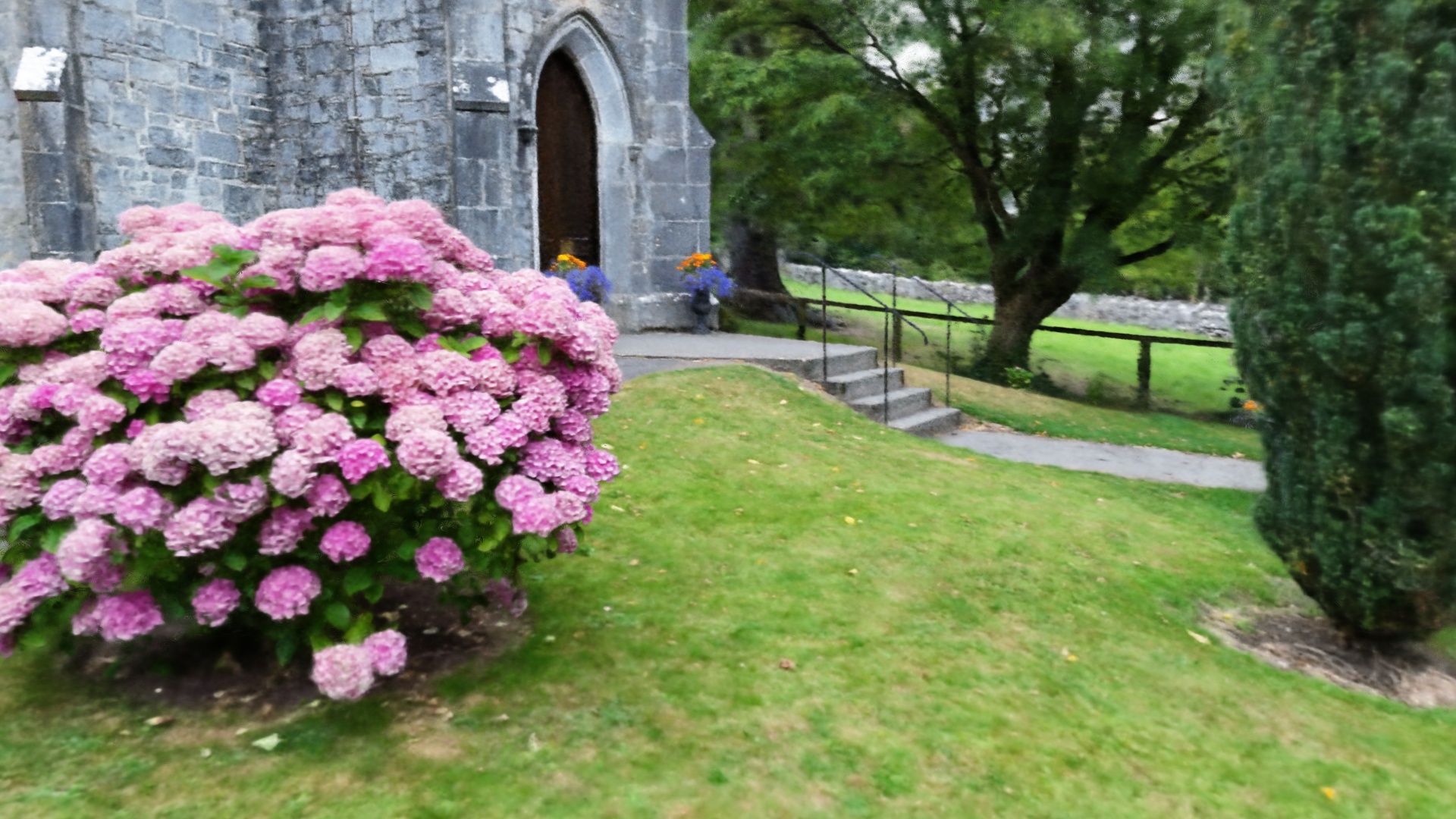}
        & \includegraphics[height=1.95cm]{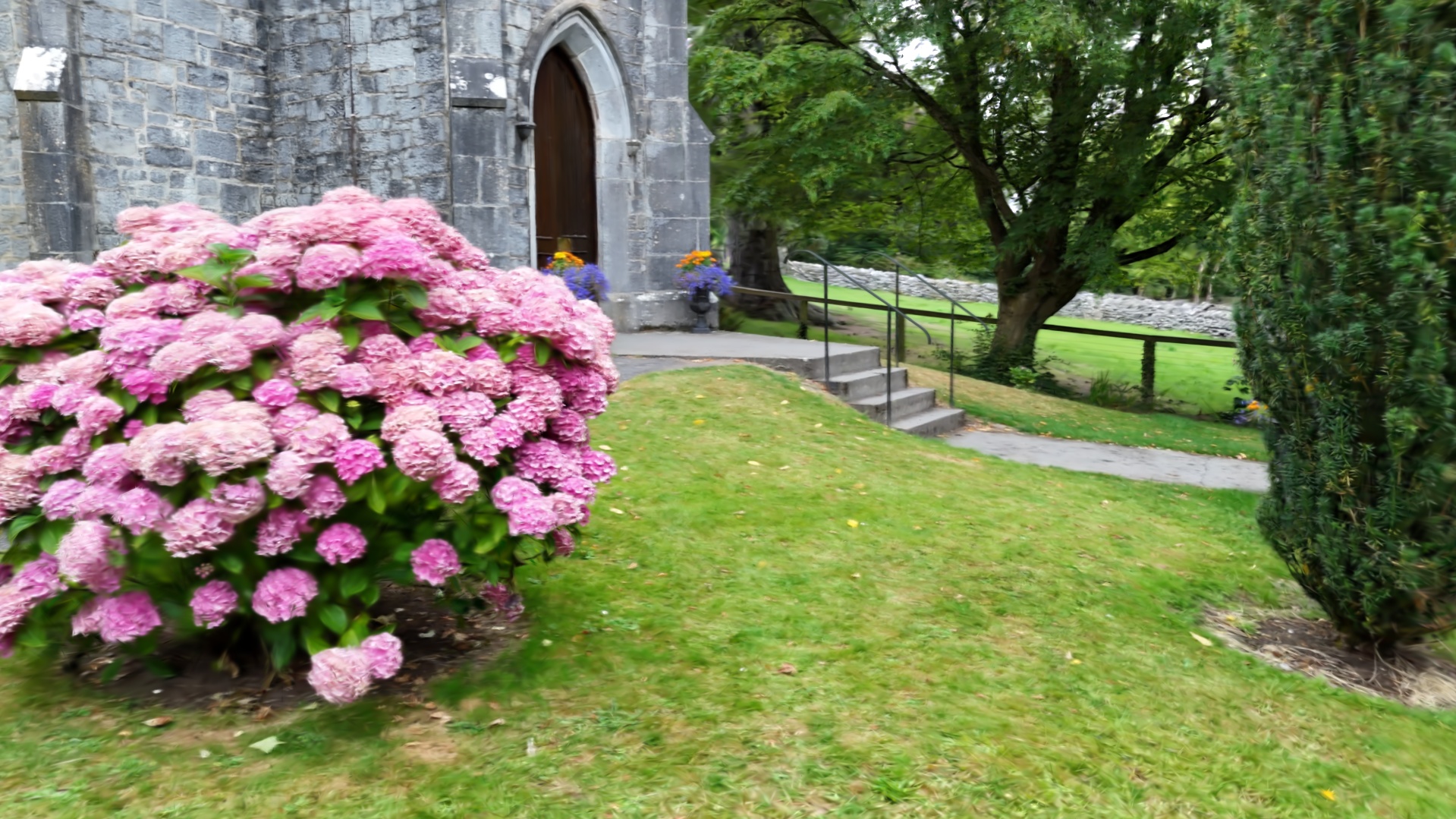} \\

        & MO
        & \includegraphics[height=1.95cm]{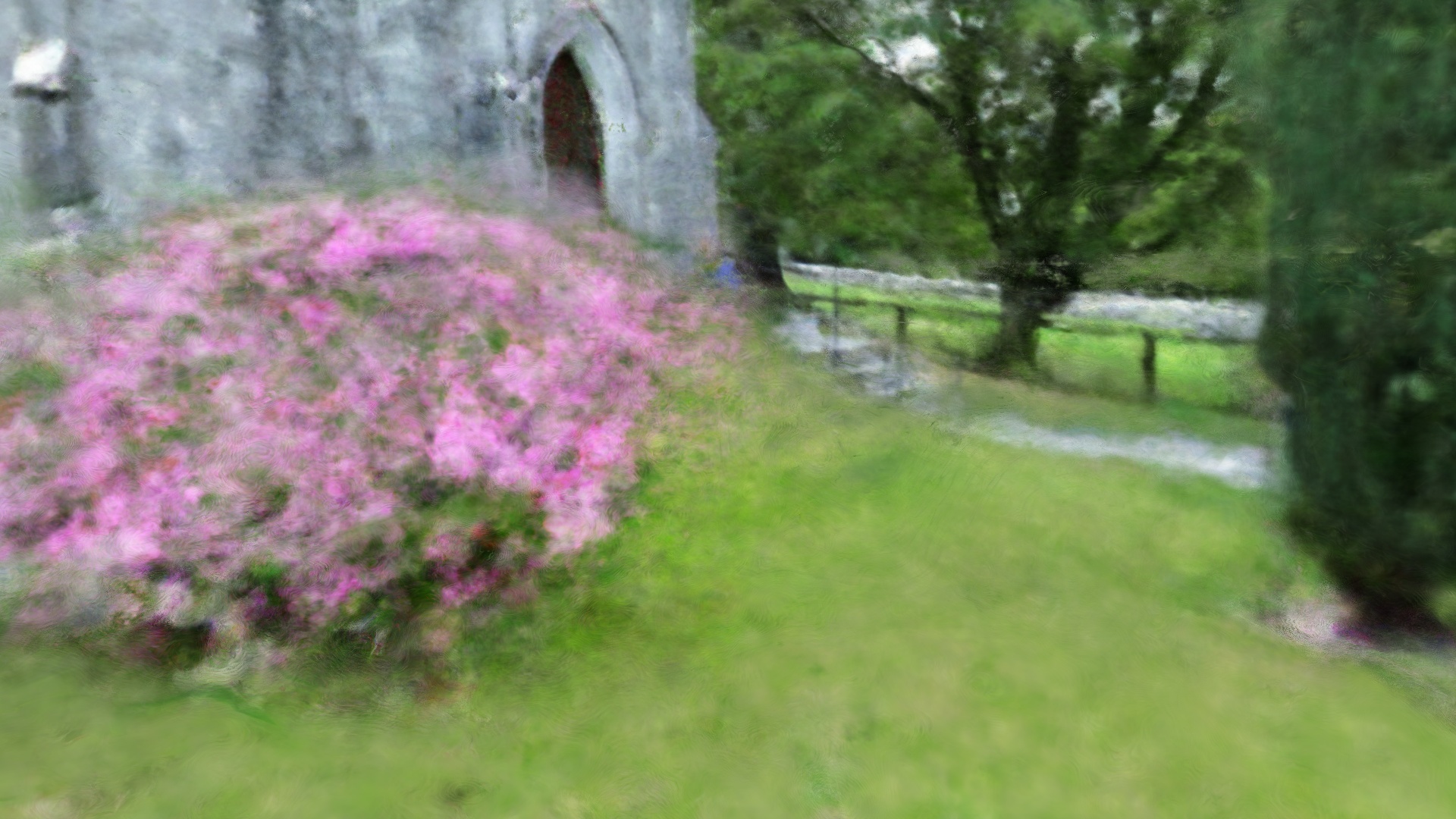}
        & \includegraphics[height=1.95cm]{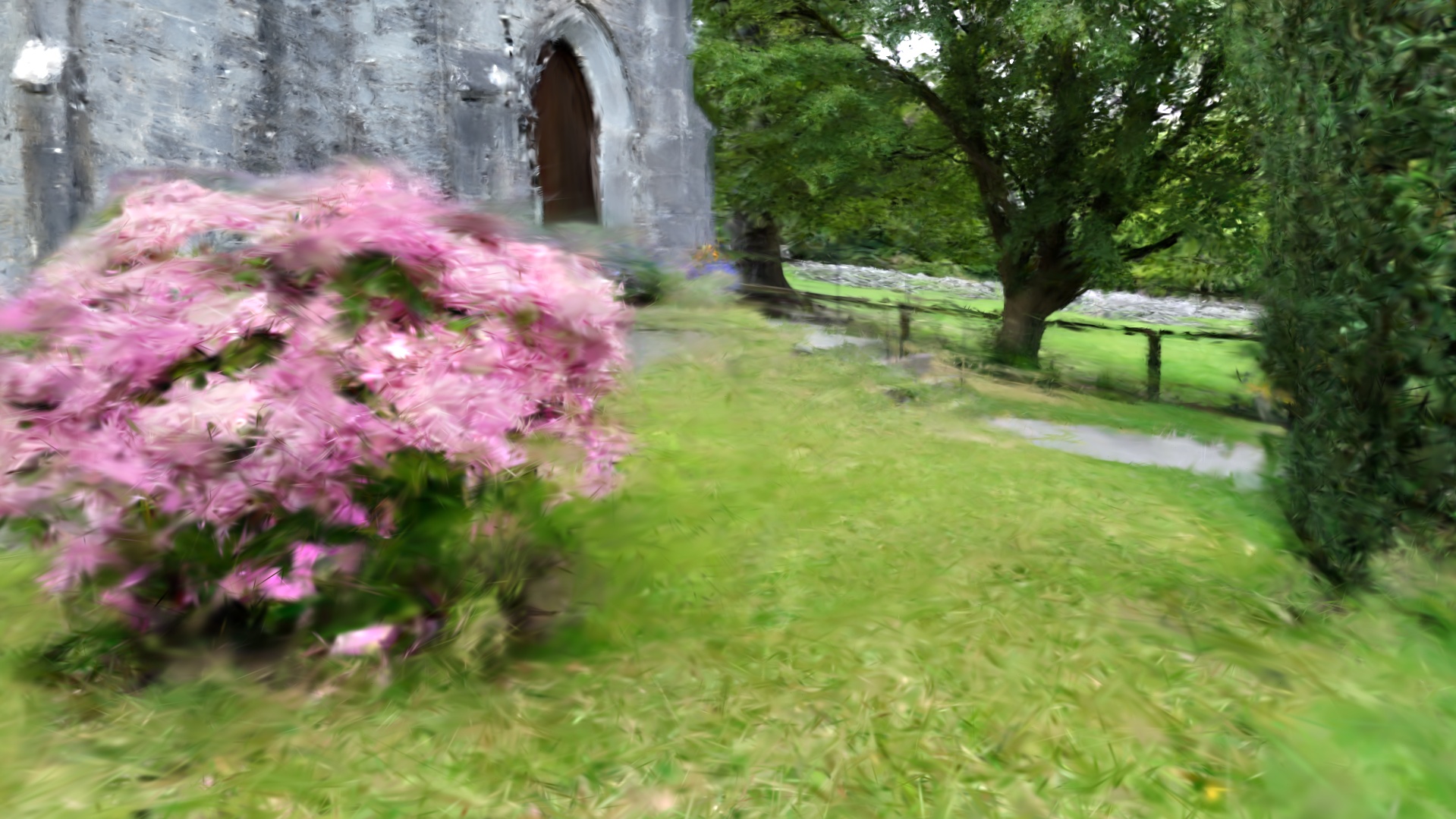} \\

        & OM
        & \includegraphics[height=1.95cm]{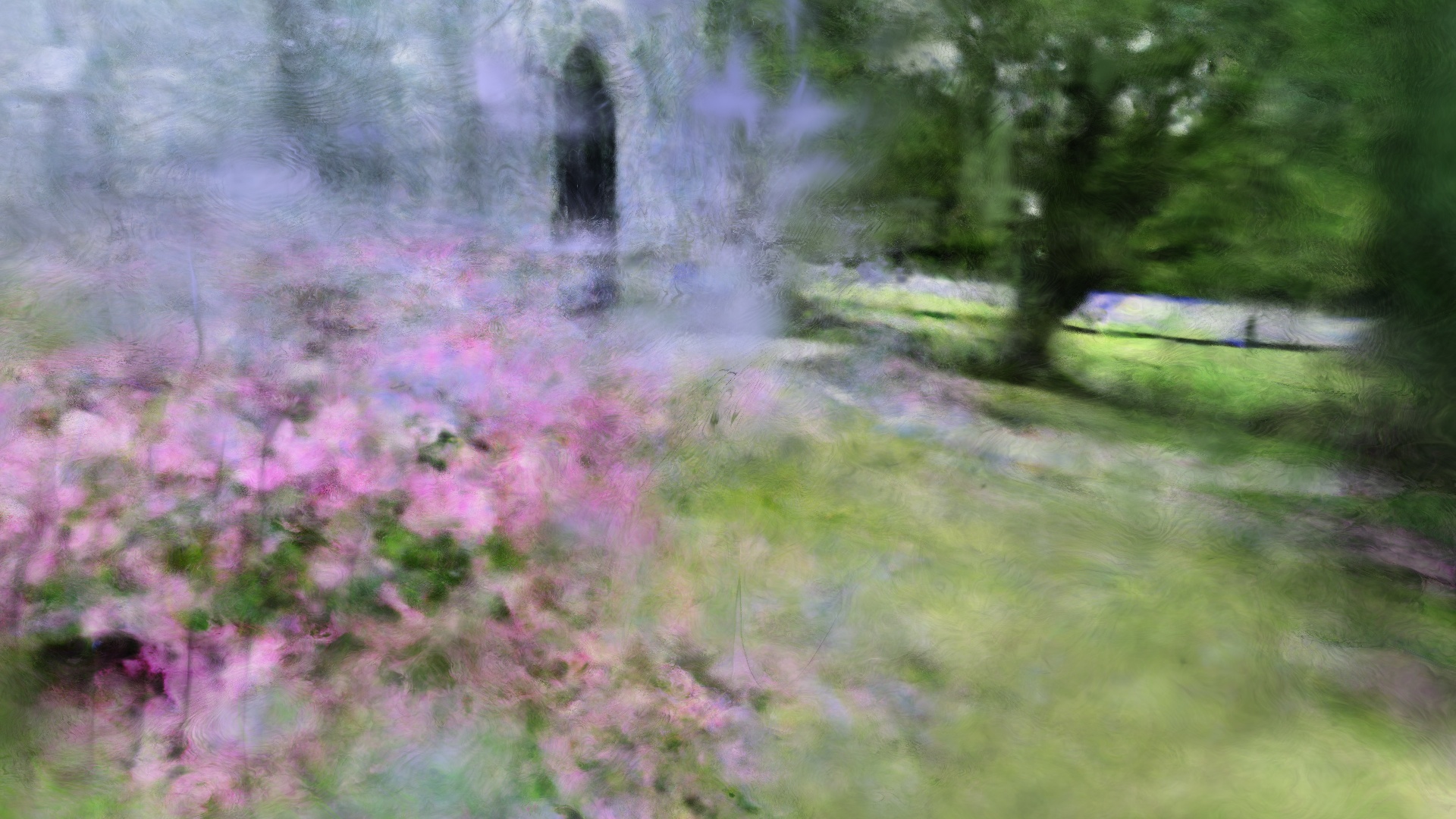}
        & \includegraphics[height=1.95cm]{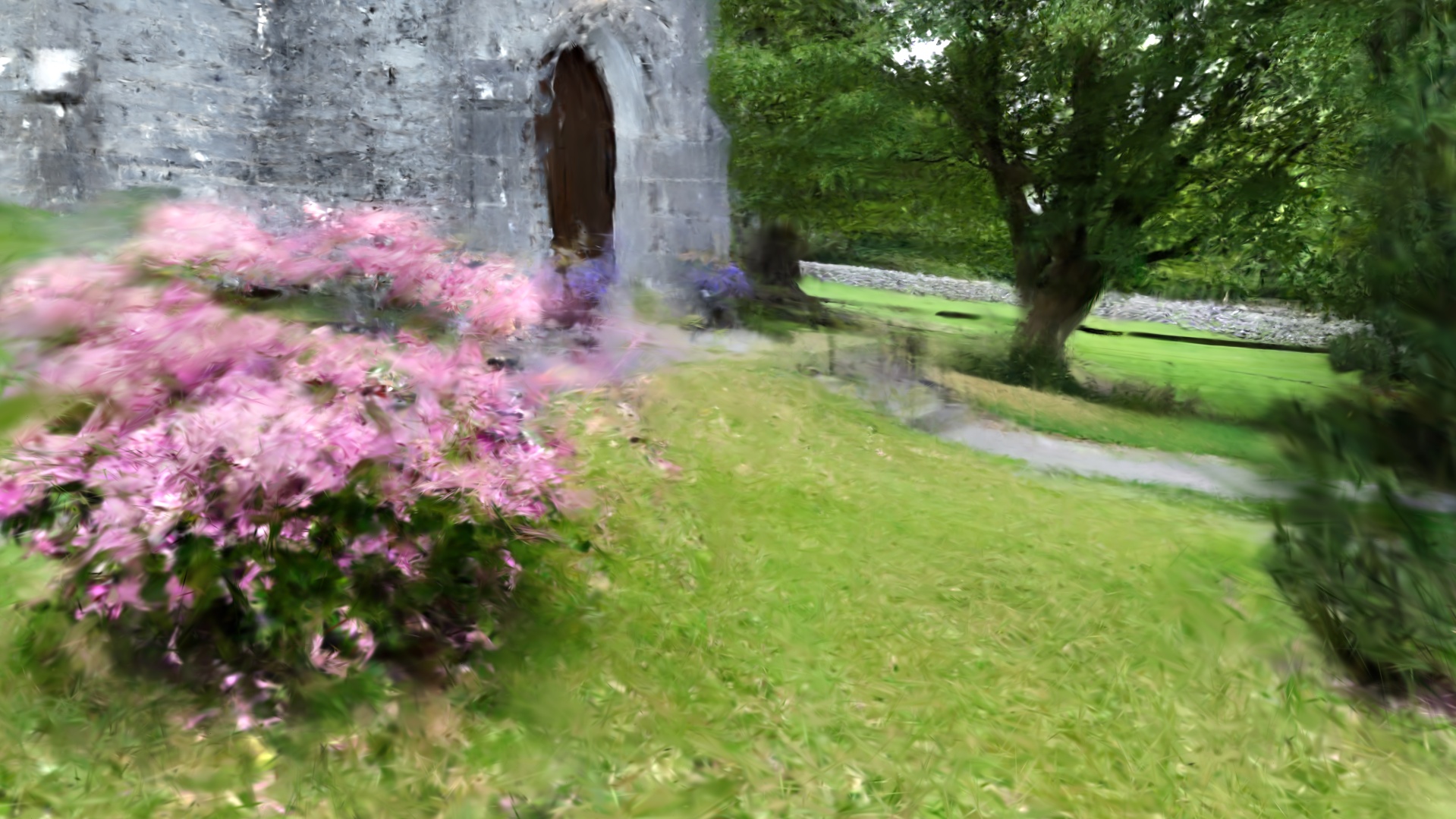} \\

        & MM
        & \includegraphics[height=1.95cm]{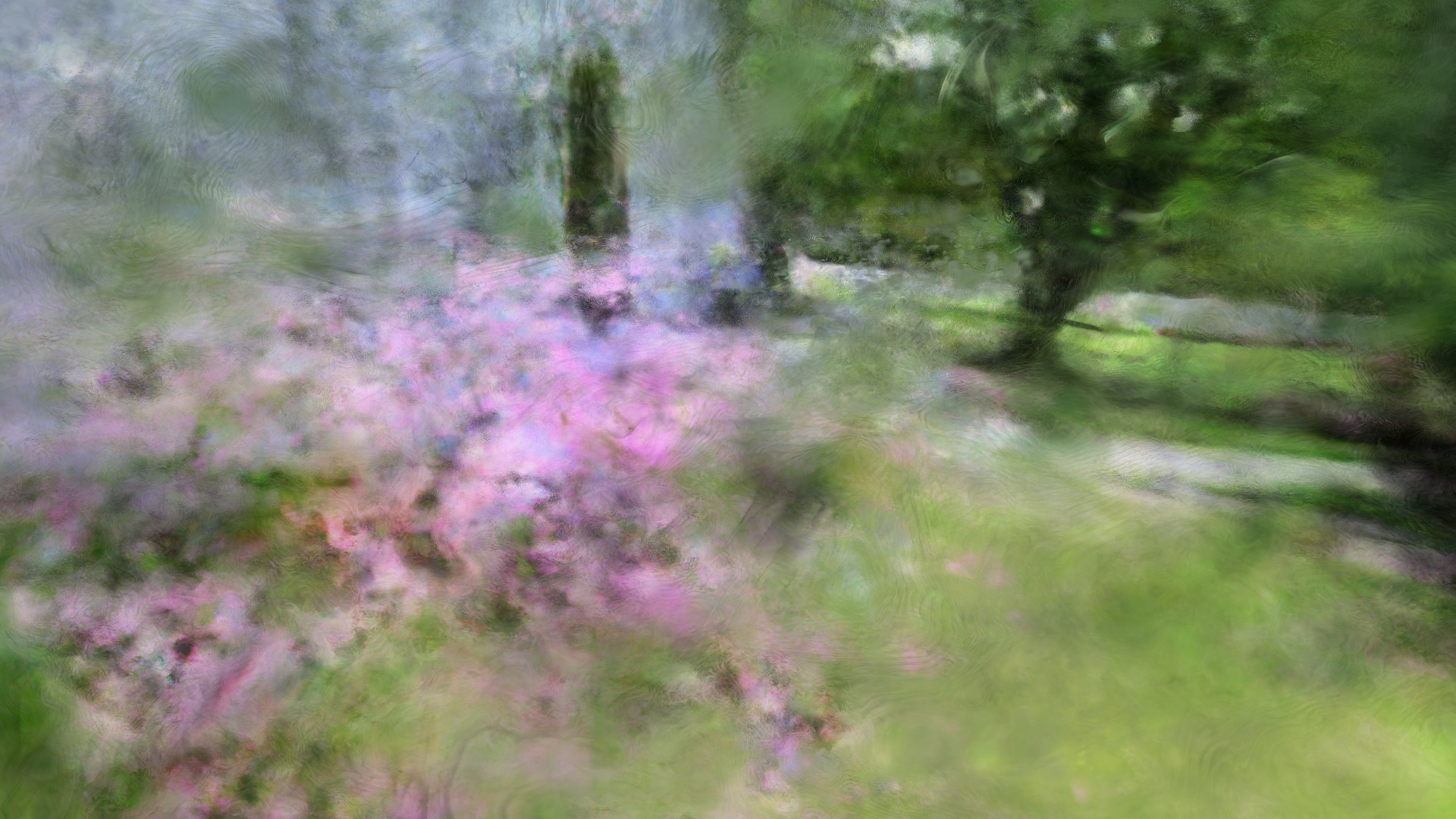}
        & \includegraphics[height=1.95cm]{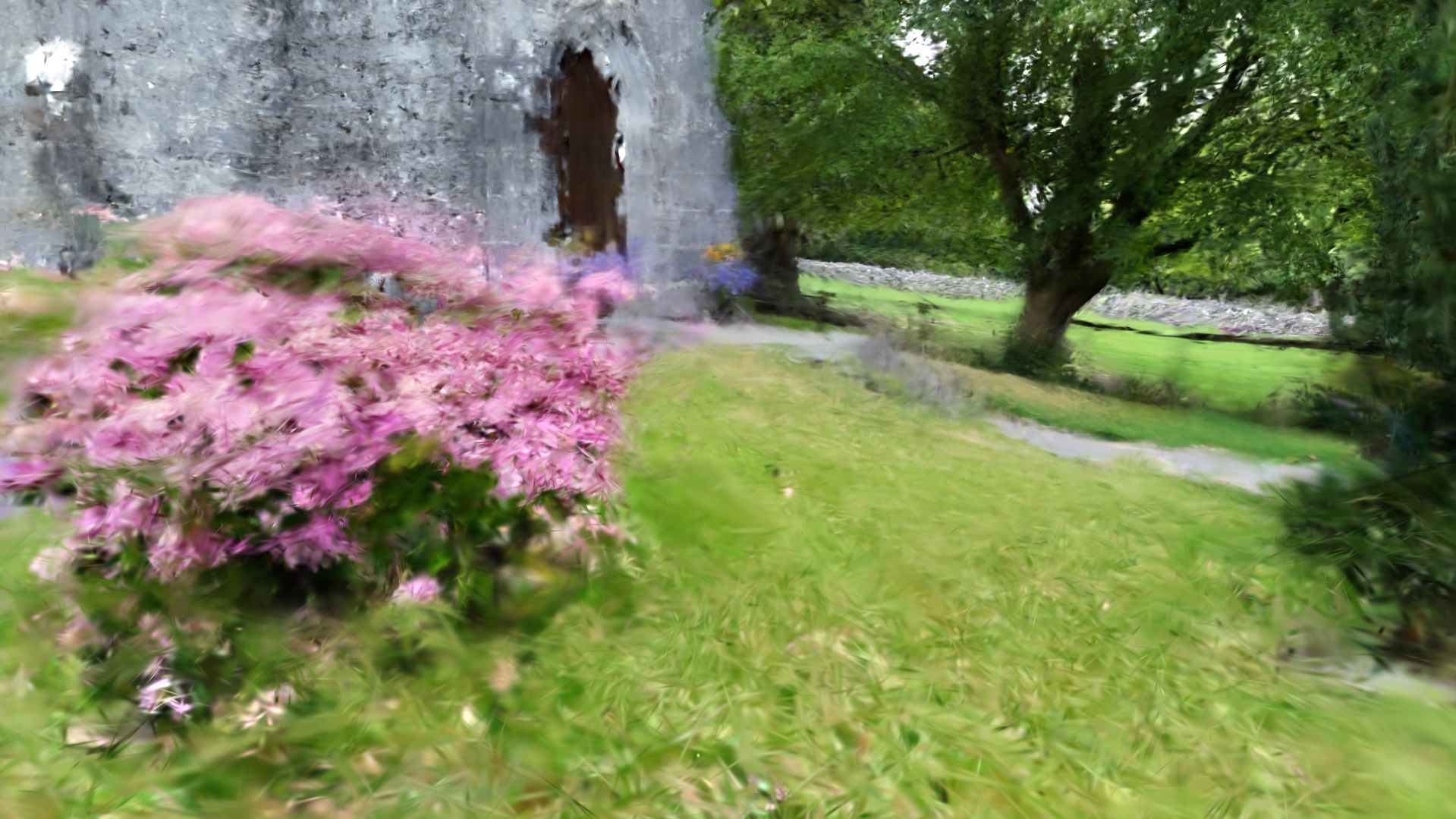} \\

    \end{tabular}

    \caption{
    Qualitative BM2:Measured vs. Optimized Poses reconstruction results for two representative scenes under
    different combinations of optimized and measured camera pose components.
    OO denotes optimized translation and optimized rotation, MO denotes measured
    translation and optimized rotation, OM denotes optimized translation and
    measured rotation, and MM denotes measured translation and measured rotation.
    Ground-truth images are shown for reference.
    }

    \label{fig:bm2_qualitative}
\end{figure*}

Figure~\ref{fig:bm2_qualitative} compares the four measured/optimized pose-source combinations for representative views.

\subsection{BM3: Calibrated vs. Optimized Intrinsics}

\begin{figure*}[!tbp]
    \centering
    \setlength{\tabcolsep}{4pt}
    \renewcommand{\arraystretch}{1.05}

    \begin{tabular}{c c c c}
        \textbf{Scene} &
        \textbf{Camera Intrinsics} &
        \textbf{Nerfacto} &
        \textbf{Splatfacto} \\
        \hline

        % ================= Village Street =================
        \multirow{3}{*}{\shortstack{Village\\Street}}

        & GT
        & \includegraphics[height=2.15cm]{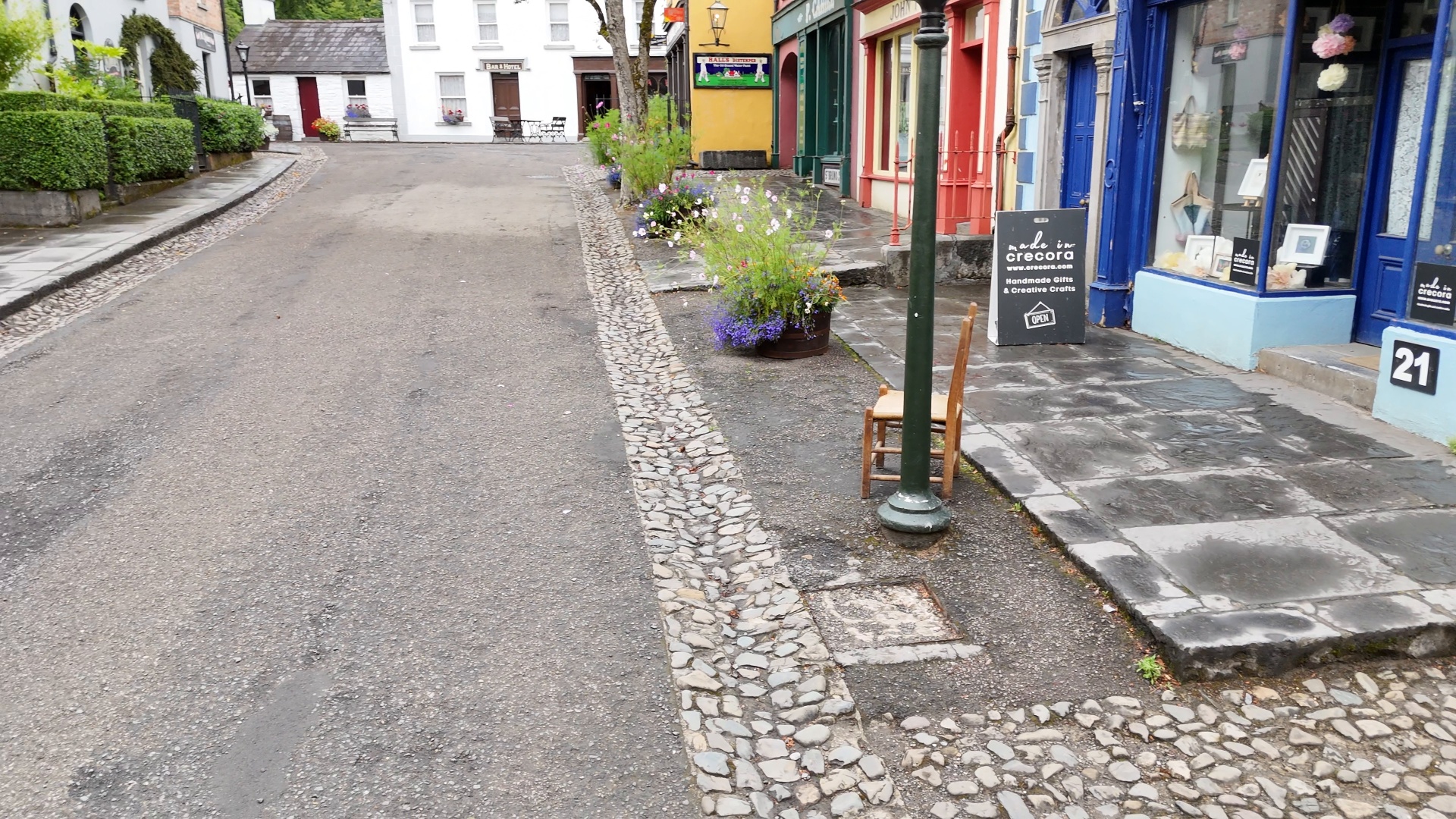}
        & \includegraphics[height=2.15cm]{fig/bm3/gt_eval_orbit_inward_mid_frame_001097.JPG} \\

        & \shortstack{COLMAP\\optimized}
        & \includegraphics[height=2.15cm]{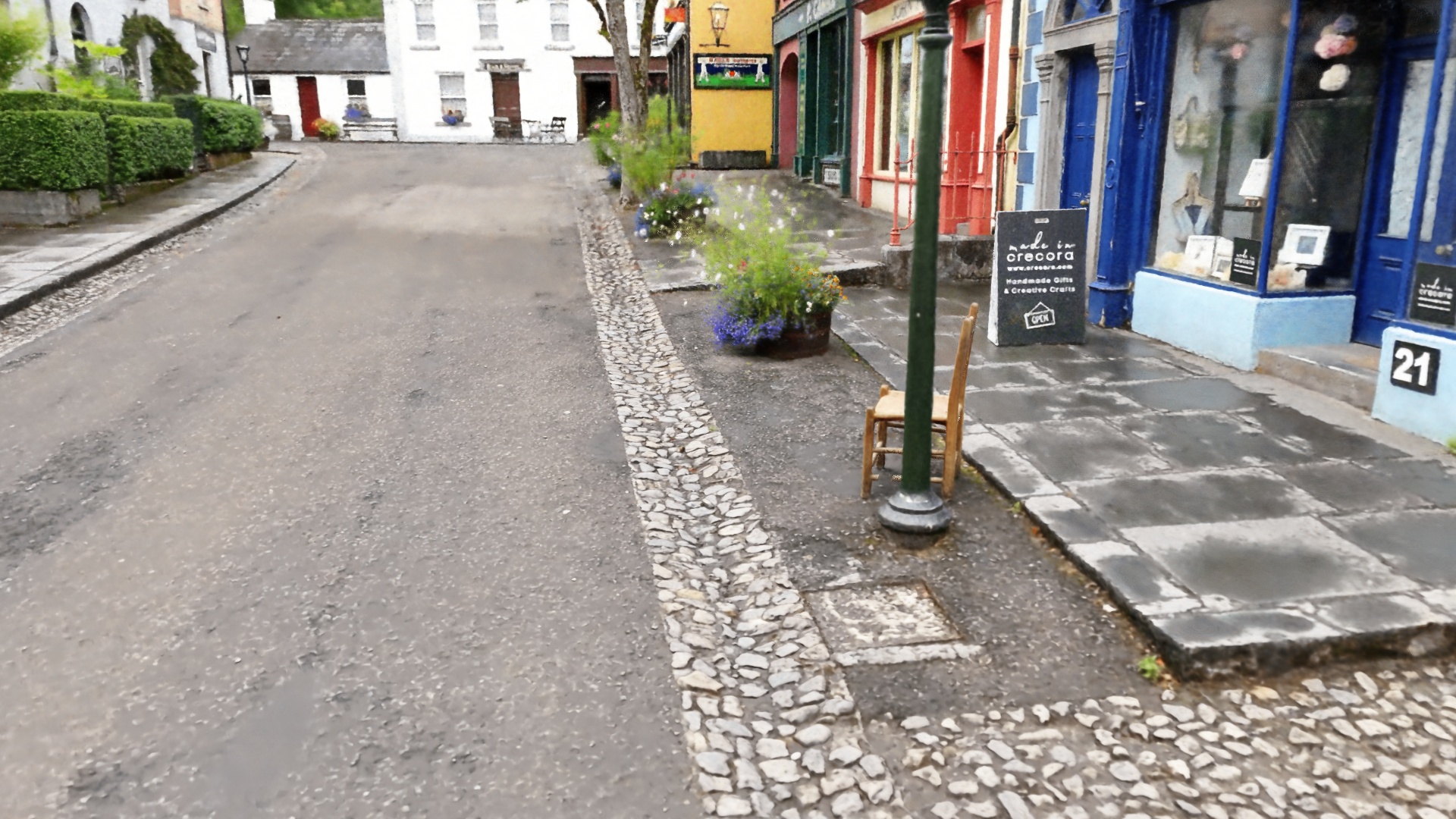}
        & \includegraphics[height=2.15cm]{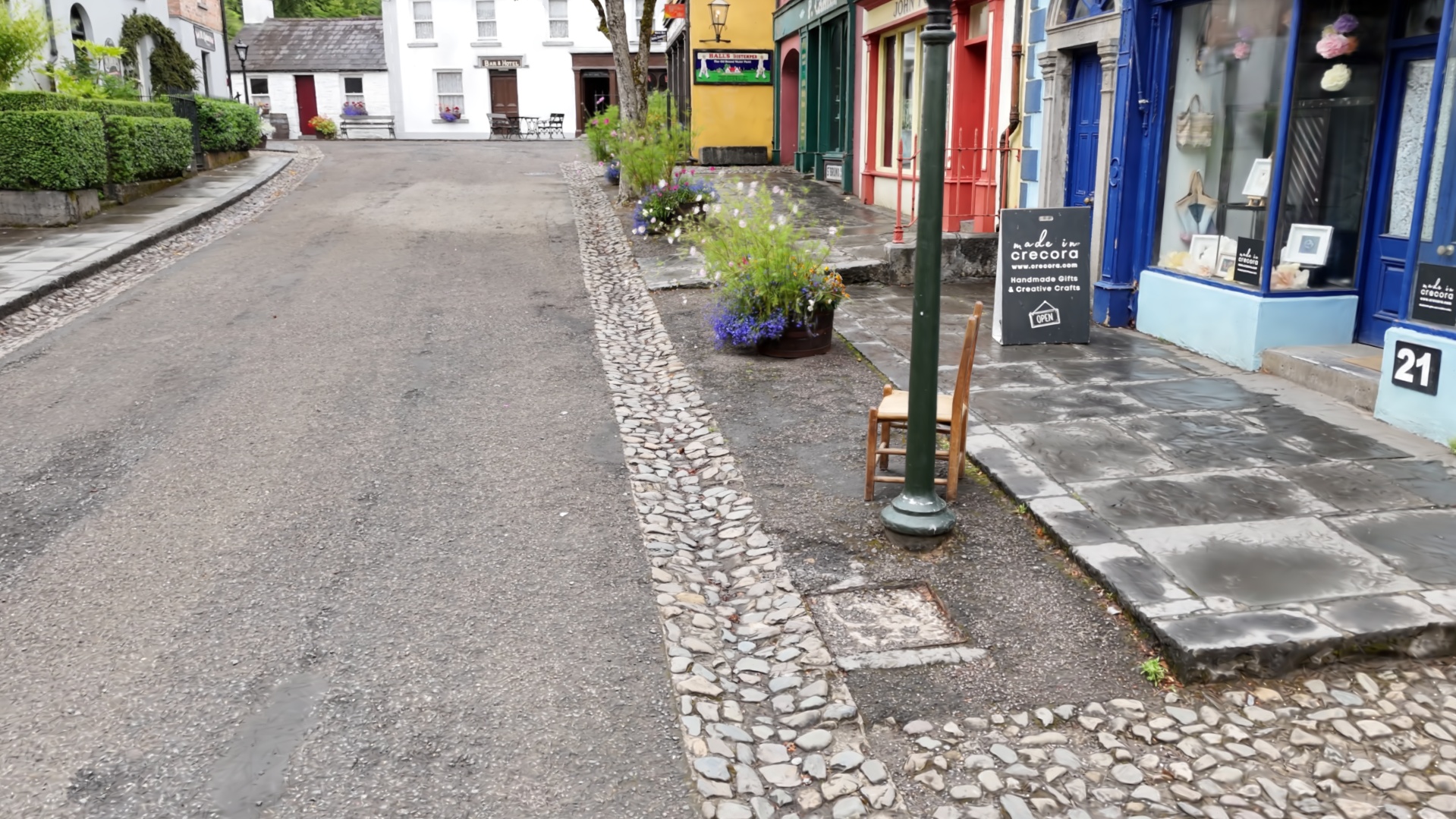} \\

        & \shortstack{OpenCV\\calibrated}
        & \includegraphics[height=2.15cm]{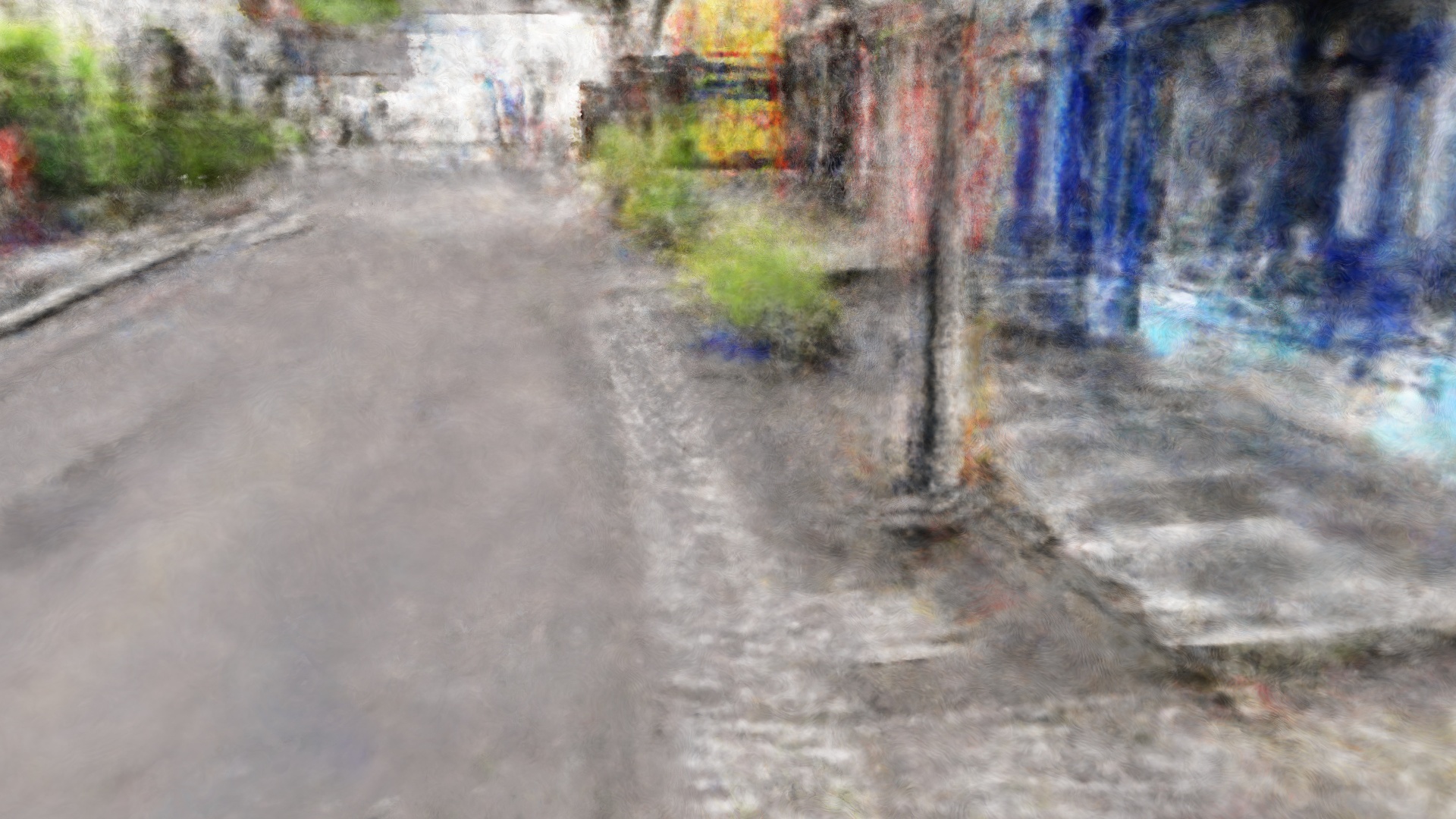}
        & \includegraphics[height=2.15cm]{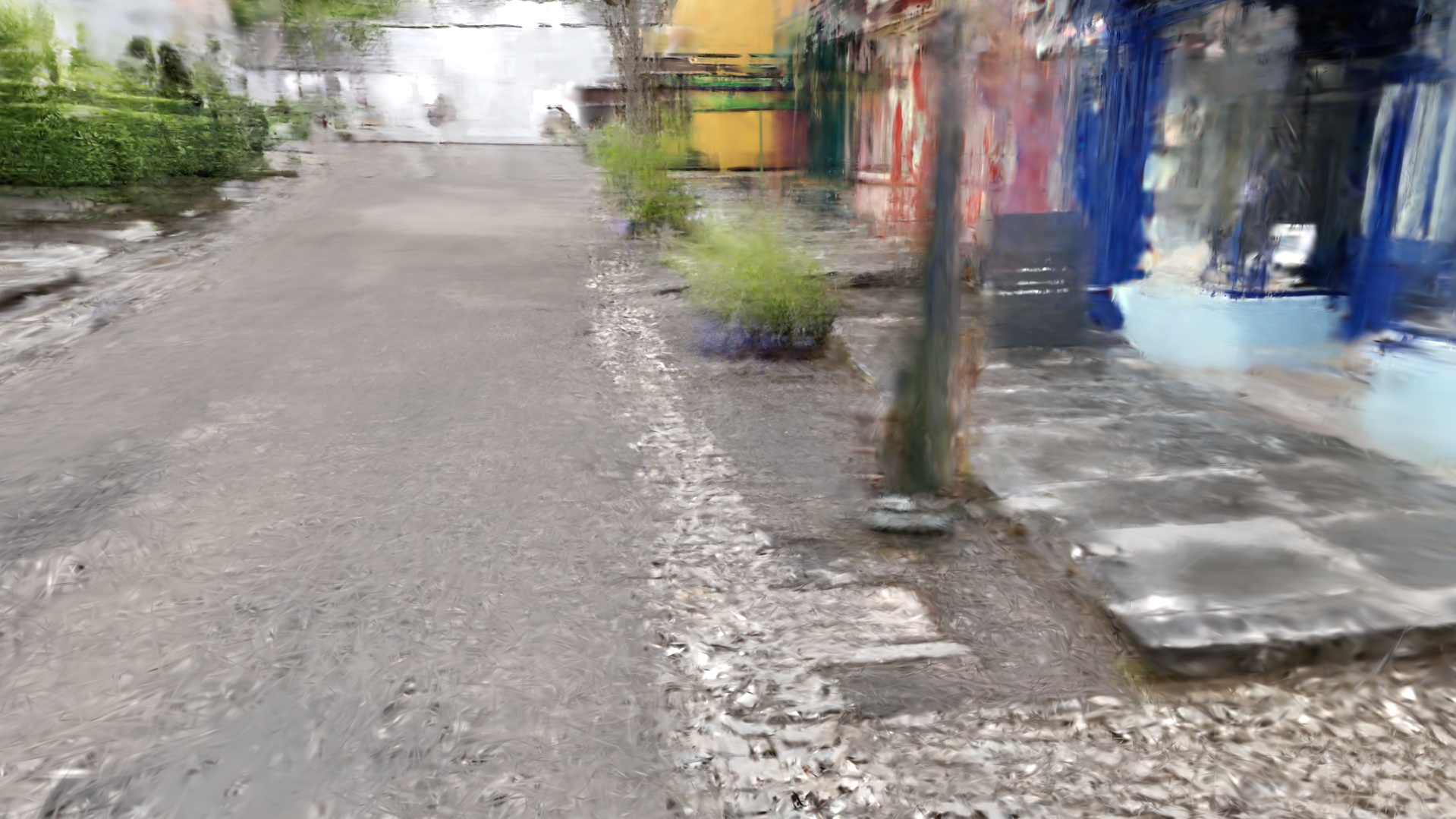} \\

        \hline

        % ================= Church =================
        \multirow{3}{*}{Church}

        & GT
        & \includegraphics[height=2.15cm]{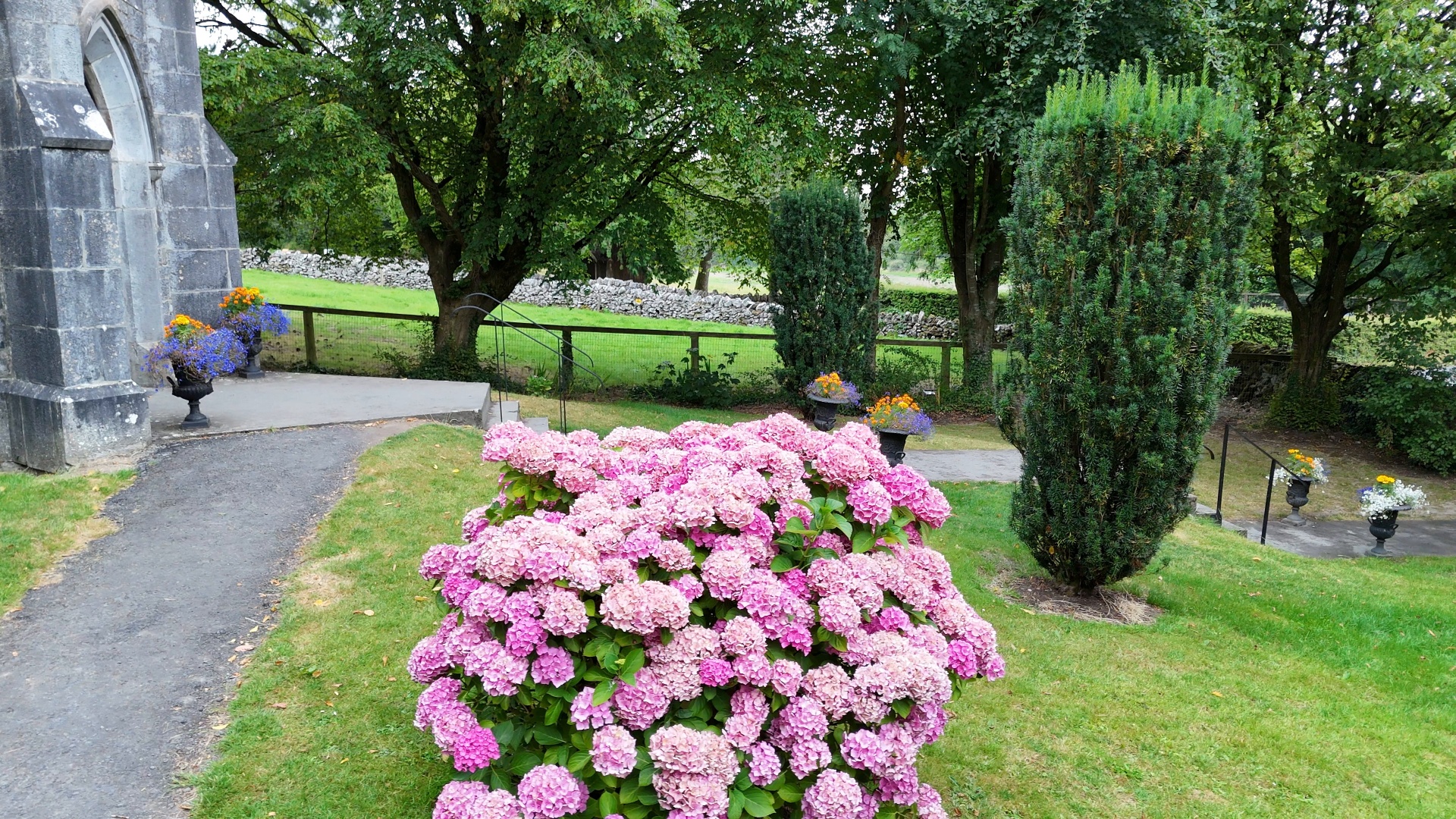}
        & \includegraphics[height=2.15cm]{fig/bm3/gt_eval_orbit_inward_mid_frame_000361.JPG} \\

        & \shortstack{COLMAP\\optimized}
        & \includegraphics[height=2.15cm]{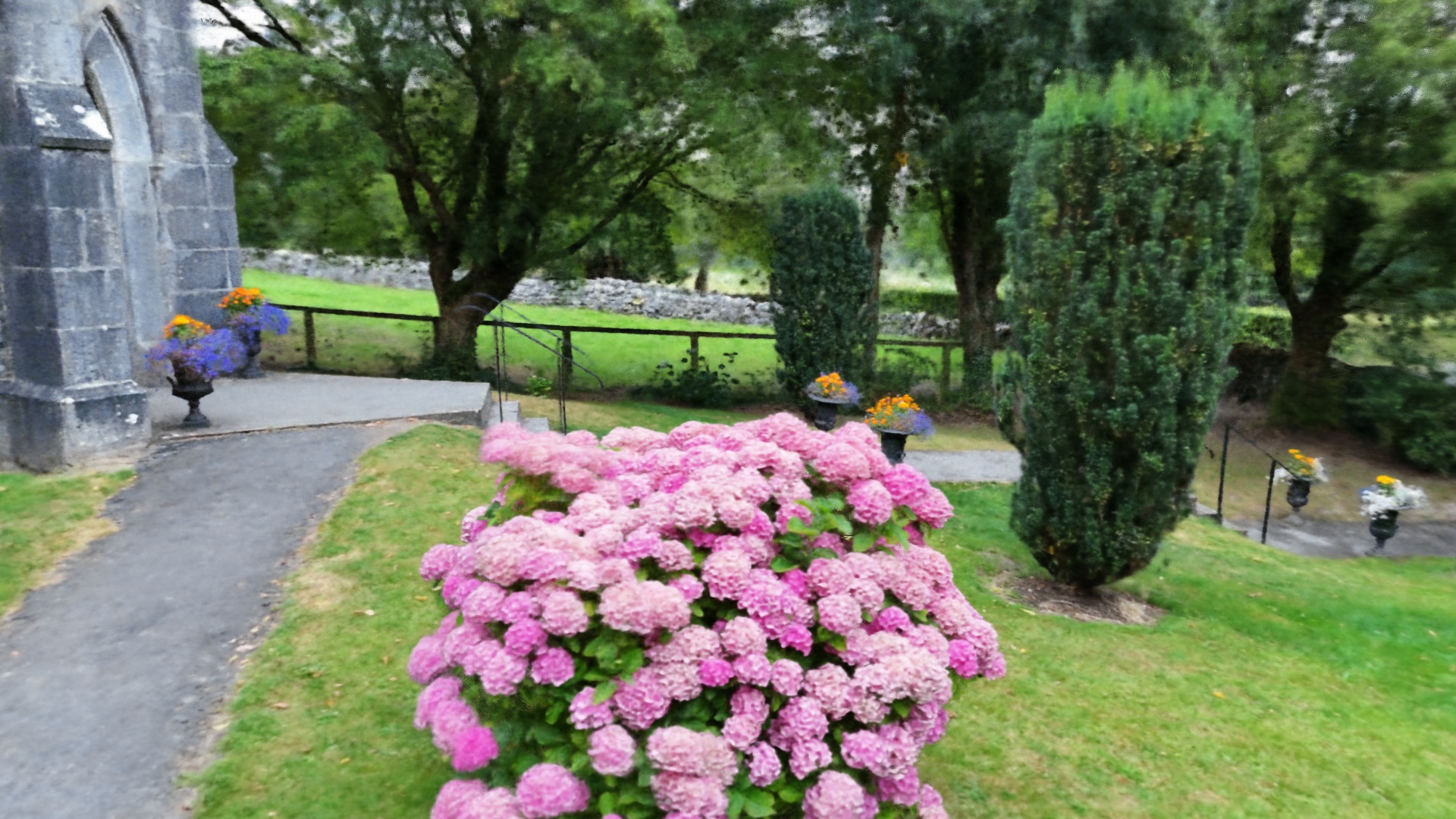}
        & \includegraphics[height=2.15cm]{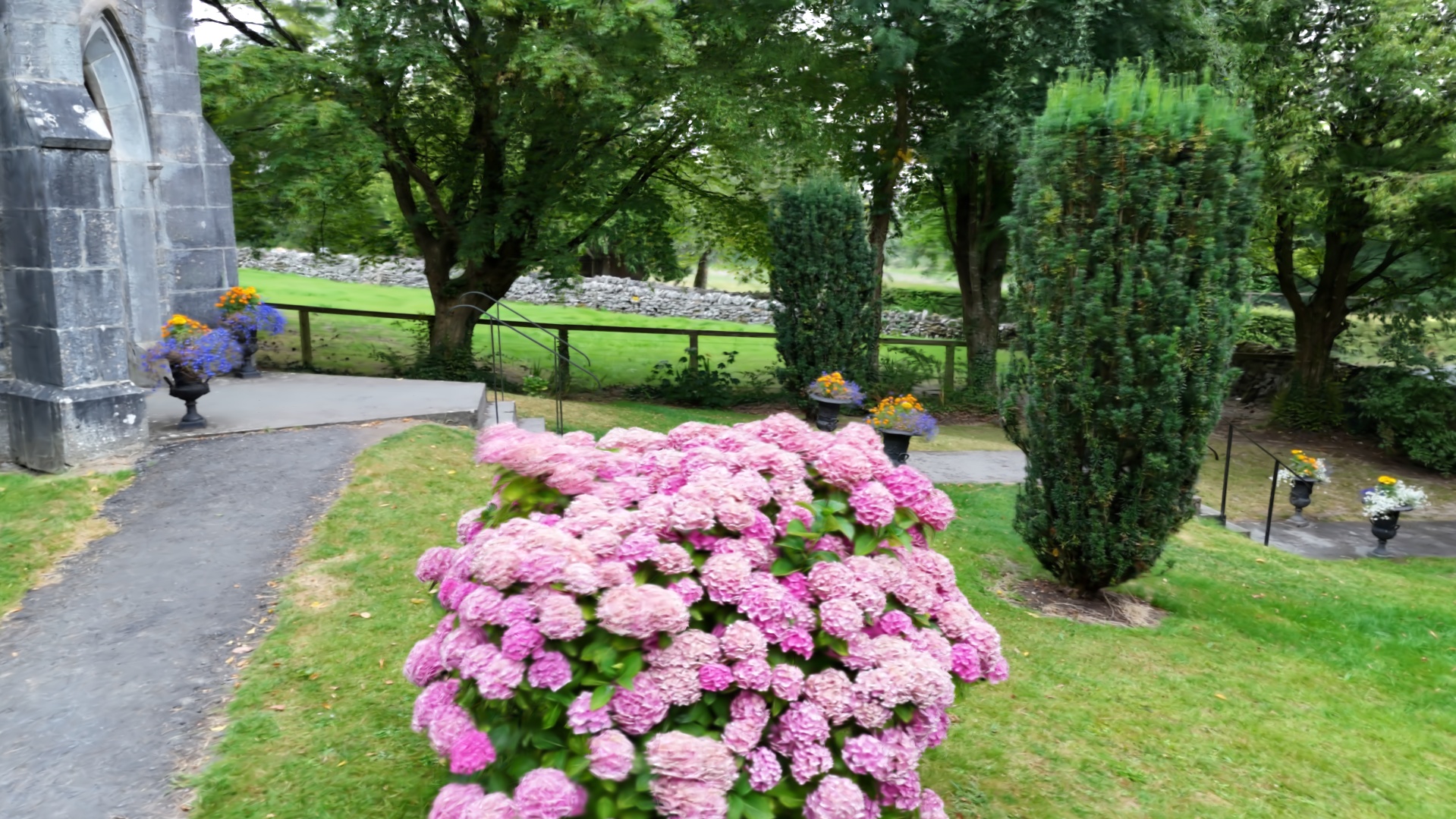} \\

        & \shortstack{OpenCV\\calibrated}
        & \includegraphics[height=2.15cm]{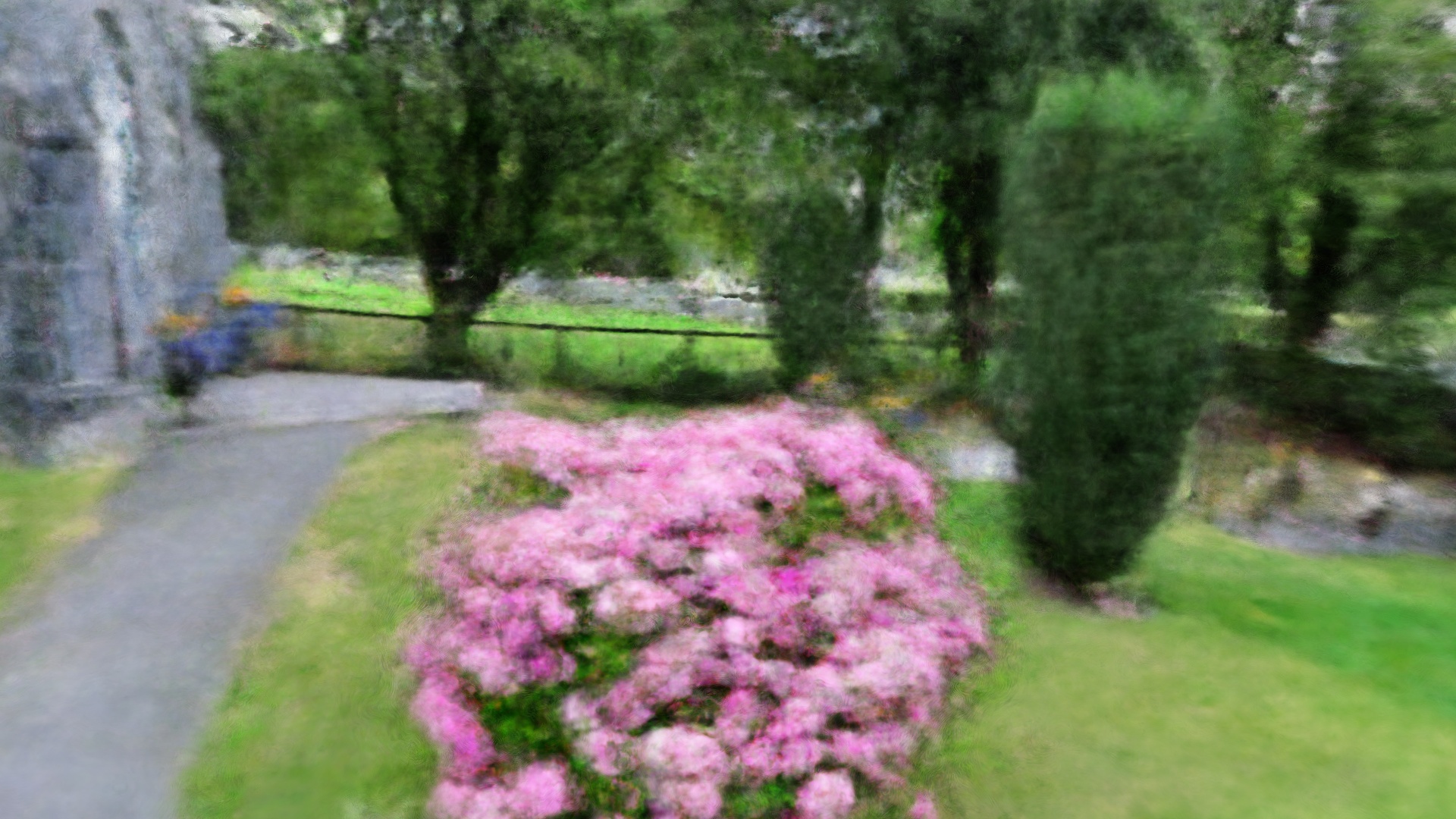}
        & \includegraphics[height=2.15cm]{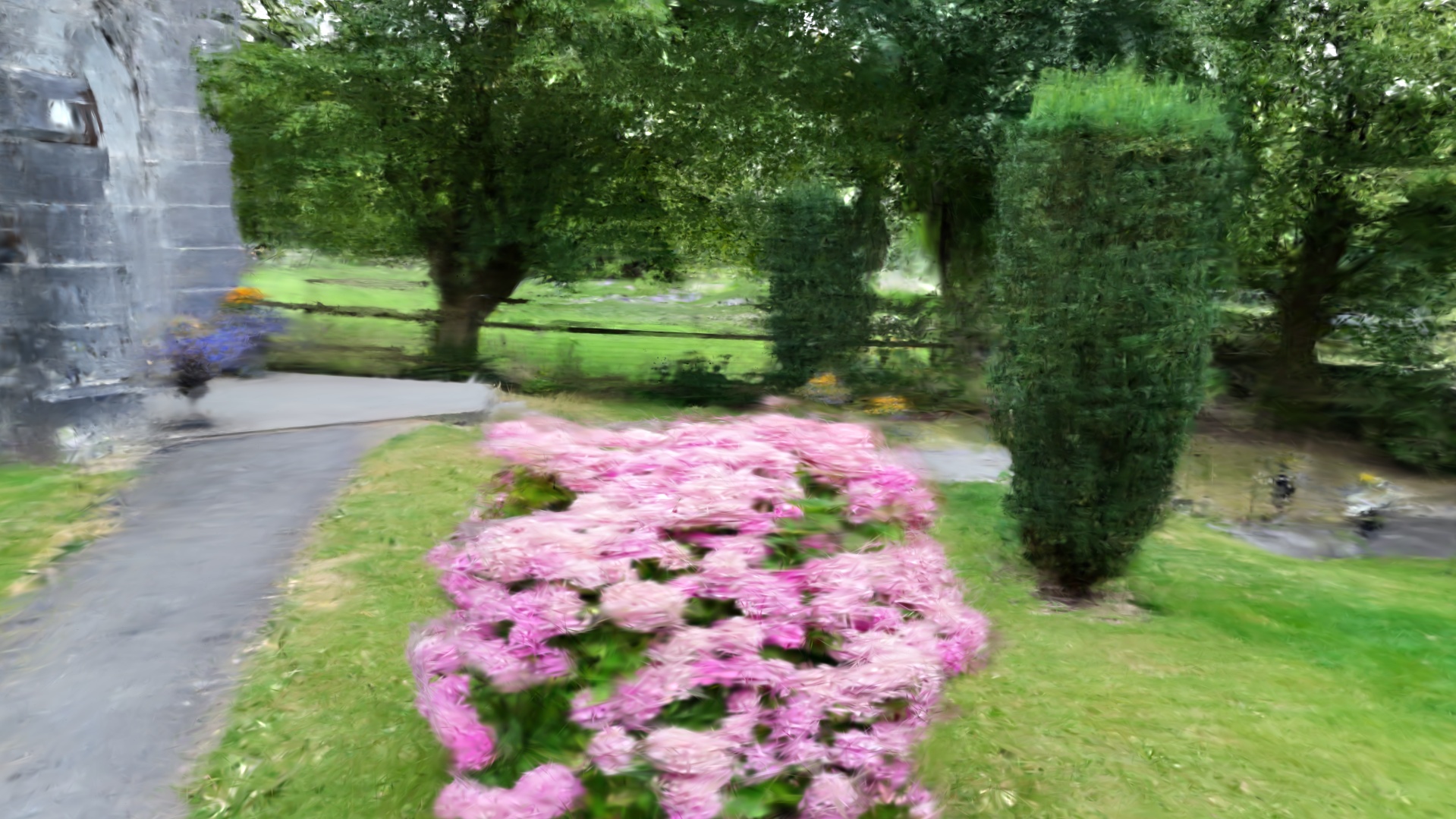} \\

    \end{tabular}

    \caption{
    Qualitative BM3 reconstruction results for two representative scenes using
    COLMAP-optimized and independently calibrated camera intrinsics.
    Ground-truth (GT) images are shown for reference. The COLMAP condition uses
    scene-optimized camera intrinsics, while the calibrated condition uses the
    fixed OpenCV camera calibration.
    }

    \label{fig:bm3_qualitative}
\end{figure*}

Figure~\ref{fig:bm3_qualitative} compares COLMAP-optimized scene intrinsics with the fixed OpenCV camera calibration.

\section{Discussion}
\label{sec:discussion}

\subsection{Trajectory-Level Evaluation}

The central motivation of \pivot{} is that held-out frames from a trajectory represented during training and views from an entirely different trajectory answer different evaluation questions. The former primarily probes interpolation within a sampled capture distribution; the latter probes how reconstruction behaves when the requested viewpoints depart structurally from that distribution.

Across all the five scenes, unseen trajectories degrade relative to seen held-out views for both evaluated models on all three reported image-quality metrics. For Nerfacto, mean PSNR decreases from 19.54 to 16.47 dB in Church, 17.01 to 15.15 dB in Victorian Garden, and 19.80 to 16.82 dB in Village Street. For Splatfacto, the corresponding decreases are 22.17 to 16.86 dB, 19.23 to 15.00 dB, and 24.50 to 16.08 dB. The magnitude is scene- and model-dependent, so these results should be interpreted as evidence from the current PIVOT v1 scenes rather than as a universal generalization law. Qualitatively, the independently captured unseen views also exhibit stronger blur, loss of detail, and rendering artifacts than representative seen views.

\subsection{Pose-Space Distance as an Evaluation Descriptor}

If reconstruction quality degrades as directed pose Chamfer distance increases, the metric can provide a compact descriptor of how far an evaluation trajectory lies from the training pose distribution. However, such an empirical relationship should not be interpreted as establishing pose-space distance as a complete measure of novel-view difficulty.

In the five scenes, seen views cluster close to the training pose distribution and generally achieve lower LPIPS, whereas unseen trajectories occupy a broader range of pose-space distances and generally higher LPIPS. The relationship is not strictly monotonic: trajectories with similar pose-space distance can differ noticeably in rendering quality. This is expected because the metric describes camera-pose coverage rather than visibility, texture, occlusion, or scene content. We therefore use directed pose Chamfer distance as a descriptive covariate rather than claiming it is a complete predictor of novel-view difficulty.

\subsection{Sensitivity to Pose Source}

The dual-pose representation allows translation and rotation to be changed independently. This is useful because sensor-derived pose errors need not affect the two components equally, and reconstruction methods may exhibit different sensitivity to each.

The BM2 runs consistently favor OO, indicating a substantial benefit from the COLMAP-optimized camera trajectory. In Church, replacing either translation or rotation with measured values reduces PSNR by several decibels for both models, with MM approximately 6.7--7.6 dB below OO. Village Street shows an even larger effect: MM is 6.93 dB below OO for Nerfacto and 11.56 dB below OO for Splatfacto. The relative ordering of OM and MO is not consistent across scenes and models, so the current evidence does not support a general claim that translation or rotation error alone is dominant.

\subsection{Sensitivity to Intrinsic Source}

The intrinsic benchmark contrasts the favorable offline setting in which intrinsics are optimized for the current scene with the deployment-oriented setting in which a fixed physical calibration is reused.

Using the fixed OpenCV calibration instead of COLMAP-optimized per-scene intrinsics reduces reconstruction quality in every all BM3 scene/model pair. The PSNR reduction ranges from 1.75 dB for Victorian Garden/Nerfacto to 9.33 dB for Village Street/Splatfacto. The effect is therefore substantial but strongly scene- and model-dependent. Importantly, this experiment measures sensitivity to the specific independently estimated calibration used in PIVOT v1; it should not be interpreted as evidence that reusable physical calibration is inherently inferior by the same amount in other systems.

\section{Limitations and Future Work}
\label{sec:limitations}

\subsection{Geometry-Aware Trajectory Distance}

The current directed pose Chamfer distance compares camera translation and orientation in pose space. Camera-pose similarity, however, does not necessarily imply similarity in scene visibility. Two camera poses can be spatially close and similarly oriented while lying on opposite sides of an occluding structure. Their pose-space distance can therefore be small even though they observe substantially different scene content.

This is a fundamental limitation of any trajectory descriptor based only on camera pose: pose-space proximity measures where cameras are and how they are oriented, but not which parts of the scene they can observe.

A promising extension is a geometry-aware trajectory similarity measure. Given an available scene reconstruction, such as the COLMAP sparse point cloud, visible scene geometry could be projected into each camera and the overlap between observations estimated, for example using an Intersection-over-Union-based visibility measure. A future trajectory metric could therefore combine:
\begin{itemize}
    \item translation difference, describing camera separation;
    \item rotation difference, describing viewing-orientation difference; and
    \item visibility overlap, describing how much reconstructed scene geometry is jointly observed.
\end{itemize}

Such a metric could distinguish cameras that are close in pose space but observe different geometry from cameras that are farther apart while retaining substantial scene overlap. The current directed pose Chamfer distance nevertheless remains useful as a simple, scene-geometry-independent descriptor that can be computed directly from camera poses. Geometry-aware distance should therefore be viewed as complementary rather than as a replacement in all settings.

\subsection{Dataset Scale and Capture Platform}

\pivot{} v1 contains five real-world scenes captured with a single DJI Mini 4 Pro drone platform. This controlled setup is useful for isolating the target experimental variables, but it limits conclusions about other cameras, localization systems, environments, and motion platforms. The processing architecture is designed to support additional capture devices, and future releases can extend the scene and device diversity while retaining the same trajectory protocol.

\subsection{Physical Calibration Quality}

The BM3 comparison depends on the quality of the independently estimated physical camera calibration. The calibration used for PIVOT v1 has a relatively high reprojection error (approximately 4 pixels in the calibration run), so part of the observed gap between fixed and COLMAP-optimized intrinsics may reflect calibration quality rather than an unavoidable limitation of reusable calibration. BM3 should therefore be interpreted as a sensitivity experiment for the calibration available in this release. Future captures should use a higher-quality calibration procedure and test calibration transfer across scenes and devices.

\subsection{Dependence on SfM Registration}

Some trajectories are deliberately difficult for SfM and may not register completely. \pivot{} records registration rates rather than silently discarding this behavior, since registration difficulty is itself relevant to trajectory design. Nevertheless, experiments requiring COLMAP-optimized poses cannot use missing optimized poses without either dropping those frames or explicitly falling back to measured poses. Benchmark configurations must therefore report how unregistered frames are handled.

\subsection{Sparse Geometry for Visibility Analysis}

The proposed geometry-aware extension would itself depend on reconstruction quality. Sparse COLMAP points do not provide complete scene visibility and may be biased toward textured, repeatedly observed regions. Future geometry-aware metrics should therefore study sensitivity to the underlying geometric representation.

\section{Reproducibility and Release}

The \pivot{} source code is released under the MIT License, while the dataset is released under Creative Commons Attribution--NonCommercial 4.0 International (CC BY-NC 4.0). The project separates the core PIVOT processing environment from a dedicated Nerfstudio integration environment used for Nerfacto/Splatfacto training and benchmark execution.

\paragraph{Project repository.}
\url{https://github.com/maryraymond/PIVOT/tree/v1.0.0}

\paragraph{Nerfstudio integration.}
\url{https://github.com/maryraymond/nerfstudio_PIVOT_integration/tree/v1.0.0}

\paragraph{Dataset}
\url{https://huggingface.co/datasets/MaryRaymond/PIVOT/tree/v1.0.0}

\paragraph{containers}
\begin{itemize}
    \item \pivot{} \url{ghcr.io/maryraymond/pivot:1.0.0}
    \item Nerfstudio integration \url{ghcr.io/maryraymond/nerfstudio_pivot_integ:1.0.0}
\end{itemize}

\section{Conclusion}

We presented \pivot{}, a multi-trajectory dataset and testbed designed to separate several favorable assumptions that are often coupled in 3D reconstruction evaluation. By preserving measured and optimized poses, calibrated and optimized intrinsics, and explicit trajectory identity, \pivot{} enables controlled experiments on pose quality, calibration, capture trajectory, and novel-view generalization. Its directed pose Chamfer distance provides a simple pose-space description of evaluation-trajectory coverage, while the benchmark design explicitly distinguishes held-out views on represented trajectories from complete unseen trajectories.

Across all scenes, the initial experiments show a consistent gap between seen held-out views and independently captured unseen trajectories, substantial degradation when measured pose components replace COLMAP-optimized poses, and measurable sensitivity to the intrinsic calibration source. The magnitude of these effects varies by scene and model, and the pose-space distance is descriptive rather than a complete predictor of rendering difficulty. Together, these results support the use of trajectory identity, pose source, and intrinsic source as explicit evaluation dimensions. The broader goal of \pivot{} is to make the gap between reconstruction benchmarks and real-world capture conditions easier to measure, reproduce, and study.

\bibliographystyle{IEEEtran}

\end{document}